\documentclass{article}

\usepackage[square, comma, sort, numbers]{natbib}
\usepackage[preprint]{neurips_data_2024}

\usepackage[utf8]{inputenc} 
\usepackage[T1]{fontenc}    
\usepackage{hyperref}       
\usepackage{url}            
\usepackage{booktabs}       
\usepackage{amsfonts}       
\usepackage{nicefrac}       
\usepackage{microtype}      
\usepackage{xcolor}         
\usepackage{algpseudocode}
\usepackage{algorithm}
\usepackage{colortbl}   
\usepackage[most]{tcolorbox}
\usepackage{amsmath}
\usepackage{multirow}
\usepackage{multicol}
\usepackage{pifont}
\usepackage{amsfonts}
\usepackage{hyperref}
\usepackage{graphicx}
\usepackage{wrapfig}

\newcommand{\highg}{\cellcolor{backgreen}}
\definecolor{deepgreen}{RGB}{0, 100, 0}
\definecolor{deepred}{RGB}{139, 0, 0}

\definecolor{backblue}{RGB}{210, 230, 250}
\newcommand{\high}{\cellcolor{backblue}}
\usepackage{xspace}
\usepackage[shortlabels]{enumitem}

\definecolor{backblue}{RGB}{210, 230, 250}
\definecolor{backgreen}{RGB}{226, 240, 217}
\definecolor{backred}{RGB}{255, 223, 223}

\hypersetup{
    colorlinks=true,
    linkcolor=blue,
    citecolor=blue,
}
\newcommand{\highr}{\cellcolor{backred}}

\newcommand{\dataset}{\textsc{We-Math}\xspace}

\usepackage{graphicx}       

\usepackage{CJKutf8}        
\usepackage{enumitem}       

\newcommand{\chinese}[1]{%
    \begin{CJK*}{UTF8}{gbsn}%
    #1%
    \end{CJK*}%
}

\usepackage[normalem]{ulem}

\title{\dataset: Does Your Large Multimodal Model Achieve Human-like Mathematical Reasoning?}

\author{%
Runqi Qiao$^1$\thanks{Equal contribution.} \ , Qiuna Tan$^1$$^*$, Guanting Dong$^1$, Minhui Wu$^2$, Chong Sun$^2$, Xiaoshuai Song$^1$,\\
{\bf Zhuoma GongQue$^1$,} {\bf Shanglin Lei$^3$,} {\bf Zhe Wei$^1$,} {\bf Miaoxuan Zhang$^1$,} {\bf Runfeng Qiao$^4$,} \\
{\bf Yifan Zhang$^1$,} {\bf Xiao Zong$^1$,} {\bf Yida Xu$^1$,} {\bf Muxi Diao$^1$,} {\bf Zhimin Bao$^2$,}\\ {\bf Chen Li$^2$,} {\bf Honggang Zhang$^1$}\thanks{Corresponding author}
\\
$^1$Beijing University of Posts and Telecommunications,$^2$Wechat, Tencent Inc.,\\
$^3$Huazhong University of Science and Technology, $^4$Beijing Institute of Technology\\
\href{https://We-Math.github.io}{https://We-Math.github.io}\\
}

\begin{document}

\maketitle
\begin{abstract}
Visual mathematical reasoning, as a fundamental visual reasoning ability,  has received widespread attention from the Large Multimodal Models (LMMs) community. 
Existing benchmarks focus more on the result-oriented performance, but neglecting the underlying principles in knowledge acquisition and generalization. 
Inspired by human-like mathematical reasoning, we introduce \dataset, the first benchmark specifically designed to explore the problem-solving principles beyond the end-to-end performance. 
We meticulously collect and categorize 6.5K visual math problems, spanning 67 hierarchical knowledge concepts and 5 layers of knowledge granularity. We firstly decompose composite problems into sub-problems according to the required knowledge concepts and introduce a novel four-dimensional metric, namely \emph{Insufficient Knowledge (IK)}, \emph{Inadequate Generalization (IG)}, \emph{Complete Mastery (CM)}, and \emph{Rote Memorization (RM)} to hierarchically assess inherent issues in LMMs' reasoning process. 
  With \dataset, we conduct a thorough evaluation of existing LMMs in visual mathematical reasoning and reveal a negative correlation between solving step and problem-specific performance. We confirm the \emph{IK} issue of LMMs can be effectively improved via knowledge augmentation strategy. 
  More notably, the primary challenge of GPT-4o has significantly transitioned from \emph{IK} to \emph{IG}, establishing it as the first LMM advancing towards the knowledge generalization stage. In contrast, other LMMs exhibit a marked inclination towards \emph{Rote Memorization} -- they correctly solve composite problems involving multiple knowledge concepts, yet fail in answering sub-problems. We anticipate that \dataset will open new pathways for advancements in visual mathematical reasoning for LMMs. The \dataset data and evaluation code are available at \href{https://github.com/We-Math/We-Math}{https://github.com/We-Math/We-Math}.
\end{abstract}
\begin{figure}[t]
    \centering
    \resizebox{\textwidth}{!}{
    \includegraphics{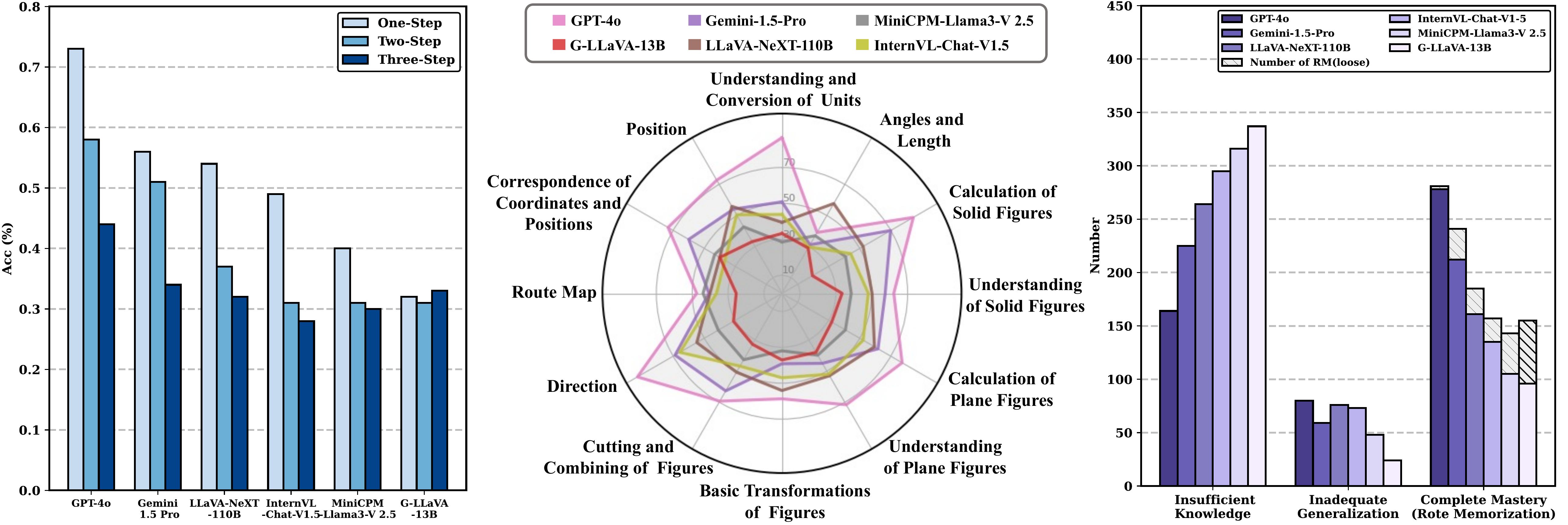}
    }
    \vspace{-0.2cm}
    \caption{ Overview of LMMs' performances on \dataset. Figures from left to right illustrates the (1) accuracy of different LMMs on various problem-solving steps, (2) the performance in different visual mathematics categories and (3) the result in knowledge based reasoning evaluation. 
     }
       \vspace{-0.4cm}
    \label{fig:eval_main}
\end{figure}
\section{Introduction}

\emph{``I think, therefore I am.'' --- René Descartes}

Human cognitive and reasoning patterns have profoundly shaped the progress of deep learning~\cite{lecun2015deep}. Initially, the design of neural networks~\cite{lecun1998gradient} is inspired by the brain's neuronal mechanisms. It uses convolution kernels and hierarchical network to mimic human cognitive process of knowledge acquisition. 
Recently, Transformers~\cite{vaswani2017attention} employ attention mechanisms to handle multiple information flows and quickly focus on critical content, thereby achieving more efficient and in-depth sequential learning.
Owing the scalability of the Transformer architecture and pre-training techniques, Large Language Models (LLMs)~\cite{ouyang2022training,achiam2023gpt,llama2,anil2023palm} and Large Multimodal Models (LMMs)~\cite{liu2024visual,dai2024instructblip,li2023blip,zhang2023llamaadapter,gao2023llamaadapter,bai2023qwenvl,su2023pandagpt,ye2024mplugowl,zhu2023minigpt} showcases strong reasoning abilities that parallel human performance across a wide range of tasks and provide a glimpse into the early outlines of Artificial General Intelligence (AGI). 

Mathematical reasoning is a critical capability of foundational models.
Existing methods employ Chain of Thought (COT)~\cite{wei2023chainofthought}, Program of Thought (POT)~\cite{pot,pal}, 
Tool-integrated techniques~\cite{ToRA,mathcoder} and data augmentation strategies~\cite{luo2023wizardmath,yuan2023scaling,yu2309metamath,li2023query} to guide LLMs towards emulating human-like reasoning patterns.
In a more challenging scenario, Visual mathematical reasoning requires the model to accurately decode the visual information in image and perform reasoning based on the textual problem. With the rapid advancements of large multimodal models (LMMs)~\cite{2023GPT4VisionSC,geminiteam2024gemini}, researchers progressively utilize the LMMs for solving visual mathematical problems~\cite{yang2023dawn,gao2023gllava}. These studies provide valuable insights into the ongoing improvements in multi-modal logical thinking capabilities.

To systematically evaluate visual mathematical reasoning capabilities, previous efforts~\cite{lu2021inter, seo2015solving, chen2021geoqa, chen2022unigeo} have focused on challenging geometric problems. Recently, several benchmarks~\cite{lu2023mathvista,zhang2024mathverse} expand the scope to include a wider range of disciplines. However, these benchmarks rely solely on end-to-end results for assessment, which fails to identify inherent issues within the LMMs' reasoning process. Moreover, MathVerse~\cite{zhang2024mathverse} attempt to directly evaluate reasoning paths based on reference answers, but limitations remain due to the knowledge-intensive nature of mathematical reasoning.
While noticing that humans solve complex math problems through gradually mastering and generalizing the \emph{knowledge concepts}~\cite{fitzpatrick2008euclid},
we claim a fair evaluation of a model's reasoning process should be based on knowledge concepts. Therefore, we pose two questions about mathematical reasoning evaluation:

\emph{\textbf{Q1}: Does the correct answer truly reflect LMM's ability to reason through such problems accurately?}

\emph{\textbf{Q2}: Does an incorrect answer suggest a lack of foundational knowledge in LMM's reasoning process?}

As the response, we present \textbf{\dataset}, a pioneering benchmark for conducting an in-depth analysis of the underlying principles of LMMs in visual mathematical reasoning. \dataset consists of over 6.5K meticulously selected visual math problems,
which can be categorized into 5 layers of knowledge granularity across 67 knowledge concepts for ensuring a comprehensive coverage.
We observe that real-world math problems typically encompass multiple foundational knowledge concepts, and their difficulty is directly related to the number of concepts involved. Upon this, we decouple the model's ability to solve composite problems with $k$ knowledge concepts into two stages:

1) LMMs can solve $k$ individual sub-problems corresponding its knowledge concept;

2) LMMs reason out the final answer by integrating the k individual knowledge concepts.

The above process can be formulated as follows:
\begin{equation}
\label{eq:intro}
P(Y|X) = \prod_{i=1}^{k} P(y_i|x_i) \cdot P_{\text{reason}}, 
\end{equation}
where $(X,Y)$ and $(x_i,y_i)$ denote the $(question, answer)$ pairs in a composite mathematical problem and the $i$-th sub-problem, respectively. $P_{\text{reason}}$ stands for the LMMs' reasoning capacity. 
It is evident that assessing the reasoning process cannot be based solely on final answers. To decompose a composite problem into individual sub-problems according to the invoked knowledge concept, we select 1.5k high-quality problems with multiple knowledge concepts in \dataset. Following equation \ref{eq:intro}, these composite problems are gradually decomposed by expert annotators into one-step problem $(x_i,y_i)$. Motivated by human reasoning patterns, \dataset further introduces a four-dimensional metric to precisely evaluate the inherent gaps in LMMs' problem-solving abilities, namely \emph{Insufficient Knowledge (IK)}, \emph{Inadequate Generalization (IG)}, \emph{Complete Mastery (CM)}, and \emph{Rote Memorization (RM)}. To further tackle the fundamental \textit{IK} issue, we propose a heuristic knowledge concept augmented (KCA) strategy, constructing descriptions for 67 knowledge concepts from Wikipedia~\cite{Wikipedia} and textbooks, thereby providing essential knowledge for LMMs' reasoning.

Figure \ref{fig:eval_main} illustrates our overview experimental results. Not surprisingly, GPT-4o~\cite{GPT4o} achieves the best overall performance across different visual mathematics categories. Closed-source LLMs (GPT-4V, Gemini 1.5 Pro) and LMMs with larger parameter scales (LLaVA-NeXT-110B~\cite{liu2024llavanext}) generally exhibit superior visual mathematical reasoning capabilities. However, most LMMs perform significantly worse on multi-step problems compared to one-step problems, suggesting that the number of knowledge concepts is positively correlated with the question's difficulty and negatively correlated with LMM performance. In specialized disciplines, most LMMs excel in calculation but consistently struggle with fine-grained visual measurement ("Angles and Length"). 

For reasoning evaluation, we emphasize that mastery of knowledge concepts is fundamental. 
Unfortunately, most LMMs still suffer from \emph{Insufficient Knowledge} issue, especially smaller-scale models (e.g., over 350 \textit{IK} issues in LLaVA-1.6-7B and DeepSeek-VL-1.3B). GPT-4o significantly addresses this knowledge gap, establishing it as the first LMM advancing towards the knowledge generalization stage. More notably, several LMMs still exhibit a marked inclination towards \textit{Rote Memorization} (e.g. G-LLaVA-13B nearly 36\% in \textit{RM (Loose)}), raising doubts about whether current LMMs truely possess the mathematical reasoning capability. In addition, our proposed KCA strategy substantially reduces the \textit{IK} issue in LMMs, and error analysis further provides empirical guidance towards human-like reasoning. We anticipate that \dataset will open new pathways for advancements in visual mathematical reasoning in LMMs.




\section{\dataset}
\label{headings}
\begin{figure}[t]
    \centering
    \resizebox{1.0\textwidth}{!}{
    \includegraphics{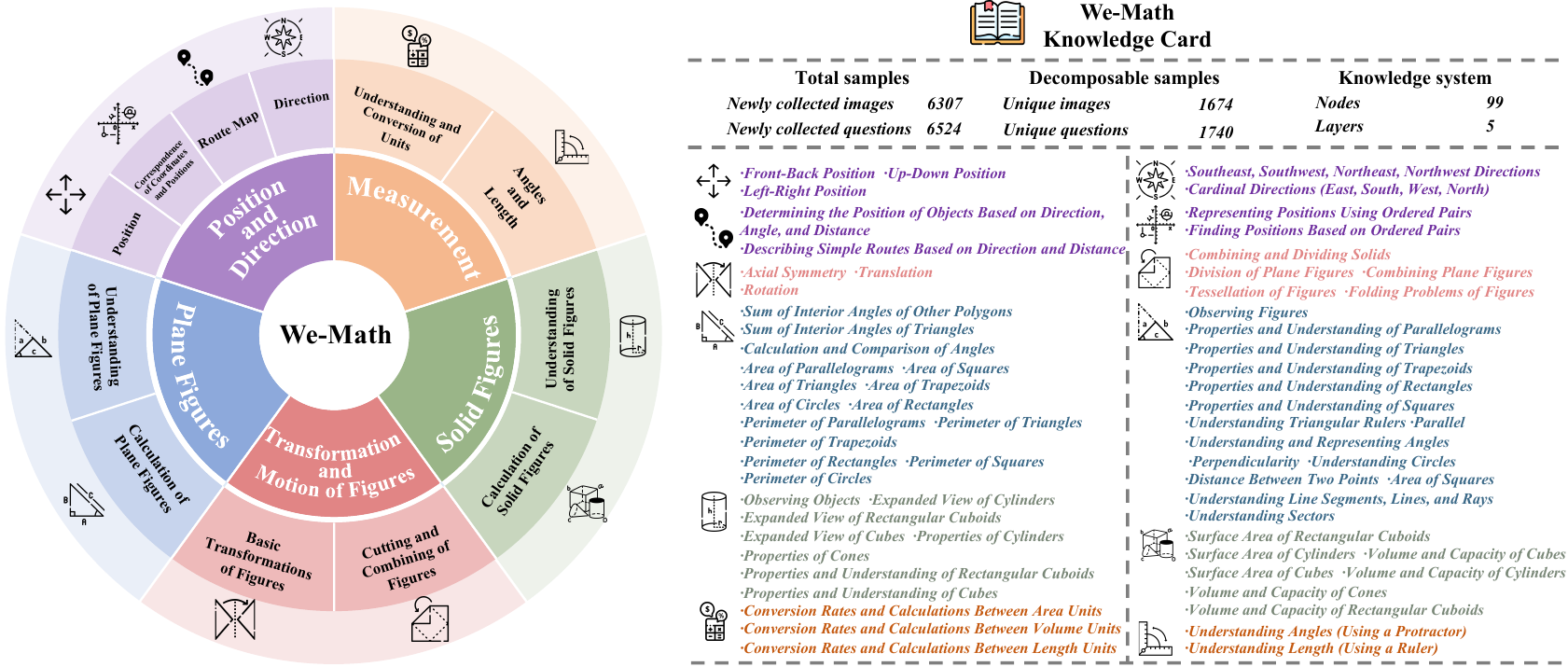}
    }
    \vspace{-0.2cm}
    \caption{Overview diagram and the statistics of \dataset. The left and right side shows the first two layers of \dataset's categories and information of different samples and terminal nodes.}
    \vspace{-0.2cm}

    \label{fig:main}
\end{figure}

\textbf{Overview of \dataset.} As previously mentioned, existing benchmarks tend to be result-oriented, while overlooking the essence of solving mathematical problems. 
%
%
This leads to the generation of some counterintuitive evaluation conclusions. For example, conclusions in MathVista~\cite{lu2023mathvista} indicate that LMMs exhibit superior performance on university-level problems compared to elementary-level ones.
%
Different from existing benchmarks, as shown in Figure~\ref{fig:main}, \dataset is constructed around textbook knowledge units, decomposing composite problem solutions into sub-problems based on the knowledge concepts. 
%
\dataset has the following characteristics:

(1) \textbf{Hierarchical Knowledge Structure.} \dataset strictly adheres to the knowledge presented in mathematics textbooks, featuring a rigorous hierarchical and multi-category architecture. 
%
%
It ensures the independence of knowledge concepts within the same level, while establishing logical relationships among concepts at different hierarchical levels.

(2) \textbf{Knowledge based Reasoning Evaluation.} \dataset is designed to explore how LMMs solve problems. Drawing upon that humans tackle problems incrementally by leveraging fundamental knowledge concepts, we break down complex mathematical problems into more manageable sub-problems. Furthermore, we employ diverse measurement dimensions for meticulous evaluations.

(3) \textbf{Knowledge Concept Augmentation.} To alleviate the inherent issues during the problem-solving process, we heuristically introduce descriptions for 67 knowledge concepts from Wikipedia and textbooks, thereby providing essential knowledge support for the reasoning processes of LMMs.


\subsection{Hierachial Structured Dataset Composition}

\textbf{Hierachial Knowledge Structure.}
\dataset emphasizes fundamental math skills, believing that complex mathematical reasoning is built upon foundation of basic mathematical reasoning processes.
Based on extensive research, mathematical problems are categorized into five distinct types, namely \textit{Plane Figures, Solid Figures, Transformations and Movements of Shapes, Positions and Directions, Measurements}. These five categories can be decomposed into 12 typical problems, which are further decomposed as 67 knowledge concepts (terminal nodes in the structure). We collect problems according to this tree structure and constrain that each terminal node contains a strict range of 10-40 samples. This rule ensures data balance across domains.
%

\textbf{Data Collection and Annotation.}
%
%
All problems (6.5K) in \dataset are sourced from publicly authoritative mathematics websites and subsequently organized based on our defined knowledge structure.
We employ three expert annotators to manually label each question with knowledge concepts.
%
Cross-validation is performed to ensure at least two experts have identical annotations for the same question.
%
%
Samples with notably inconsistent labels will be considered of low quality and subsequently excluded.
%
To prepare for the subsequent decomposition of problems, we further annotate problem-solving steps based on the knowledge concepts labels.
We categorize each problem into three distinct classes: "One-Step", "Two-Step", and "Three-Step". This categorization enables us to gain a deeper understanding on how LMMs solve problems.
%
Further details about annotation can be found in Appendix. After the annotation, all problems are double checked by an expert team in terms of three aspaces: (1) The consistency between the questions and dagrams; (2) The correctness of the answers to the questions; (3)  The alignments between problems and the 67 knowledge concepts. 
%

%
%


\subsection{Knowledge based Reasoning Evaluation}
\label{sec:decompose}

\begin{figure}[ht]
    \centering
    \resizebox{\textwidth}{!}{
    \includegraphics{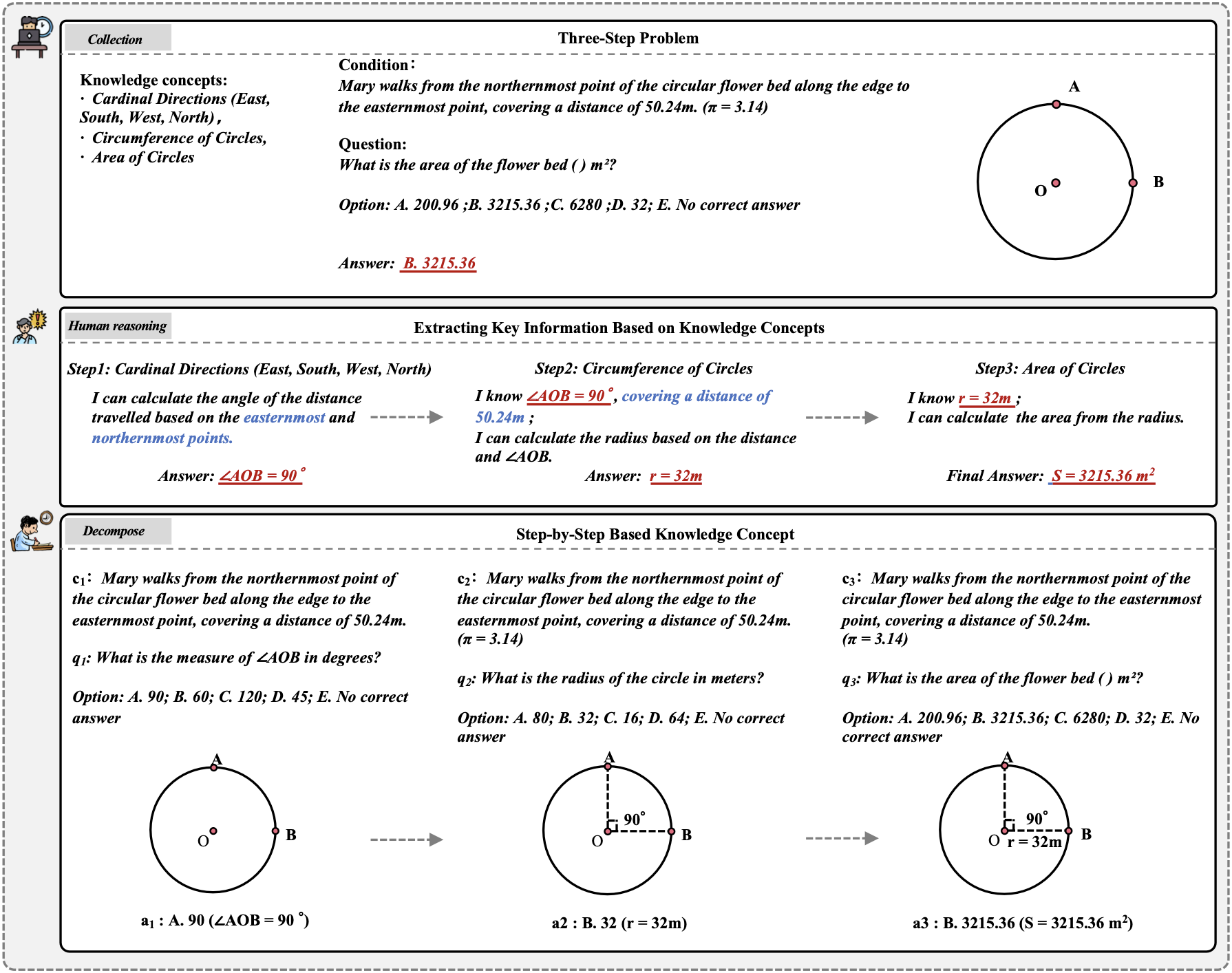}
    }
    \caption{The pipeline of knowledge-based data decomposition (an example of a three-step problem in \dataset).}
    \label{fig:kuangjia}
\end{figure}

\textbf{Problem Definition.} For the visual mathematical reasoning task, given text question $Q_i$, image $I_i$ and corresponding answer $A_i$. We define the LMMs evaluation dataset $D_{\rm eval} = \{(Q_i,I_i,A_i) | K_i,C_i\}_{i=1}^N$. where $K_i$ and $C_i$ are two prior constraints for question $Q_i$. In detail, $K_i=\{k_i\}_{i=1}^M$ denote $M$ knowledge concepts within the question. $C_i$ represents the prerequisite conditions needed to solve the problem $Q_i$ (see Figure~\ref{fig:kuangjia} for example). 
For the convenience of presenting, we define the problem containing $k$ knowledge concepts as a "$k$-step problem" in our paper. 




%
\textbf{Knowledge-based Data Decomposition.} Real-world mathematical problems are composed of multiple atomic knowledge concepts. However, existing benchmarks usually overlook this information, leading to unreasonable evaluation results.
Inspired by Euclid's Elements~\cite{fitzpatrick2008euclid}, we argue that the evaluation of mathematical reasoning ability in LMMs essentially involves assessing their mastery of fundamental knowledge concepts.
%
%
It is quite a natural and objective way to exploit basic knowledge concepts for reasoning evaluation of LMMs. 
Given an $i$-th test sample $\{(Q_i, I_i, A_i) | K_i, C_i\} \in D_{\dataset}$ with M concepts $K_i=\{k_i^m\}_{m=1}^M$, we ask human experts to decompose each problem step by step into $M$ sub-problems based on knowledge concepts, which can be formulated as:
%
\begin{equation}
\{(q_i^m, i_i^m, a_i^m) | k_i^m, c_i^m\}_{m=1}^M = \mathop{\rm Decompose}_{(Q_i, I_i, A_i) \in D_{\dataset}}\{(Q_i, I_i, A_i) | K_i, C_i\}
\end{equation}
where $k_i$, $c_i$ denote the individual knowledge and prior condition for the sub-problem. 
``$\mathop{\rm Decompose}$'' represents the Human decomposition process based on $M$ knowledge concepts. 
%
%
To ensure logical coherence of decomposition, the condition $c_i^m$ is initialized as $C_i$. Then it is
recursively computed by concatenating the answer $a_i^{m-1}$ and condition $c_i^{m-1}$ of the $m-1$-th concept:
\begin{equation}
c_i^m = c_i^{m-1} + \textit{a}_i^{m-1} \quad \text{for } m = 2, 3, \ldots, M
\end{equation}
where ``$+$'' denotes the concatenation operation. In addition, the equation $\left\{
\begin{array}{l}
q_i^M = Q_i \\
a_i^M = A_i
\end{array}
\right\}$ must be satisfied, which is also a constraint for logical coherence.
%
Finally, we can obtain the original multi-step problem and $M$ one-step sub-problems for reasoning evaluation. The overall pipeline of Knowledge-based Data Decomposition are displayed in the left side of Figure \ref{fig:kuangjia}.

\textbf{Metric for Reasoning Evaluation.}
%

\begin{figure}[!t]
    \centering
    \resizebox{\textwidth}{!}{
    \includegraphics{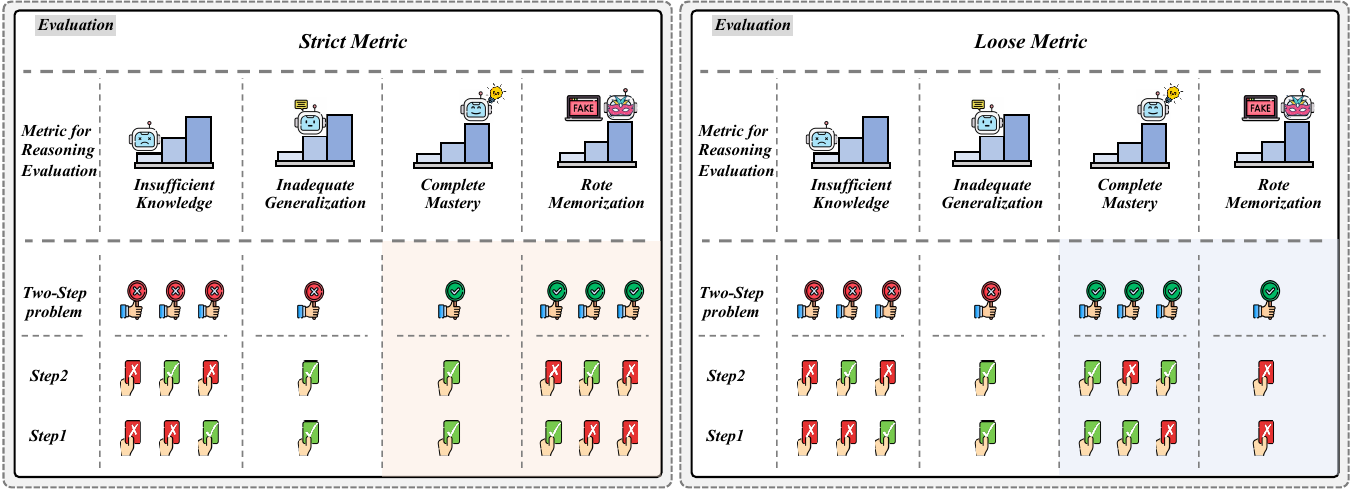}
    }
    \vspace{-0.3cm}
    \caption{An example of the four-dimensional metrics for evaluating a two-step problem, using both loose and strict settings.}
    \label{fig:metric-main}
\end{figure}

Based on the decomposed multi-step problems, we further reveal the inherent issues of LMMs in problem-solving process. 
We feed both the $M$ one-step sub-problems and the original problem into LMMs, and classifying the responses into four categories:

%

\textbf{1. Insufficient Knowledge (IK):} Part of one-step problems contain errors, and the multi-step problem is wrong. It is reasonable because model's insufficient grasp of single knowledge concept may lead to errors in multi-step problem.

\textbf{2. Inadequate Generalization (IG):} One-Step problems are all correct, but the multi-step problem is incorrect. 
This is also considered reasonable. While LMMs are capable of understanding individual knowledge concepts, they may struggle to generalize that knowledge to solve composite problems.

\textbf{3. Complete Mastery (CM):} One-Step problems are all correct, and multi-step problem is also answered correctly. This result demonstrates that the model's results are both reliable and accurate.

\textbf{4. Rote Memorization (RM):} One-Step problems contain errors, but the multi-step problem is answered correctly, which contradicts human logical thinking. If a model can solve composite multi-step problems but fails to answer the one-step problems needed in the process, it raises doubts about the model's reliability.

Considering \emph{IK}, \textit{IG}, and \textit{CM}, it is evident that results falling under the \textit{IG} category are generally more preferred compared to those classified as \textit{IK}.
The reason is that \textit{IK} reflects the model’s struggle with both single and multiple knowledge concepts, while \textit{IG} shows the model's proficiency one-step problem. 
By enhancing the model's generalization ability in reasoning process, we can potentially shift results from \emph{IG} to \textit{CM}.
Therefore, we establish a reasoning capability hierarchy as $\textit{IK} < \textit{IG} < \textit{CM}$. 
We believe that \textit{RM} is an unreasonable scenario (models can solve multi-step problems without mastering one-step problems, which completely contradicts human reasoning intuition).

In light of the model's instability, the current criteria for determining whether a result belongs \textit{RM} is strict.
We thus propose a more flexible loose metric.
As illustrated in Figure \ref{fig:metric-main}, the TFT and FTT situations in the two-step problems are regard as \textit{CM} (rather than \textit{RM}), according to the loose metric. We also discuss the situation of four-dimensional metrics on three-problem in Appendix \ref{sec:Metric}.
We propose the following metric to judge the reliability of the model's reasoning process:
\begin{equation}
\label{eq:metric1}
S_{\rm IK} = \frac{N_{\rm IK}}{N}, \quad S_{\rm IG} = \frac{N_{\rm IG}}{N}, \quad S_{\rm CM} = \frac{N_{\rm CM}}{N}, \quad S_{\rm RM} = \frac{N_{\rm RM}}{N_{\rm RM}+N_{\rm CM}}
\end{equation}

where \( \text{N} \) denotes the total number of samples and \( N_{IK} \), \( N_{IG} \), \( N_{CM} \), \( N_{RM} \) represents the number of samples for a specific situation. Therefore, we obtain our final reasoning confidence scores:

\begin{equation}
\text{Score}_{\text{average}} = \alpha S_{\rm IK} + \beta S_{\rm IG} + S_{\rm CM}
\end{equation}


where $\alpha, \beta$ denotes the weight for each case. To ensure the reasoning capability hierarchy is "\textit{IK} < \textit{IG} < \textit{CM}", we control the params $\alpha<\beta<1$, and set the default value of $\alpha$ to 0.0 and $\beta$ to 0.5.



\subsection{Knowledge Concept Augmentation}
\begin{figure}[ht]
    \centering
    \resizebox{\textwidth}{!}{
    \includegraphics{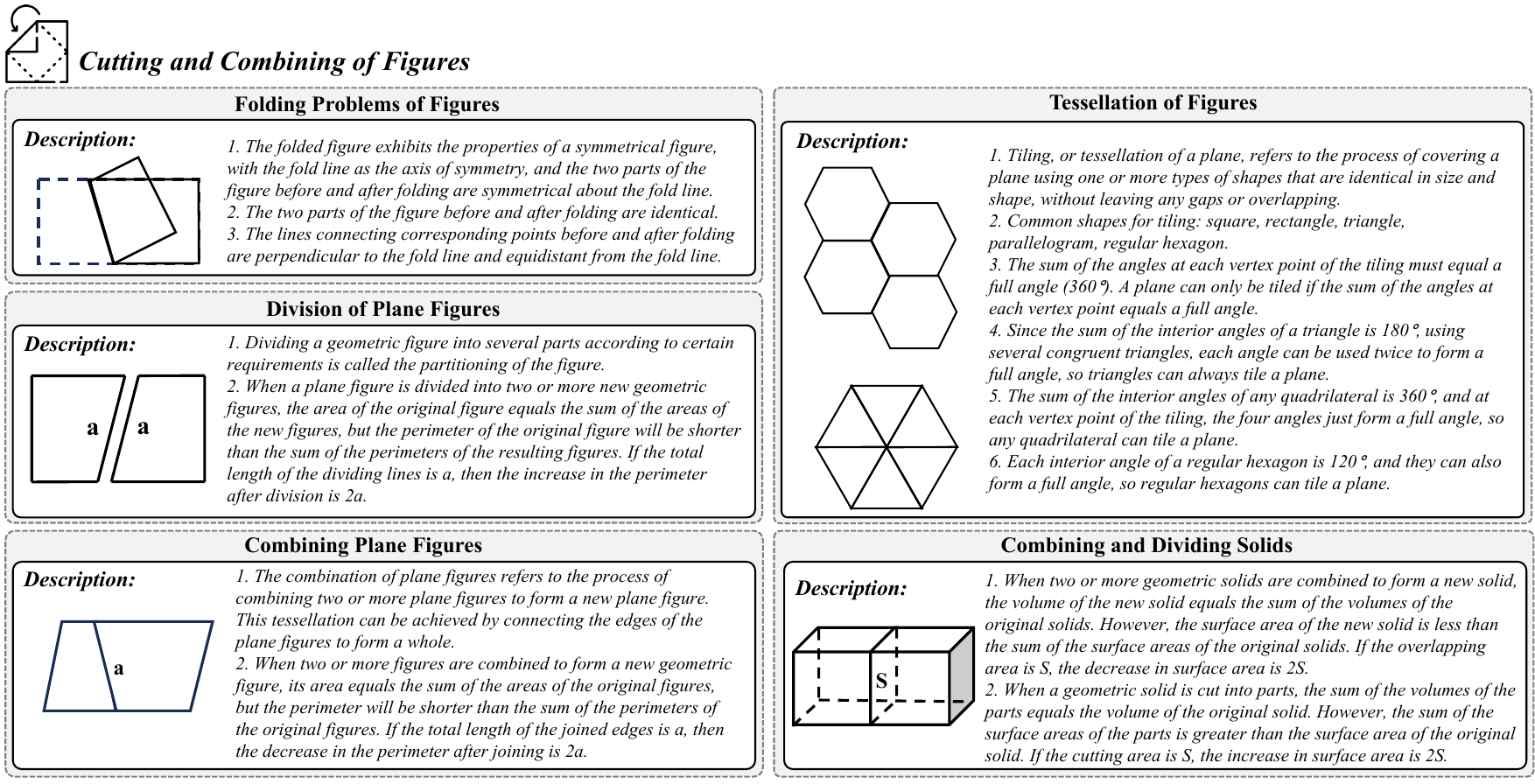}
    }
    \caption{The cases of knowledge concept cards.}
    \label{fig:kn card}
\end{figure}

In the previous section, we identify the \textit{Insufficient Knowledge (IK)} as the foundation challenge in mathematical reasoning. 
To heuristically tackle this issue, we enlist human experts to create 67 knowledge concept cards, which is essential for LMM's reasoning process. 
Initially, expert annotators offer precise summaries derived from the definitions in Euclid's Elements~\cite{fitzpatrick2008euclid}, Wikipedia and textbooks.
Subsequently, these experts further condense the content examined by a series of questions related to a specific knowledge concept, extracting crucial knowledge hints for incorporation into the knowledge cards. 
After several rounds of review, we confirm the accuracy and utility of each card. 
Figure \ref{fig:kn card} showcases typical knowledge concept cases and their descriptions. 
Consequently, with a given problem $Q_i$ and its respective knowledge concept $K_i$, LMMs utilize the relevant knowledge cards to deduce the answer $A_i$. The detailed information of KCA can be found in Appendix.

\section{Experiment}


\paragraph{Evaluation Protocols.} To accelerate the evaluation speed, \dataset comprises a \textit{testmini} set with 1740 samples, including 1215 one-step samples, 360 two-step samples, and 165 three-step samples. In subsequent experiments, we utilize the \dataset \textit{testmini} subset for evaluation. For automated evaluation, we standardize all samples into a multiple-choice format. We use regex to match the LMMs' predictions and then calculate their accuracy against the ground-truth answers for main results. For analyses in section \ref{exp:reason} and \ref{exp:error}, we utilize the four-dimensional metric described in section \ref{sec:decompose} for assessment. To avoid LMMs deduce answers from options, we introduce an extra \textit{uncertain} option to mitigate this issue.

\paragraph{Evaluation Models.} We examine the performance of foundation models across two distinct categories on \dataset: (a) Closed-source LMMs: GPT-4o~\cite{GPT4o}, GPT-4V~\cite{2023GPT4VisionSC}, Gemini 1.5 Pro~\cite{gemini}, Qwen-VL-Max~\cite{bai2023qwenvl}, (b) Open-source LMMs: LLaVA-NeXT-110B, LLaVA-NeXT-70B~\cite{liu2024llavanext}, LLaVA-1.6-13B, LLaVA-1.6-7B~\cite{liu2023improvedllava}, DeepSeek-VL-1.3B, DeepSeek-VL-7B~\cite{lu2024deepseekvl}, Phi3-Vision-4.2B~\cite{abdin2024phi3}, MiniCPM-Llama3-V 2.5~\cite{hu2024large}, InternLM-XComposer2-VL-7B~\cite{internlmxcomposer2}, InternVL-Chat-V1.5~\cite{chen2024internvl}, GLM-4V-9B~\cite{glm2024chatglm}, LongVA~\cite{zhang2024longcontexttransferlanguage}, G-LLaVA-13B~\cite{gao2023gllava}.

\subsection{Main Result }
\label{exp:main}

\begin{table}[!t]

\caption{\textbf{Accuracy scores of LMMs on the \textit{testmini} subset of \dataset.} The first 3 columns report the overall performance on one-step, two-step, three-step problems, while the other columns display the result on one-step problems in different problem categories. The highest accuracy for closed-source and open-source LMMs is marked in \colorbox{backblue}{blue} and \colorbox{backgreen}{green} respectively. (S1: one-step problem, S2: two-step problem, S3: three-step problem, Mem: Measurement, PF: Plane Figures, SF: Solid Figures, TMF: Transformations and Motion of Figures, PD: Position and Direction. AL: Angles and Length, UCU: Understanding and Conversion of Units, CPF: Calculation of Plane Figures, UPF: Understanding of Plane Figures, CSF: Calculation of Solid Figures, USF: Understanding of Solid Figures, BTF: Basic Transformations of Figures, CCF: Cutting and Combining of Figures, Dir: Direction, Pos: Position, RoM: Route Map, CCP: Correspondence of Coordinates and Positions).}
\centering
\renewcommand{\arraystretch}{1.1} 

\resizebox{\textwidth}{!}{%
\begin{tabular}{c|ccc|cc|cc|cc|cc|cccc}

\toprule
\multirow{2}{*}{\textbf{Model}} &\multirow{2}{*}{\textbf{S1}} & \multirow{2}{*}{\textbf{S2}} &\multirow{2}{*}{\textbf{S3}} & \multicolumn{2}{|c|}{\textbf{Mem}} & \multicolumn{2}{|c|}{\textbf{PF}} & \multicolumn{2}{|c|}{\textbf{SF}} & \multicolumn{2}{|c|}{\centering{\textbf{TMF}}} & \multicolumn{4}{|c}{\textbf{PD}} \\
\cmidrule{5-16}

& & & & UCU & AL & CPF & UPF & CSF & USF & BTF & CCF & Dir & Pos & RoM & CCP \\
\midrule
\multicolumn{16}{c}{\textit{Closed-source}}\\ \midrule


GPT-4o & \high{72.84} & \high{58.06} & \high{43.64} & \high{86.61} & \high{39.12} & \high{77.35} & \high{71.56} & \high{84.50} & \high{62.27} & \high{58.74} & \high{69.37} & \high{93.10} & \high{72.67} & 47.53 & \high{73.33} \\
GPT-4V & 65.51 & 49.17 & 38.18 & 82.54 & 38.42 & 70.67 & 60.22 & 76.58 & 56.32 & 57.76 & 67.67 & 79.29 & 57.48 & \high{47.80} & 63.33 \\
Gemini 1.5 Pro & 56.13 & 51.39 & 33.94 & 50.99 & 31.23 & 61.75 & 45.03 & 69.95 & 57.54 & 39.24 & 62.65 & 68.81 & 54.13 & 40.66 & 60.00 \\
Qwen-VL-Max & 40.82 & 30.28 & 20.61 & 19.35 & 25.26 & 39.82 & 41.44 & 43.64 & 48.02 & 43.82 & 43.39 & 41.43 & 35.09 & 40.66 & 26.67 \\

\midrule
\multicolumn{16}{c}{\textit{Open-source}}\\ \midrule

LLaVA-NeXT-110B & \highg{53.74} & 36.94 & 31.52 & 39.48 & \highg{57.72} & \highg{59.48} & \highg{53.06} & \highg{52.25} & \highg{50.22} & \highg{54.09} & 50.76 & 54.76 & \highg{55.86} & 40.11 & 40.00 \\

LLaVA-NeXT-72B & 42.88 & 35.56 & 30.91 & 31.65 & 25.26 & 43.25 & 42.39 & 46.14 & 41.76 & 44.22 & \highg{51.02} & 44.29 & 38.93 & 32.97 & 36.67 \\

InternVL-Chat-V1.5 & 49.38 & 30.56 & 28.48 & 43.95 & 29.82 & 52.23 & 52.06 & 44.19 & 48.15 & 47.05 & 46.82 & \highg{65.71} & 50.47 & 36.54 & 36.67 \\

LLaVA-1.6-13B & 29.38 & 25.28 & 32.73 & 21.73 & 23.16 & 23.37 & 34.72 & 25.26 & 26.36 & 37.52 & 41.65 & 26.90 & 28.87 & 37.09 & 30.00 \\

G-LLaVA-13B & 32.43 & 30.56 & 32.73 & 33.33 & 29.12 & 32.04 & 37.88 & 19.57 & 33.51 & 37.12 & 32.79 & 31.19 & 33.21 & 25.55 & 40.00 \\

GLM-4V-9B & 47.33 & \highg{37.22} & \highg{38.18} & \highg{53.37} & 37.02 & 51.32 & 46.52 & 50.60 & 38.22 & 44.09 & 45.22 & 40.95 & 49.27 & 36.81 & \highg{53.33} \\

MiniCPM-LLaMA3-V 2.5 & 39.75 & 31.11 & 29.70 & 28.57 & 37.02 & 40.81 & 39.82 & 40.97 & 38.61 & 31.96 & 42.66 & 40.95 & 42.70 & 43.96 & 43.33 \\
LongVA-7B & 43.54 & 30.56 & 28.48 & 24.50 & 39.82 & 45.09 & 40.75 & 51.85 & 42.49 & 45.60 & 44.56 & 44.52 & 40.74 & 47.53 & 20.00 \\
LLaVA-1.6-7B & 22.96 & 20.83 & 15.76 & 18.45 & 20.53 & 16.92 & 29.63 & 15.57 & 18.60 & 42.67 & 24.05 & 17.62 & 43.31 & 28.85 & 26.67 \\
DeepSeek-VL-7B & 32.59 & 26.67 & 25.45 & 16.57 & 35.09 & 27.27 & 38.01 & 24.18 & 38.65 & 50.02 & 30.09 & 24.52 & 41.01 & 51.65 & 23.33 \\
InternLM-XComposer2-VL-7B & 47.00 & 33.06 & 33.33 & 31.25 & 46.49 & 47.70 & 42.57 & 51.44 & 43.87 & 41.13 & 50.58 & 65.48 & 53.87 & \highg{55.22} & 40.00 \\
Phi3-Vision-4.2B & 42.14 & 34.17 & 27.88 & 28.67 & 15.96 & 47.23 & 38.83 & 49.99 & 44.41 & 28.76 & 31.22 & 48.57 & 49.19 & 26.37 & 50.00 \\
DeepSeek-VL-1.3B & 31.44 & 27.78 & 23.03 & 27.78 & 23.86 & 22.76 & 36.92 & 30.36 & 34.18 & 44.46 & 28.29 & 48.10 & 41.77 & 37.09 & 33.33 \\

\bottomrule
\end{tabular}
}
\centering
\vspace{-0.2cm}
\label{tab:main_result}
\end{table}

\begin{figure}[!t]
    \centering
    \resizebox{0.98\textwidth}{!}{
    \includegraphics{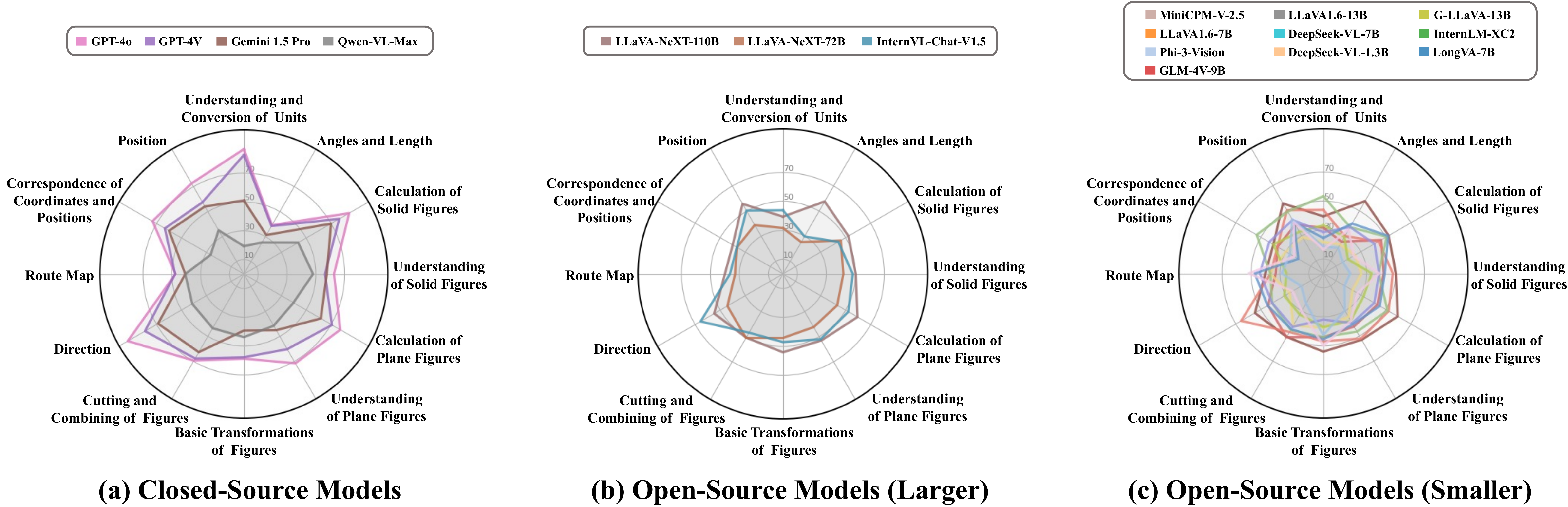}
    }
    \caption{The visualization of different LMMs' performances on each category.}
    \label{fig:leida-main}
\end{figure}

Table \ref{tab:main_result} shows the overall performance of different LMMs on One-Step / Two-Step / Three-Step problems and different problem domains. We have the following observations:

\textbf{The Nums of Knowledge Concepts are negatively correlated with LMMs' Performance.} Regarding problems of varying complexities (one-step vs. two-step vs. three-step), GPT-4o consistently achieve an advantage across all settings. Other closed-source models, such as GPT-4V and Gemini 1.5 Pro, also demonstrate competitive performance. However, most LMMs perform significantly worse on multi-step problems compared to one-step problems. For instance, GPT-4o's accuracy drops from 72.84\% to 43.64\%. This trend is even more pronounced in stronger models like LLaVA-NeXT-110B and InternVL-Chat-V1.5. These observations suggest that the number of knowledge concepts in a question is positively correlated with its difficulty and negatively correlated with LMMs' performance, supporting the rationale for decomposing questions to a certain extent.

\textbf{Larger Parameter Scales in LLMs generally achieve Better Generalization Abilites}. 
To explore what role LLM plays in LMMs, we conduct pairwise comparisons on the LMMs with the same LLM backbone (e.g. LLaVA-NeXT-110B vs LLaVA-NeXT-72B; DeepSeek-VL-7B vs DeepSeek-VL-1.3B). Focusing on the strict metric, we observe that larger parameter scales in LLMs generally perform better, which reveals that the parameter scales in the text decoder is a key factor in achieving the generalization ability in visual mathematical reasoning.

\textbf{LMMs excel in Calculation but struggle with Fine-grained Visual Measurement.} 
Focusing on different math categories, GPT-4o still maintains impressive results across various subfields. In contrast, as shown in Figure \ref{fig:leida-main}, other LMMs generally struggle with "Angle Measurement" and "Unit Conversion". After analyzing these cases, we reveals that the main challenge for LMMs lies in their inability to perform precise visual angle and unit measurements. Furthermore, most LMMs demonstrate better proficiency in calculation (e.g., \textit{Calculations of Solid Figures, Calculations of Plane Figures}) compared to conceptual understanding (e.g., \textit{Understanding of Solid Figures, Understanding of Plane Figures}), which indicates that current LMMs excel at directly applying formulas based on given conditions on but are still limited in understanding and comprehensively applying knowledge.

\begin{table}[!t]
\centering
\caption{The performance of different LMMs on four-dimensional metrics for reasoning evaluation. The best and the two worst performances are marked in \colorbox{backblue}{blue} and \colorbox{backred}{red} (Avg: $\text{Score}_{\text{average}}$).}
\vspace{-0.3cm}

\renewcommand{\arraystretch}{1.4} 
\resizebox{\textwidth}{!}{%
\begin{tabular}{c|c|c|c|c|c|c|c|c|c|c}

\toprule
\multirow{2}{*}{\textbf{Model}} & \multicolumn{5}{|c|}{\textbf{Strict}} & \multicolumn{5}{|c}{\textbf{Loose}} \\
\cmidrule{2-11}
 & \textbf{Avg} ($\uparrow$) & \textbf{IK} ($\downarrow$) & \textbf{IG} ($\downarrow$) & \textbf{CM} ($\uparrow$) & \textbf{RM} ($\downarrow$) & \textbf{Avg} ($\uparrow$) & \textbf{IK} ($\downarrow$) & \textbf{IG} ($\downarrow$) & \textbf{CM} ($\uparrow$) & \textbf{RM} ($\downarrow$) \\ 
\midrule
\multicolumn{11}{c}{\textit{Closed-source}}\\ \midrule
GPT-4o & \high{42.86\%} & \high{31.24\% (164)} & \highr{15.24\% (80)} & \high{35.24\% (185)} & \high{34.16\% (96)} & \high{60.57\%} & \high{31.24\% (164)} & \highr{15.24\% (80)} & \high{52.95\% (278)} & \high{1.07\% (3)} \\

GPT-4V & 31.05\% & 39.81\% (209) & \highr{14.48\% (76)} & 23.81\% (125) & 47.92\% (115) & 51.43\% & 39.81\% (209) & \highr{14.48\% (76)} & 44.19\% (232) & 3.33\% (8) \\
        
Gemini 1.5 Pro & 26.38\% & 42.86\% (225) & 11.24\% (59) & 20.76\% (109) & 54.77\% (132) & 46.00\% & 42.86\% (225) & 11.24\% (59) & 40.38\% (212) & 12.03\% (29) \\


Qwen-VL-Max & 10.48\% & 65.14\% (342) & 7.62\% (40) & 6.67\% (35) & 75.52\% (108) & 25.52\% & 65.14\% (342) & 7.62\% (40) & 21.71\% (114) & 20.28\% (29) \\ \midrule

\multicolumn{11}{c}{\textit{Open-source}}\\ \midrule

LLaVA-NeXT-110B & 19.24\% & 50.29\% (264) & \highr{14.48\% (76)} & 12.00\% (63) & 65.95\% (122) & 37.90\% & 50.29\% (264) & \highr{14.48\% (76)} & 30.67\% (161) & 12.97\% (24) \\

LLaVA-NeXT-72B & 13.43\% & 58.86\% (309) & 7.05\% (37) & 9.90\% (52) & 70.95\% (127) & 31.52\% & 58.86\% (309) & 7.05\% (37) & 28.00\% (147) & 17.88\% (32) \\

InternVL-Chat-V1.5 & 14.95\% & 56.19\% (295) & 13.90\% (73) & 8.00\% (42) & 73.25\% (115) & 32.67\% & 56.19\% (295) & 13.90\% (73) & 25.71\% (135) & 14.01\% (22) \\

LLaVA-1.6-13B & \highr{5.24\%} & 69.14\% (363) & 3.24\% (17) & \highr{3.62\% (19)} & \highr{86.90\% (126)} & 22.00\% & 69.14\% (363) & 3.24\% (17) & 20.38\% (107) & 26.21\% (38) \\

G-LLaVA-13B & 6.48\% & 64.19\% (337) & 4.57\% (24) & 4.19\% (22) & 86.59\% (142) & 22.29\% & 64.19\% (337) & 4.57\% (24) & 20.00\% (105) & \highr{35.98\% (59)} \\

GLM-4V-9B & 14.86\% & 52.95\% (278) & 9.52\% (50) & 10.10\% (53) & 73.10\% (144) & 35.05\% & 52.95\% (278) & 9.52\% (50) & 30.29\% (159) & 19.29\% (38) \\

MiniCPM-LLaMA3-V 2.5 & 9.52\% & 60.19\% (316) & 9.14\% (48) & 4.95\% (26) & 83.85\% (135) & 28.00\% & 60.19\% (316) & 9.14\% (48) & 23.43\% (123) & 23.60\% (38) \\

LongVA-7B & 11.52\% & 61.14\% (321) & 8.95\% (47) & 7.05\% (37) & 76.43\% (120) & 27.71\% & 61.14\% (321) & 8.95\% (47) & 23.24\% (122) & 22.29\% (35) \\

LLaVA-1.6-7B & \highr{3.33\%} & \highr{78.29\% (411)} & \high{2.48\% (13)} & \highr{2.10\% (11)} & \highr{89.11\% (90)} & \highr{13.81\%} & \highr{78.29\% (411)} & \high{2.48\% (13)} & \highr{12.57\% (66)} & \highr{34.65\% (35)} \\

DeepSeek-VL-7B & 6.29\% & 69.14\% (363) & 4.57\% (24) & 4.00\% (21) & 84.78\% (117) & \highr{20.95\%} & 69.14\% (363) & 4.57\% (24) & \highr{18.67\% (98)} & 28.99\% (40) \\

InternLM-XComposer2-VL-7B & 12.67\% & 56.38\% (296) & 10.48\% (55) & 7.43\% (39) & 77.59\% (135) & 30.95\% & 56.38\% (296) & 10.48\% (55) & 25.71\% (135) & 22.41\% (39) \\

Phi3-Vision-4.2B & 10.57\% & 58.86\% (309) & 8.95\% (47) & 6.10\% (32) & 81.07\% (137) & 29.81\% & 58.86\% (309) & 8.95\% (47) & 25.33\% (133) & 21.30\% (36) \\

DeepSeek-VL-1.3B & 5.90\% & \highr{71.05\% (373)} & 2.67\% (14) & 4.57\% (24) & 82.61\% (114) & 21.52\% & \highr{71.05\% (373)} & 2.67\% (14) & 20.19\% (106) & 23.19\% (32) \\

\bottomrule

\end{tabular}
}
\label{tab:four_dimention_result}
\end{table}

\begin{figure}[!t]
    \centering
    \resizebox{\textwidth}{!}{
    \includegraphics{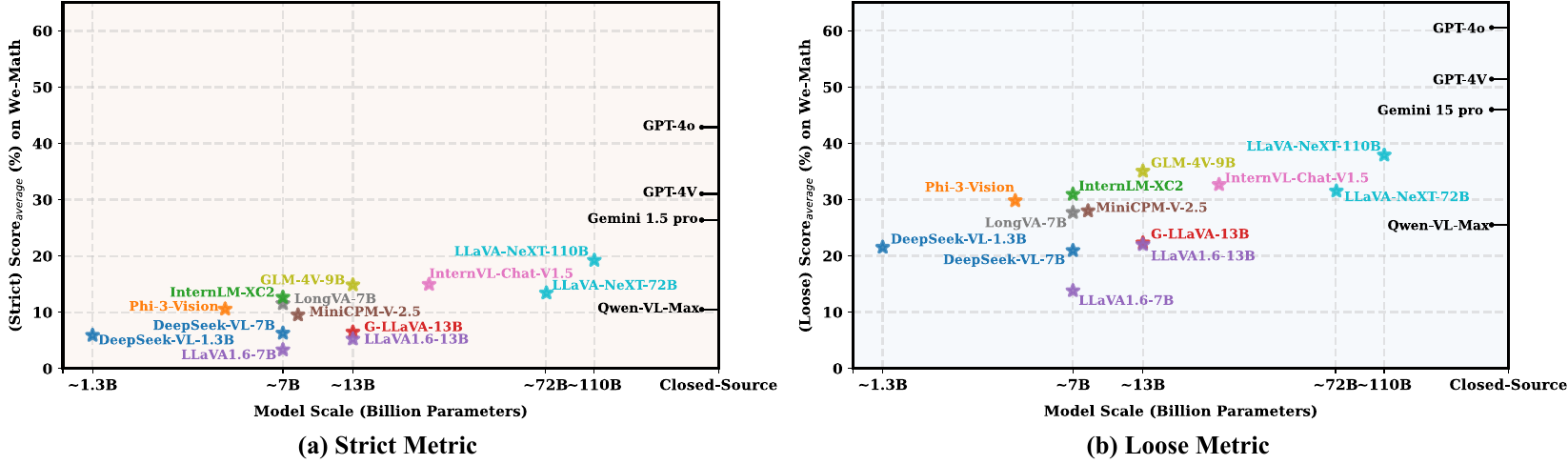}
    }
    \vspace{-0.3cm}
    \caption{The Leaderboard of different LMMs under the strict and loose metric (average score \%). "$\sim$" represents an approximate estimate of the total parameters nums in LMMs.}
    \label{fig:619-leaderboard}
\end{figure}

\textbf{LMMs exhibit Strong Potential for Parameter Compression.} In terms of different LMMs, LLaVA-NeXT-110B demonstrates performance closest to GPT-4. Surprisingly, despite having smaller parameter scales, Phi3-Vision-4.2B and MiniCPM-Llama3-V 2.5 also show competitive performance compared to LLaVA-NeXT-72B. Moreover, the recent GLM-4V-9B and InternVL-Chat-V1.5 have allocated a larger proportion of parameters to the visual encoder (as shown in Table \ref{tab:LMMs Architecture}), thereby demonstrating notable capabilities. This underscores the importance of optimizing visual representations and suggests that LMMs still have significant potential for parameter compression.significant potential for parameter compression.


\subsection{Knowledge based Reasoning Analysis }
\label{exp:reason}

\begin{figure}[!t]
    \centering
    \resizebox{0.8\textwidth}{!}{
    \includegraphics{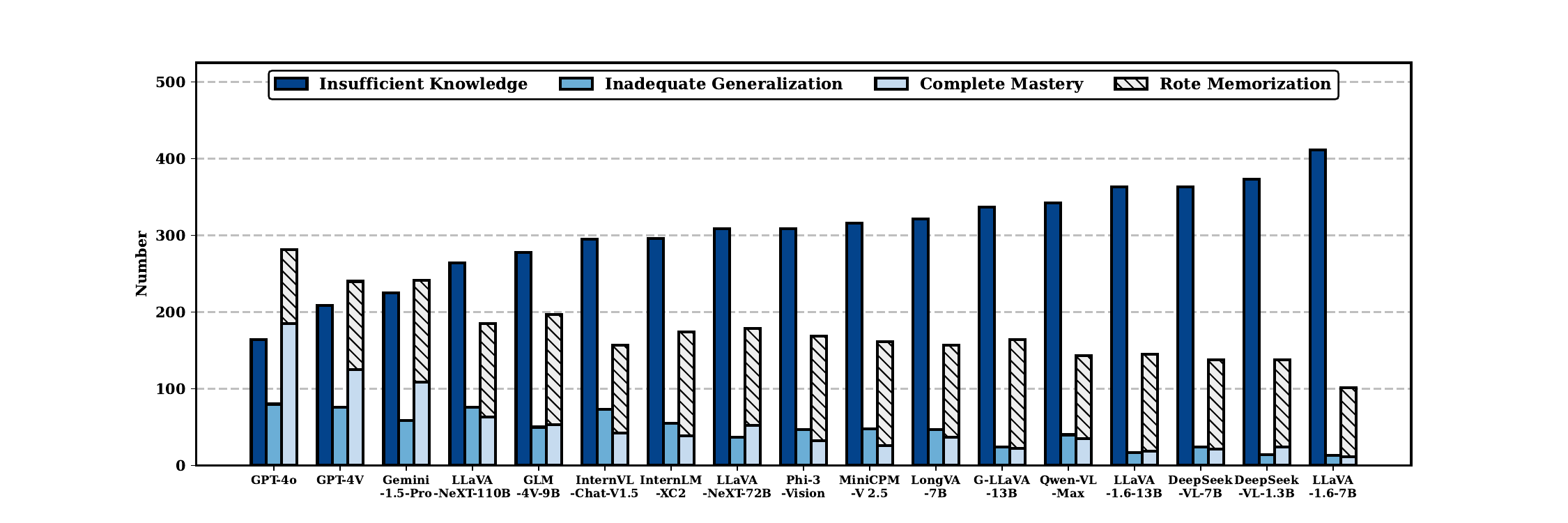}
    }
    \caption{The performance of different LMMs on four-dimensional metrics under strict metric.}
    \label{fig:main_strict_4metric}
\end{figure}

\begin{figure}[!t]
    \centering
    \resizebox{0.8\textwidth}{!}{
    \includegraphics{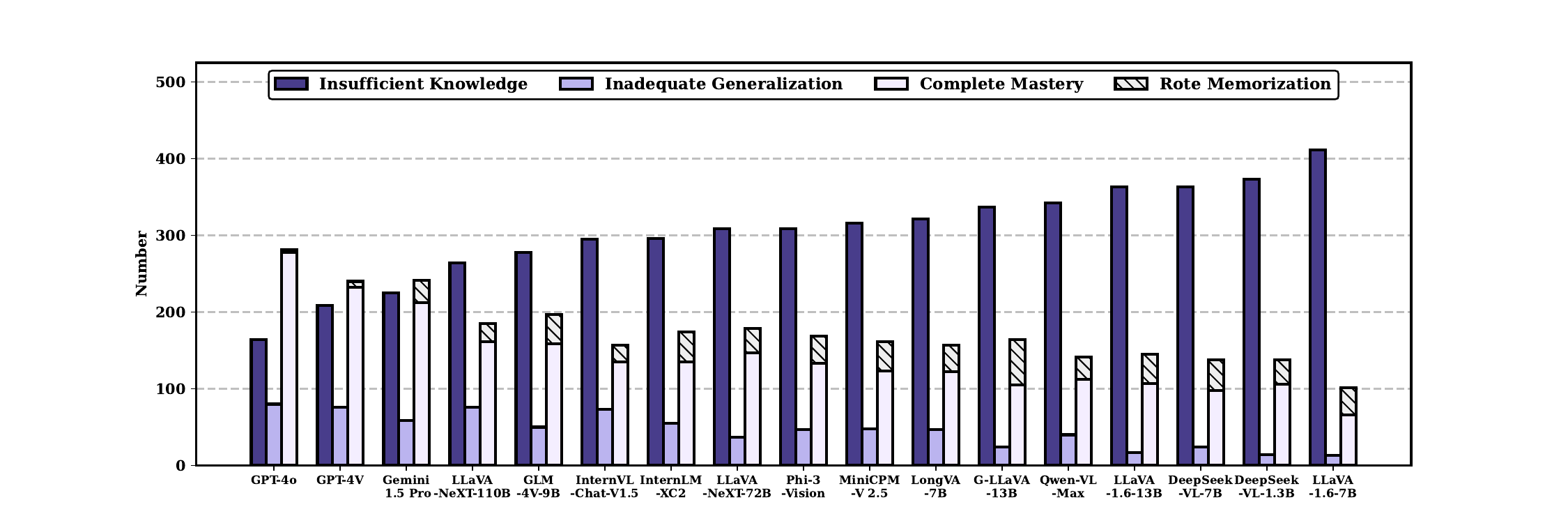}
    }
    \caption{The performance of different LMMs on four-dimensional metrics under loose metric.}
    \label{fig:main_loose_4metric}
\end{figure}

Table \ref{tab:four_dimention_result} and Figure~\ref{fig:619-leaderboard},~\ref{fig:main_strict_4metric},~\ref{fig:main_loose_4metric} illustrate the results of knowledge based reasoning evaluation, including four distinct conditions (\textit{IK, IG, CM, RM}). We have the following observations:


\textbf{IK is the Greatest Vulnerability of LMMs.} All LMMs consistently demonstrate an \textit{Insufficient Knowledge} issue during the reasoning process, especially in models with smaller parameter scales (LLaVA-1.6-7B, DeepSeek-VL-1.3B). As discussed in section \ref{sec:decompose}, addressing \textit{IK} is crucial for progressing towards \textit{Inadequate Generalization (IG)} and \textit{Complete Mastery (CM)}. This knowledge gap in solving one-step problems hinders further progress in reasoning about more composite mathematical problems. This finding also supports the rationale behind our proposed KCA strategy.


\label{exp:ka}
\begin{figure}[!t]
    \centering
    \resizebox{\textwidth}{!}{
    \includegraphics{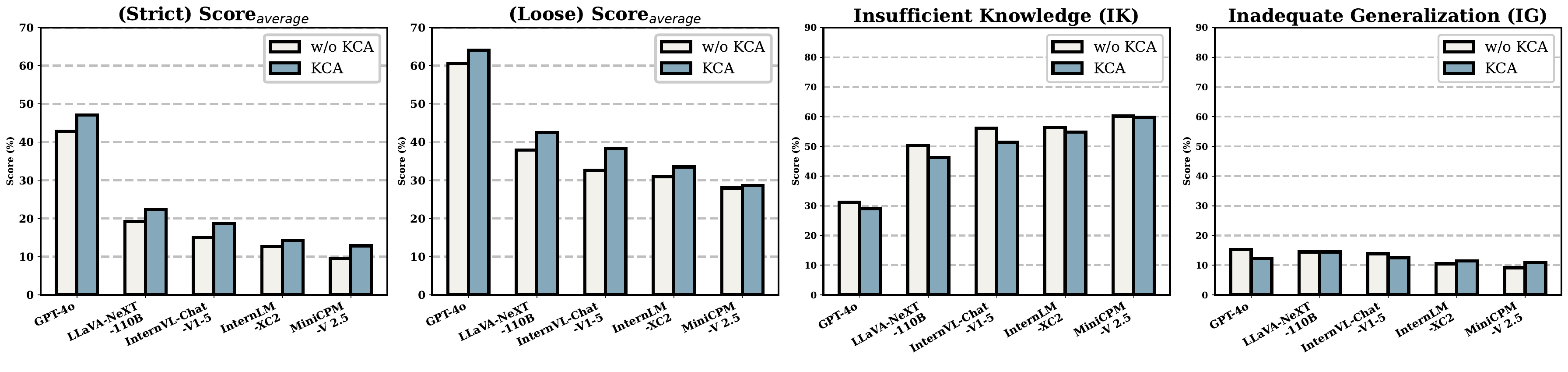}
    }
    \vspace{-0.5cm}
    \caption{Quantitative Analysis on KCA. The left two figures show the impact of KCA on the average performance of LMMs under strict and loose conditions. The right two figures compare the results between \textit{IK} and \textit{IG}.}
    \label{fig:aug}
\end{figure}

\textbf{GPT-4o's Main Challenge has gradually shifted from \textit{IK} to \textit{IG}, highlighting it as the First LMM towards the  Knowledge Generalization Stage.} Focusing on \textit{IK} and \textit{IG}, GPT-4o exhibits a substantial lead in addressing the \textit{IK} issue, but the weakest performance in \textit{IG}. Further analyzing the logical relationships between \textit{IK}, \textit{IG}, and \textit{CM} (\textit{IK} $\rightarrow$ \textit{IG} $\rightarrow$ \textit{CM}), we are  pleasantly surprised to find that GPT-4o is markedly superior to the open-sourced LLaVA-NeXT-110B in \textit{IK} (19.05\%), suggesting it has successfully converted a considerable amount of \textit{IK} into \textit{IG} issue. This revelation indicates that GPT-4o's challenges in reasoning have shifted from addressing \textit{Insufficient Knowledge} in one-step problems to the knowledge generalization stage, leading us to speculate that there may have been groundbreaking changes in GPT-4o's training strategy. However, other LMMs remain stuck at the \textit{IK} phase. We argue that it is pointless to compare \textit{IG} without a solid grasp of \textit{IK}, highlighting the significance of our hierarchical metrics (\textit{IK} < \textit{IG} < \textit{CM}).




\textbf{The Unreasonable RM issue remains widespread across Most LMMs.} GPT-4o achieves a significant lead on the \textit{RM} issue, particularly on the loose metric ($S_{RM} < 2\%$). However, other LMMs still exhibit nearly 25\% $S_{RM}$ on the loose metric. When focusing on the changes in $S_{RM}$ between strict and loose metrics, several models (LLaVA-NeXT-110B, GLM-4V-9B, DeepSeek-VL-1.3B, MiniCPM-Llama3-V 2.5) show significant variations. This is a beneficial phenomenon, indicating that these models possess a certain ability to solve one-step problems, but their performance fluctuates due to external factors such as prompting templates and hyper-parameters.

\begin{wrapfigure}{r}{0.48\textwidth}
    \centering

\vspace{-0.2cm}

\includegraphics[width=0.47\textwidth]{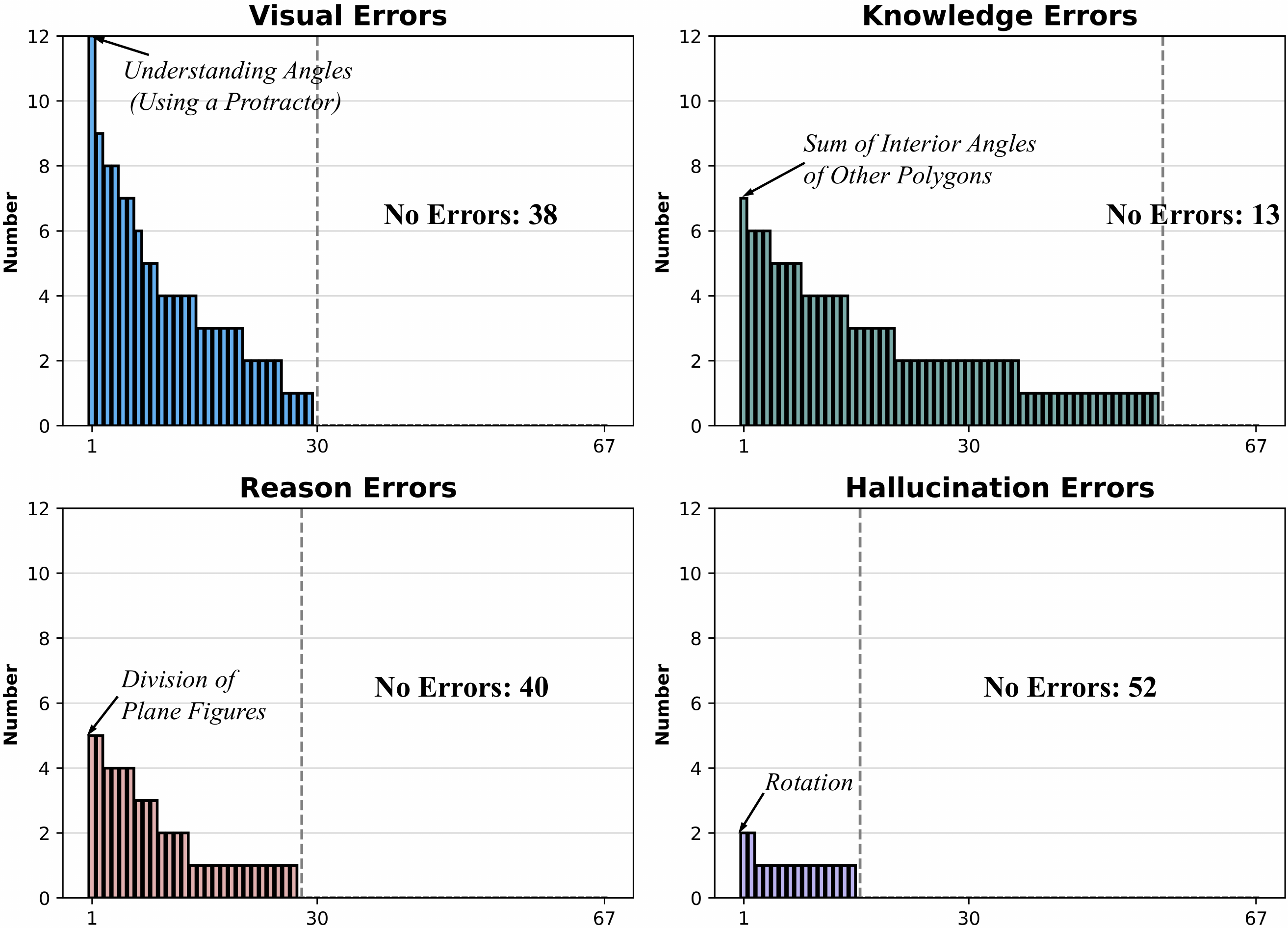} 
\vspace{-0.3cm}
\caption{Error analysis of GPT-4o, The definitions of 4 types errors are listed in Appendix.}
\vspace{-0.3cm}
    \label{fig:error}
\end{wrapfigure}

\subsection{Quantitative Analysis}
\textbf{The Effectiveness on KCA.} Figure \ref{fig:aug} displays the quantitative analysis of the LMMs with knowledge concept augmented (KCA). We find that LMMs with different parameter scales show consistent performance improvements on both strict and loose metrics after introducing the KCA strategy. 
Moreover, the KCA strategy significantly alleviates \textit{IK} issues but does not noticeably improve \textit{IG}. This aligns with human intuition, as the knowledge descriptions primarily address gaps in reasoning knowledge. Nevertheless, alleviating \textit{IG} issues requires a comprehensive enhancement of the LMMs' knowledge generalization abilities, which we consider a direction for future exploration.






\textbf{Error Anaysis.}
\label{exp:error}
Figure \ref{fig:error} shows the occurrence of the four types of errors across the 67 knowledge concepts. Knowledge errors are the most frequent, appearing in over 45 knowledge concepts. Notably, although visual errors are the second most common, they are more concentrated in specific concepts (e.g., \textit{"Understanding Angles"} >10), and over 38 concepts have no visual errors. This finding underscores the urgent need to enhance the fine-grained measurement capabilities of visual encoders in LMMs for mathematical reasoning, rather than blindly improving their overall capabilities.



\section{Related Work}
\textbf{Mathematical Reasoning Benchmarks.} Assessing mathematical reasoning abilities is crucial for the development of large foundational models (LLMs and LMMs). Early efforts, such as MathQA~\citep{amini-etal-2019-mathqa}, focus on solving mathematical word problems and highlight the importance of operation-based reasoning. Following this, datasets like GSM8K~\cite{cobbe2021training} and MATH~\cite{hendrycks2021measuring} set the stage for evaluating text-based mathematical problems at various difficulty levels. Other benchmarks, such as MMLU~\citep{hendrycks2021measuringmassivemultitasklanguage} and MT-Bench~\citep{zheng2023judgingllmasajudgemtbenchchatbot}, also consider mathematical evaluation as a key part of assessing LLMs. Beyond text-only evaluations, datasets like GeoQA~\cite{chen2021geoqa}, UniGeo~\cite{chen2022unigeo}, and Geometry3K~\cite{lu2021inter} have pioneered the evaluation of geometric problems. Recently, several benchmarks~\cite{lu2023mathvista}~\cite{yue2023mmmu} have expanded their scope to cover a broader range of subjects. Additionally, MathVerse~\cite{zhang2024mathverse} aims to evaluate reasoning paths based on reference answers. However, challenges remain due to the complex nature of mathematical reasoning. In this paper, we introduce \dataset, a comprehensive benchmark designed to evaluate the reasoning abilities of LMMs across a wide range of mathematical categories.

\textbf{Benchmarks for Large Multimodal Model.} The rapid advancement of Large Language Models (LLMs) and Large Multimodal Models (LMMs) have highlighted the necessity for more comprehensive evaluation benchmarks. 
At first, the emergence of a series of text-only benchmarks and evaluations give us a clearer understanding of the strengths and weaknesses of large language models~\citep{huang2023ceval,hendrycks2021measuringmassivemultitasklanguage,zheng2023judgingllmasajudgemtbenchchatbot,dong2023revisitinputperturbationproblems,song2024csbenchcomprehensivebenchmarklarge,song2023largelanguagemodelsmeet,dong2024understandllmneedsdual,xue2023occuquestmitigatingoccupationalbias,zellers2019hellaswagmachinereallyfinish,cobbe2021training,hendrycks2021measuring,sarlin2020supergluelearningfeaturematching}. Focusing on the visual aspect, early benchmarks predominantly focused on narrow tasks like Visual Question Answering (VQA)~\citep{antol2015vqa,goyal2017making,kafle2017analysis,singh2019towards,hudson2019gqa,docvqa,okvqa,stvqa,mmvet,ocrvqa,mme,vizwiz,sciencqa} and image captioning~\citep{lin2014microsoft,agrawal2019nocaps,plummer2015flickr30k}, showcasing significant progress but not fully addressing the broader spectrum of multimodal perception and reasoning. This gap has driven recent research to assess LMMs from multiple angles. Notable efforts include MMBench~\citep{liu2023mmbench} and SEED-bench~\citep{li2023seed,li2024seed}, which probe models' abilities through common-sense queries incorporating multiple-choice formats. For domain-specific expertise, MMMU~\citep{yue2023mmmu} utilize academic content to gauge deeper knowledge levels. Yet, benchmark such as MMStar~\citep{chen2024we} reveals that certain evaluations allow models to respond without images, risking data leakage and failing to adequately measure logic and reasoning skills. The challenge of understanding image implications, requiring multi-hop reasoning and theory of mind (ToM)~\citep{desai2022nice,hessel2022androids,street2024llms,street2024llms,dong2024selfplayexecutionfeedbackimproving,liu2024ii}, underscores this shortfall. In parallel, the intersection of large language models (LLMs) and Large Multimodal Models (LMMs) has surged, extending the applicability of LMMs evaluations across diverse modalities including 2D images~\citep{zhang2024imagenetdbenchmarkingneuralnetwork,deng2009imagenet,everingham2010pascal}, 3D point clouds~\citep{wu20153d,chang2015shapenet,geiger2013vision}, audio~\citep{panayotov2015librispeech,veaux2017cstr,yang2024airbenchbenchmarkinglargeaudiolanguage,wang2024audiobenchuniversalbenchmarkaudio}, and video~\citep{ning2023videobenchcomprehensivebenchmarktoolkit,abu2016youtube,huang2023vbenchcomprehensivebenchmarksuite}. Moreover, a series of works have positioned LMMs as agents with various tools, such as APIs~\citep{xie2024large,wang2024mllmtoolmultimodallargelanguage,liu2023textunveilingmultimodalproficiency}, retrievers~\citep{long2024generativemultimodalknowledgeretrieval,zhao-etal-2023-retrieving} , thereby broadening the development avenues for the model evaluation community~\citep{liang2021multibenchmultiscalebenchmarksmultimodal,ge2024mllmbenchevaluatingmultimodalllms,ying2024mmtbenchcomprehensivemultimodalbenchmark,wei2023uniirtrainingbenchmarkinguniversal}.

\section{Conclusion}

In this paper, we propose \dataset, a comprehensive benchmark for in-depth analysis of LMMs in visual mathematical reasoning. \dataset encompasses 6.5K visual math problems, covering 5 layers and 67 knowledge concepts. Moreover, we pioneeringly decompose composite problems into sub-problems according to the required knowledge concepts and introduce a novel four-dimensional metric for fine-grained reasoning evaluation. With \dataset, we thoroughly evaluate existing LMMs in visual mathematical reasoning and reveal a negative correlation between solving steps and problem-specific performance. Furthermore, we identify \textit{IK} issues as the greatest vulnerability of LMMs. However, GPT-4o's main challenge has shifted from \textit{IK} to \textit{IG}, highlighting it the first LMM towards the next stage. Lastly, analyses on KCA strategy and error cases further heuristically guides existing LMMs towards human-like visual mathematical reasoning.

\bibliography{references}
\bibliographystyle{unsrt}

\clearpage
\appendix
\begin{center}
{\Large \textbf{Appendix}}
\end{center}

\setcounter{section}{0}
\renewcommand{\thesection}{\Alph{section}}
\tableofcontents
\clearpage

\section{Broaden Impact}
\label{app:Broaden Impact}

\textbf{Bridging Human-Like Inspiration and Reliability.} As previously mentioned, works such as neural networks~\cite{lecun1998gradient} and attention mechanisms~\cite{vaswani2017attention} draw their design inspiration from human thinking patterns. This is fundamentally because the purpose of designing AI is to assist humans. Currently, LMMs have already been helping people in various scenarios, which was unimaginable in the past. Therefore, we firmly believe that a new era is coming, where people will focus not only on the performance of models in specific fields but also on the reliability of a model. In some fundamental scenarios, a reliable model is more important, which is one of the primary motivations behind the creation of \dataset. Furthermore, after completing our experiments, we find that in a loose setting, GPT-4o's \textit{RM} metric is only 1.07\%, showing us the possibility of a reliable and accurate model emerging in the future.

\textbf{Fine-grained Evaluation and Versatile Applications.} From the model's perspective, \dataset can provide LMMs with an assessment of mathematical abilities. Additionally, \dataset's \textit{IK}, \textit{IG}, and \textit{CM} metrics offer a fine-grained evaluation of the model's capabilities. Furthermore, the \textit{RM} metric reflects a model's reliability to address our concern of not desiring a model that can solve complex problems but makes errors on sub-problems within the solution process. Ultimately, we introduce the $\text{Score}_{\text{average}}$ metric to quantify the model's overall performance.
Moreover, since \dataset is constructed from the decomposition of a multi-step problem's necessary solution process, it provides new perspectives for interactive tasks (multi-turn dialogues), self-supervised learning, information extraction, and other tasks. It also offers crucial references and support for the deployment of models in education and other fields.

\textbf{Ethics Statement.} We ensure that \dataset complies with legal and ethical guidelines throughout its construction process, with no violations. We provide fair compensation to all annotators involved. \dataset focuses on elementary mathematics problems, and during its construction, data collection was sourced from publicly available test questions, textbooks, and professional websites. Since mathematics problems inherently have standard answers, they are not subject to cultural differences. Additionally, we guarantee that \dataset is solely for academic research purposes, and we uphold the strict prohibition of any commercial use. Additionally, we declare that we will bear full responsibility in the event of any rights violations and confirm the data license.

\section{More Details on \dataset}
\label{sec:dataset}

\subsection{Hierarchical Knowledge Structure}
\label{sec:Hierarchical}

\begin{figure}[ht]
    \centering
    \resizebox{\textwidth}{!}{
    \includegraphics{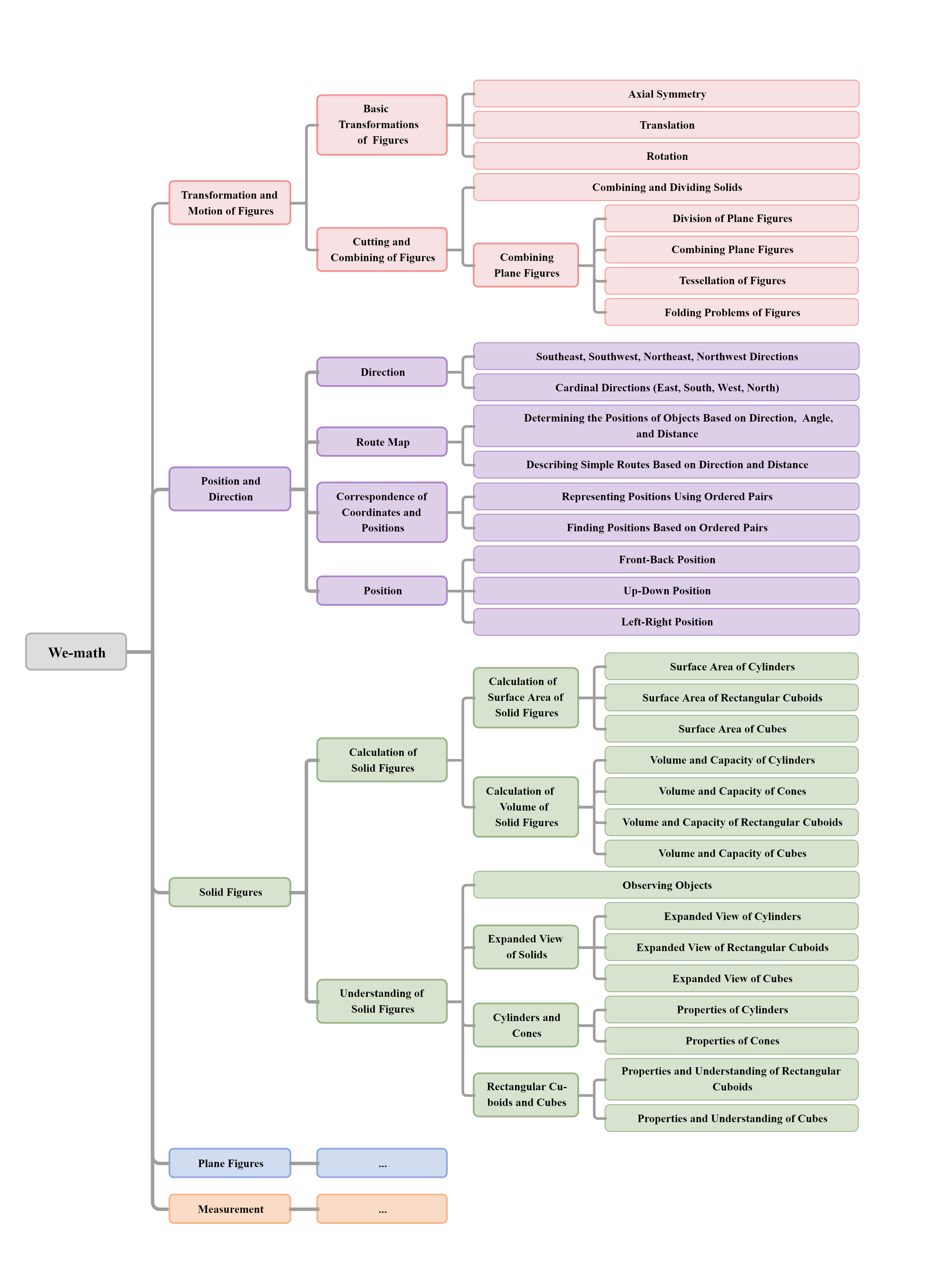}
    }
    \caption{The Hierarchical Knowledge Structure of \dataset(1).}
    \label{fig:tree1}
\end{figure}

\begin{figure}[ht]
    \centering
    \resizebox{\textwidth}{!}{
    \includegraphics{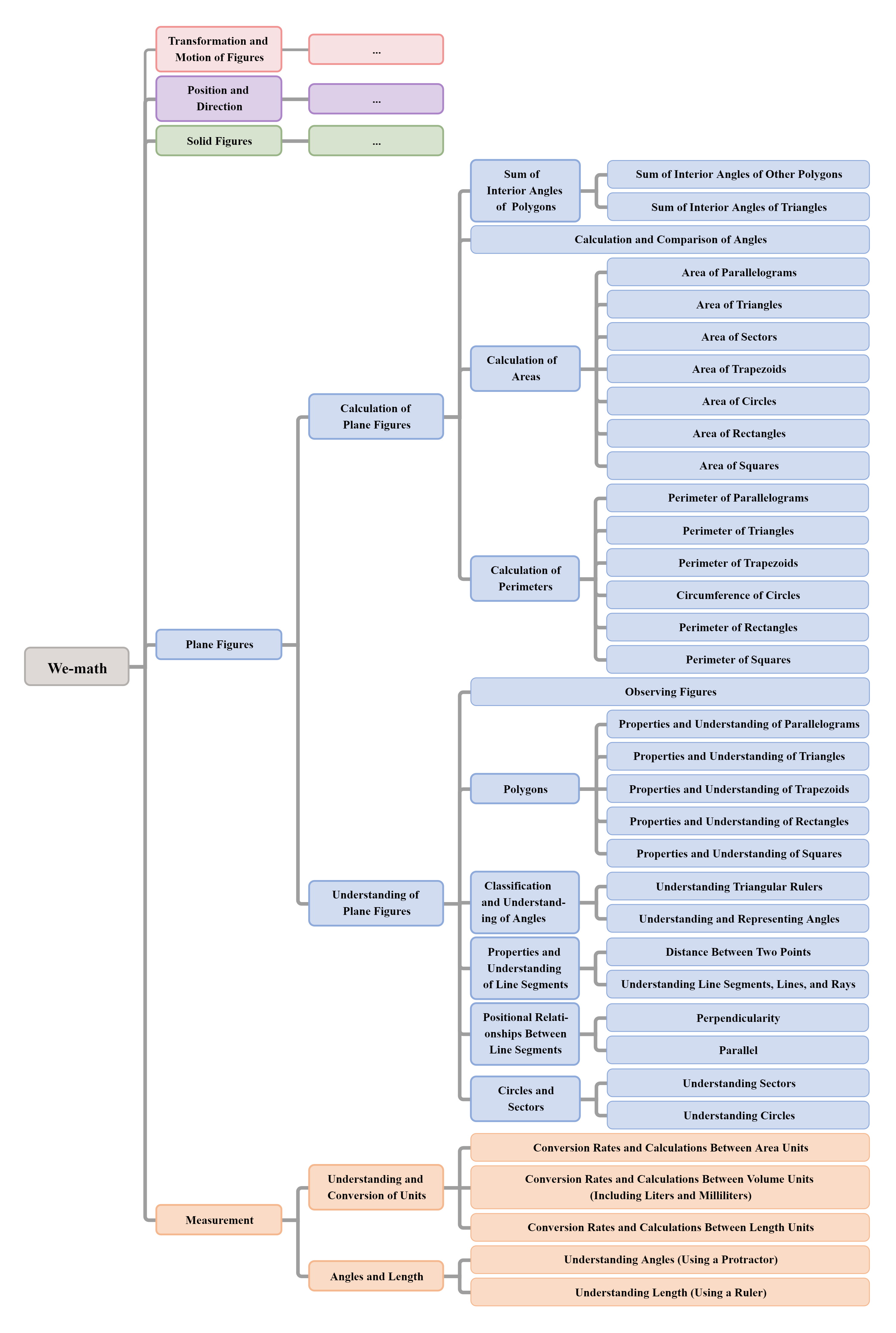}
    }
    \caption{The Hierarchical Knowledge Structure of \dataset(2).}
    \label{fig:tree2}
\end{figure}

Figure~\ref{fig:tree1},~\ref{fig:tree2} shows the detailed hierarchical structure of \dataset, which includes 5 levels, 99 nodes, and 67 leaf nodes.

In the initial stages of constructing the benchmark, we aimed to address two key objectives. We believe that the purpose of designing a benchmark is to evaluate the performance of models and provide guidance on areas that need improvement. However, existing benchmarks offer only broad guides in these aspects. Additionally, the core contribution mentioned earlier is that \dataset is the first benchmark specifically designed to study the mathematical problem-solving mechanisms of models. Inspired by the learning paradigm of humans, which is based on knowledge concepts, \dataset constructs its dataset with knowledge concepts as the basic unit, resulting in evaluations with rigorous scientific accuracy and better guidance.


\subsection{Knowledge-based Data Decomposition}
\label{subapp:Decomposition}

\begin{figure}[ht]
    \centering
    \resizebox{0.9\textwidth}{!}{
    \includegraphics{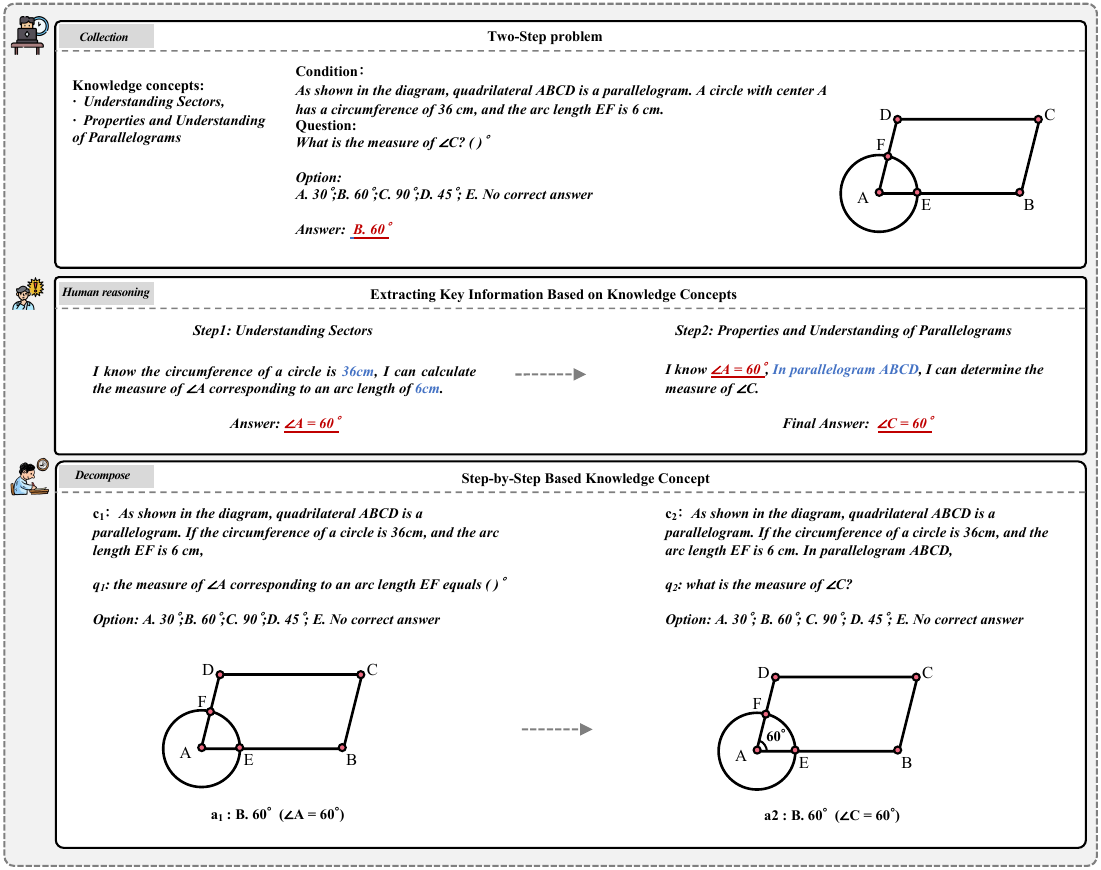}
    }
    \vspace{-0.2cm}
    \caption{An example of a two-step problem in \dataset.}
    \label{fig:example1-2}
\end{figure}

\begin{figure}[ht]
    \centering
    \resizebox{0.9\textwidth}{!}{
    \includegraphics{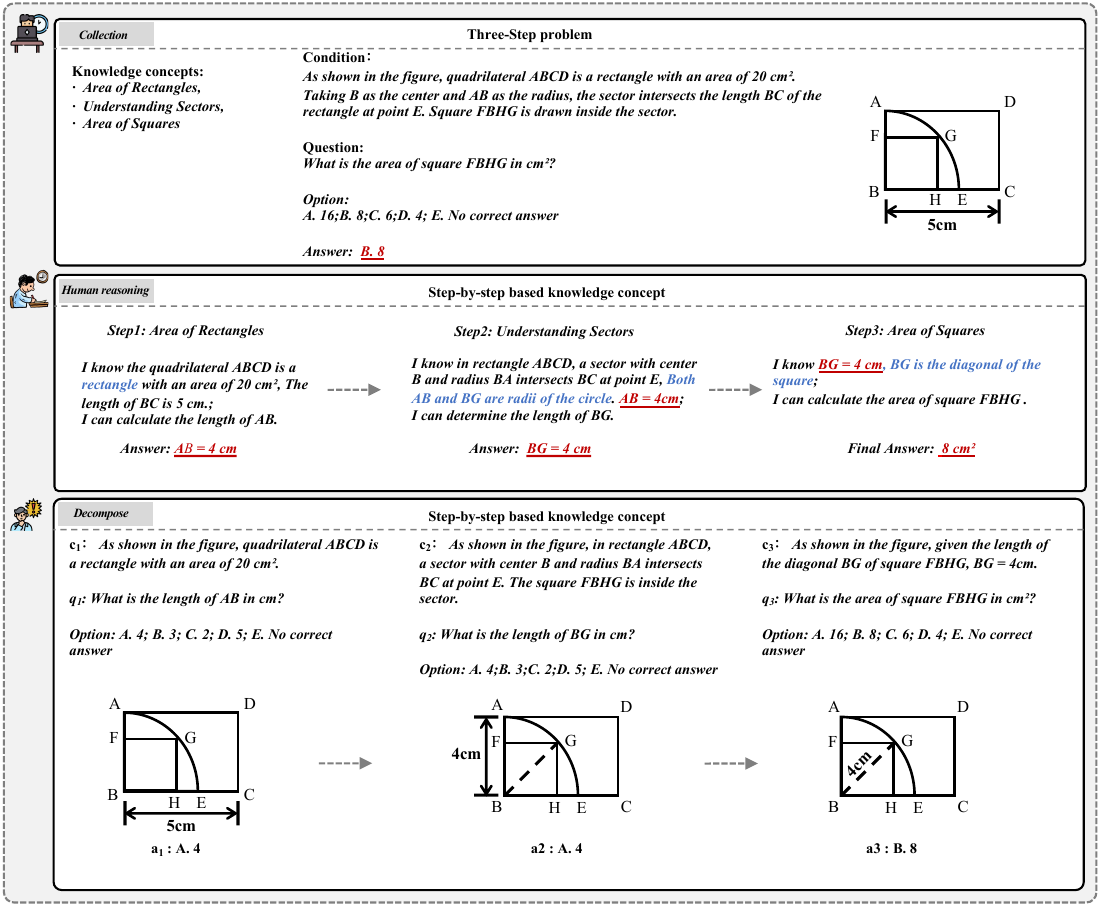}
    }
      \vspace{-0.2cm}
    \caption{An example of a three-step problem in \dataset.}
    \label{fig:example2-3}
\end{figure}

Figures~\ref{fig:example1-2},~\ref{fig:example2-3} illustrate the process of Knowledge-based Data Decomposition.

\textbf{Collection.} In each example, the Collection section presents specific information about each multi-step problem in the dataset.

\textbf{Human reasoning.} The Human reasoning section shows the process required before decomposing each multi-step problem, where educational experts extract the key information needed for each sub-problem based on the reasoning path for the knowledge concepts included in the multi-step problem. 

\textbf{Decompose.} The Decompose section uses the key information extracted in the Human reasoning section to formulate sub-problems, refine the options, and ultimately achieve the decomposition of a multi-step problem.

It is necessary to further explain that to ensure each sub-problem has a rigorous logical relationship and is independent, the text condition for the first sub-problem is derived from the text condition of the multi-step problem, and the image condition for the first sub-problem is the same as the image condition of the multi-step problem.

Furthermore, in constructing the second sub-problem, two situations may arise. The first situation is where the answer of the first sub-problem is injected as a key condition into the image condition of the second sub-problem, presenting the information visually. The second situation is where the answer of the first sub-problem is injected as a key condition into the text condition of the second sub-problem, while the image condition remains unchanged.

In \dataset, the vast majority of cases are of the first type. However, for some information that is extremely difficult to present in images, we opt for the second type, presenting the information in text form. To ensure fairness in the decomposition of the problems, only one of these situations will occur in the decomposition of the same multi-step problem. This approach ensures that the question of the final sub-problem will match the original multi-step problem, completing the decomposition.

\vspace{-0.2cm}

\clearpage






\begin{table}[!t]
  \centering
    \caption{Prompt templates for response generations.}
    \footnotesize
\begin{tabular}{c|l}
\toprule
\multicolumn{1}{l|}{\textbf{Type}}     & \multicolumn{1}{c}{\textbf{Prompt Template}}\\ 
\midrule
\begin{tabular}[c]{@{}c@{}}Multiple\\Choice\end{tabular}   & \begin{tabular}[c]{@{}l@{}}Now, we require you to solve a multiple-choice math question. Please briefly \\ describe your thought process and provide the final answer(option).\\ \textbf{Question}: \textless{}Question\textgreater\\
\textbf{Option}: \textless{}Option\textgreater\\
Regarding the format, please answer following the template below, and be \\ sure to include two \textless{}\textgreater{} symbols: \\
\textbf{\textless{}Thought process\textgreater{}}: \textless{}\textless{}your thought process\textgreater{}\textgreater{} \textbf{\textless{}Answer\textgreater{}}: \textless{}\textless{}your option\textgreater{}\textgreater{} \\
\end{tabular} \\
\midrule

\begin{tabular}[c]{@{}c@{}}Knowledge Concept\\Augmentation\end{tabular}   & \begin{tabular}[c]{@{}l@{}}Now, we require you to solve a multiple-choice math question. We will provide \\ you with the relevant knowledge concepts of this question for your reference. \\ Please briefly describe your thought process and provide the final answer(option).\\ 
\textbf{Knowledge concept}: \textless{}Knowledge concept\textgreater\\
\textbf{Question}: \textless{}Question\textgreater\\
\textbf{Option}: \textless{}Option\textgreater\\
Regarding the format, please answer following the template below, and be \\ sure to include two \textless{}\textgreater{} symbols: \\
\textbf{\textless{}Thought process\textgreater{}}: \textless{}\textless{}your thought process\textgreater{}\textgreater{} \textbf{\textless{}Answer\textgreater{}}: \textless{}\textless{}your option\textgreater{}\textgreater{}  \\
\end{tabular} \\
\bottomrule
\end{tabular}
\label{tab:prompt_query}
\end{table}

\subsection{Knowledge Concepts Augmentation}
\label{subapp:Augmentation}

Table~\ref{tab:prompt_query} report the prompt templates in our experiments. We concatenate the textual descriptions into the prompt. Additionally, each knowledge concept description is accompanied by its corresponding visual content, which helps the experimenter understand and facilitates further enhancement when models can incorporate sufficient visual information as part of the prompt in the future. 

In section \ref{ex des}, we illustrates the specific content of descriptions for 67 knowledge concepts. For example, as shown in Figure~\ref{fig:CPF}, for the knowledge concept "Perimeter of Squares," it is necessary to know that "c=4a", relying solely on textual descriptions is insufficient for understanding this concept, so we include visual information to aid comprehension.


\subsection{Details of Data Collection}
\label{subapp:Collection Details}

With the hierarchical knowledge structure, we select geometric problems with images from publicly authoritative mathematics websites from various countries, including professional exams and practice tests (detailed sources list can be found in section \ref{data sorce}). To ensure comprehensive coverage of fundamental and critical areas in primary math, we select the five most foundational and prevalent domains within the field of primary geometry, including:
\begin{itemize}[leftmargin = 15pt] 
    \item \textbf{Plane figures}: Questions involving identification and properties of two-dimensional shapes.
    \item \textbf{Solid figures}: Questions related to the recognition and characteristics of three-dimensional objects.
    \item \textbf{Transformation and motion of figures}: Problems focusing on geometric transformations such as translation, rotation, and reflection.
    \item \textbf{Position and direction}: Questions that involve understanding spatial relationships and directions.
    \item \textbf{Measurement}: Problems requiring the measurement of length, area, volume, and angles.
\end{itemize}
The selection criteria are as follows:
(1) The problems include multiple knowledge concepts and can be decomposed into steps for solution.
(2) The problems and images are consistent.
(3) The correct answer is unique, and the distractor options are highly confusing.

\clearpage



\subsection{Details of Data Statistics}
\label{subapp:Statistics}

\vspace{-0.3cm}
\begin{figure}[ht]
    \centering
    \resizebox{\textwidth}{!}{
    \includegraphics{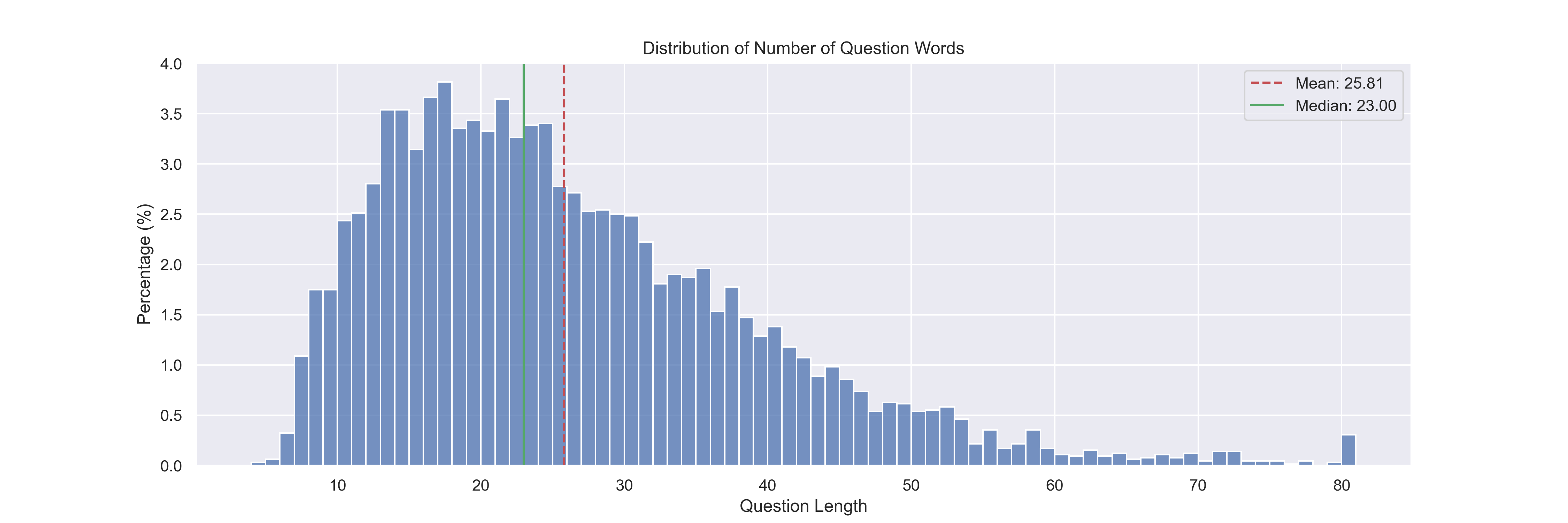}
    }
    \caption{The distribution of the number of words per question in \dataset. Questions with a length greater than 80 are categorized as 81 for visualization simplicity.}
    \label{fig:statistics}
\end{figure}

\begin{wraptable}{r}{0.53\textwidth}
\caption{Key statistics of \dataset.} \label{tab:statistic}
\begin{tabular}{lc}
\toprule
\textbf{Statistic} & \textbf{Number} \\
\midrule
Total questions & 6,524 \\
Newly collected questions & 6,524 \\
Multiple-choice questions & 6,524 \\
-First-layer nodes & 5 \\
-Second-layer nodes & 12 \\
-Terminal nodes & 67 \\
\midrule
Question options & \\
-Total options & 25,178 \\
-Average options & 3.859 \\
-Proportion of answer A & 6,524 (25.9\%) \\
-Proportion of answer B & 6,524 (25.9\%) \\
-Proportion of answer C & 6,505 (25.8\%) \\
-Proportion of answer D & 4,419 (17.6\%) \\
-Proportion of answer E & 1,198 (4.8\%) \\
-Proportion of answer F\&G & 11 (0.04\%) \\
\midrule
Question length & \\
-Maximum length (word) & 143 \\
-Maximum length (character) & 852 \\
-Average length (word) & 25.8 \\
-Average length (character) & 135.3 \\
\bottomrule
\end{tabular}
\end{wraptable}

\textbf{Question distribution.} 
The \dataset consists entirely of English questions, all newly collected from public authoritative mathematics websites, and presented in the format of multiple-choice questions. As illustrated in Table \ref{tab:statistic}, the average number of words in the English questions within \dataset is 25.81, with the maximum number of words in a question reaching 143. Figure \ref{fig:statistics} further elaborates on the distribution of word counts, highlighting the diverse patterns of the questions.

\textbf{Advantages of Multiple-Choice Questions.}

In \dataset, all problems are presented as multiple-choice questions. Even if some problems did not originally conform to the multiple-choice format during the initial selection, our researchers manually converted them into the format. Using multiple-choice questions offers several advantages:

\textbf{Standardization}: Ensures uniformity across all questions, facilitating consistent assessment and comparison across different hierarchical subjects.

\textbf{Objective Grading}: The use of single correct answers eliminates subjectivity in grading, enhancing the reliability of the evaluation.

\textbf{Efficiency}: Allows for rapid and scalable assessment, suitable for large datasets and automated systems.

\textbf{Focused Assessment}: Carefully designed distractors help in accurately identifying specific knowledge gaps and common misconceptions.

\section{More Details on the Metrics}
\label{sec:Metric}

\begin{figure}[!t]
    \centering
    \resizebox{\textwidth}{!}{
    \includegraphics{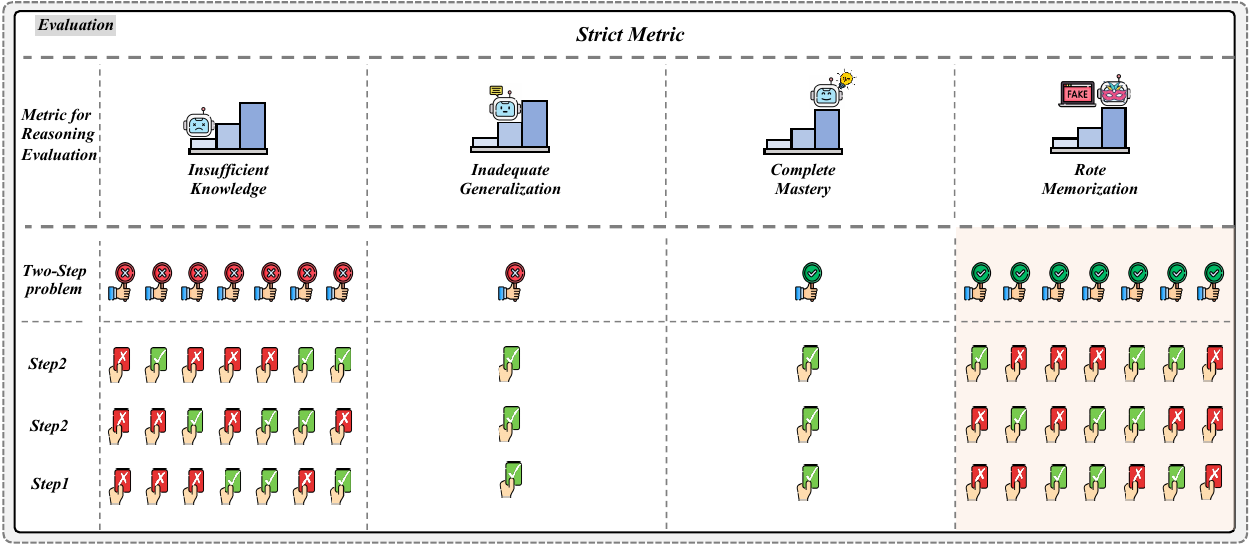}
    }
    \caption{Diagram illustrating strict metric in three-step problem.}
    \label{fig:strict-3}
\end{figure}

\begin{figure}[!t]
    \centering
    \resizebox{\textwidth}{!}{
    \includegraphics{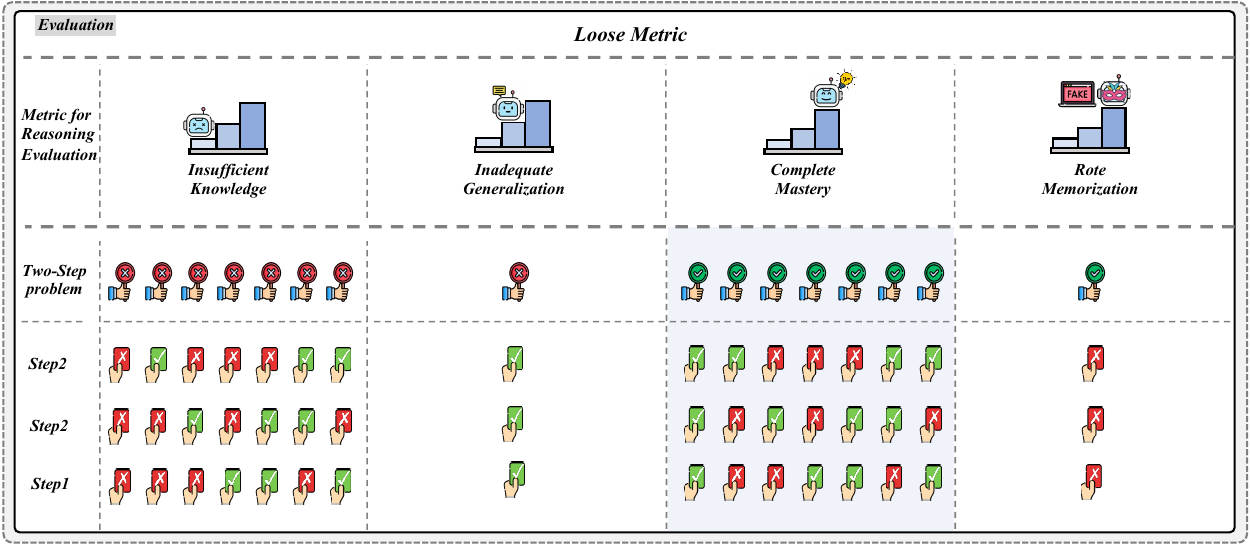}
    }
    \caption{Diagram illustrating loose metric in three-step problem.}
    \label{fig:loose-3}
\end{figure}

\textbf{Distinguishing Metric.} Considering the model's instability, Figure~\ref{fig:metric-main} and Figure~\ref{fig:strict-3},~\ref{fig:loose-3} illustrate the two metrics we propose for distinguishing between \textit{RM} and \textit{CM} metrics. Figure~\ref{fig:metric-main} represents the two-step problem, while Figures~\ref{fig:strict-3} and Figures~\ref{fig:loose-3} represent the three-step problem. Specifically, under the strict metric, if there is any error in the corresponding sub-problems of a multi-step problem that is answered correctly, it is classified as \textit{RM (Rote Memorization)}. Only if all corresponding sub-problems are answered correctly (TTTT, TTT) is it classified as \textit{CM (Complete Master)}. Under the loose metric, it is classified as \textit{RM} only if the model answers all sub-problems incorrectly (FFFT, FFT), otherwise, it is classified as \textit{CM}. Therefore, the $\text{Score}_{\text{average}}$ under the loose metric is slightly higher. We hope to see models like GPT-4o~\cite{GPT4o} and GPT-4V~\cite{2023GPT4VisionSC}, which have already performed nearly perfectly under the loose metric and are far ahead of other models, bring us even greater surprises under the strict metric in the next update.

\textbf{Metrics' Intrinsic Logic.} As shown in Figure~\ref{fig:metric-main},~\ref{fig:strict-3},~\ref{fig:loose-3}, it is evident in the Metric for Reasoning Evaluation Section that \textit{IK}, \textit{IG}, and \textit{CM} have a logical relationship. In the early stages of constructing \dataset, we recorded all the model's responses and analyzed the answers to each multi-step problem and its corresponding sub-problems. We believe that for both humans and models, a reasonable learning process should involve first mastering each knowledge concept individually and then learning to comprehensively apply them to achieve complete mastery. The situation where the multi-step problem is answered correctly but the sub-problems are answered incorrectly (\textit{RM}) is an unreasonable phenomenon.
Therefore, we developed a four-dimensional fine-grained metric to further evaluate the model's performance.



\section{More Details on Experiment Setup}
\label{subapp:Experiment set up}


\subsection{Details of the Evaluated Models}
To evaluate the mathematical reasoning abilities of various LMMs, we selected their latest model versions. Table \ref{tab:release_time_model_source} presents their release dates and specific sources. Given the intuition that smaller models (with parameters of 7B or less) perform poorly on \textit{Insufficient Knowledge (IK)}, we also included evaluations of the latest models with 7B, 4.2B, and 1.3B parameters. This was done to explore whether these models could achieve significant improvement under the KCA strategy.

\label{subapp:prompt}

\setlength{\extrarowheight}{2pt}
\hypersetup{breaklinks=true}
\begin{table}[!]
    \centering
    \caption{The release time and model source of LMMs used in \dataset}
    \label{tab:release_time_model_source}
        \renewcommand{\arraystretch}{1.2} 
    \resizebox{\textwidth}{!}{
    \begin{tabular}{lcc}
    \toprule
    \textbf{Model} & \textbf{Release Time} & \textbf{Source} \\
    \midrule
    GPT-4o~\cite{GPT4o} & 2024-05 & \url{https://gpt4o.ai/} \\\hline
    GPT-4V~\cite{2023GPT4VisionSC} & 2024-04 & \url{https://openai.com/index/gpt-4v-system-card/} \\\hline
    Gemini 1.5 Pro~\cite{gemini} & 2024-05 & \url{https://deepmind.google/technologies/gemini/pro/} \\\hline
    Qwen-VL-Max~\cite{bai2023qwenvl} & 2024-01 & \url{https://huggingface.co/spaces/Qwen/Qwen-VL-Max/} \\\hline
    LLaVA-NeXT-110B~\cite{liu2024llavanext} & 2024-05 & \url{https://huggingface.co/lmms-lab/llava-next-110b/} \\\hline
    LLaVA-NeXT-72B~\cite{liu2024llavanext} & 2024-05 & \url{https://huggingface.co/lmms-lab/llava-next-72b/} \\\hline
    LLaVA-1.6-13B~\cite{liu2023improvedllava} & 2024-03 & \url{https://huggingface.co/llava-hf/llava-v1.6-vicuna-13b-hf/} \\\hline 
    LLaVA-1.6-7B~\cite{liu2023improvedllava} & 2024-03 & \url{https://huggingface.co/llava-hf/llava-v1.6-vicuna-7b-hf/} \\\hline 
    DeepSeek-VL-1.3B~\cite{lu2024deepseekvl} & 2024-03 & \url{https://huggingface.co/deepseek-ai/deepseek-vl-1.3b-chat/} \\\hline
    DeepSeek-VL-7B~\cite{lu2024deepseekvl} & 2024-03 & \url{https://huggingface.co/deepseek-ai/deepseek-vl-7b-chat/} \\\hline
    Phi3-Vision-4.2B~\cite{abdin2024phi3} & 2024-05 & \url{https://huggingface.co/microsoft/Phi-3-vision-128k-instruct/} \\\hline
    MiniCPM-LLaMA3-V 2.5~\cite{hu2024large} & 2024-05 & \url{https://huggingface.co/openbmb/MiniCPM-Llama3-V-2_5/} \\\hline
    InternLM-XComposer2-VL-7B~\cite{internlmxcomposer2} & 2024-04 & \url{https://huggingface.co/internlm/internlm-xcomposer2-vl-7b/} \\\hline 
    InternVL-Chat-V1.5~\cite{chen2024internvl} & 2024-04 & \url{https://huggingface.co/OpenGVLab/InternVL-Chat-V1-5/} \\\hline 
    GLM-4V-9B~\cite{glm2024chatglm} & 2024-06 & \url{https://huggingface.co/THUDM/glm-4v-9b} \\\hline 
    
    LongVA~\cite{zhang2024longcontexttransferlanguage} & 2024-06 & \url{https://huggingface.co/lmms-lab/LongVA-7B} \\\hline 
    
    G-LLaVA-13B~\cite{gao2023gllava} & 2024-03 & \url{https://huggingface.co/renjiepi/G-LLaVA-13B/} \\
    \bottomrule
    \end{tabular}
    }
\end{table}

\subsection{Details of the Model Hyperparameters}
For all closed-sourced models with API access, we adopt the generation scheme shown in Table \ref{tab:closed_model_hyperparameter} and simply run the inference with CPUs, which typically completes within a day. For all open-source models, we utilize a cluster with 8 NVIDIA A800-SXM4-80GB GPUs to run the inference, and we follow the hyper-parameter settings specified in the model source's inference samples. If no specific instructions are provided, we use the default settings. Table \ref{tab:open_model_hyperparameter} details the specific generation parameters.

\begin{table}[!]
  \centering
    \caption{Generating parameters for Closed-Source LMMs.}
    \label{tab:closed_model_hyperparameter}
    \footnotesize

\begin{tabular}{c|l}
\toprule
\multicolumn{1}{c|}{\textbf{Model}}     & \multicolumn{1}{c}{\textbf{Generation Setup}}\\ 
\midrule
\begin{tabular}[c]{@{}c@{}}GPT-4o\end{tabular}   & \begin{tabular}[c]{@{}l@{}} "model" : "gpt-4o", "temperature" : 0, "max\_tokens" : 1024 \\
\end{tabular} \\
\midrule

\begin{tabular}[c]{@{}c@{}}GPT-4V\end{tabular}   & \begin{tabular}[c]{@{}l@{}} "model" : "gpt-4-turbo", "temperature" : 0, "max\_tokens" : 1024 \\
\end{tabular} \\
\midrule

\begin{tabular}[c]{@{}c@{}}Gemini 1.5 Pro\end{tabular}   & \begin{tabular}[c]{@{}l@{}} "model" : "gemini-1.5-pro-latest", "temperature" : 0, "max\_tokens" : 1024 \\
\end{tabular} \\
\midrule

\begin{tabular}[c]{@{}c@{}}Qwen-VL-Max\end{tabular}   & \begin{tabular}[c]{@{}l@{}} "model" : "qwen-vl-max", "temperature" : 0, "max\_tokens" : 1024 \\
\end{tabular} \\

\bottomrule
\end{tabular}
\end{table}

\begin{table}[!]
  \centering
    \caption{Generating parameters for Open-Source LMMs.}
    \label{tab:open_model_hyperparameter}
    \footnotesize

\begin{tabular}{c|l}
\toprule
\multicolumn{1}{c|}{\textbf{Model}}     & \multicolumn{1}{c}{\textbf{Generation Setup}}\\ 
\midrule
\begin{tabular}[c]{@{}c@{}}LLaVA-NeXT-110B\end{tabular}   & \begin{tabular}[c]{@{}l@{}} do\_sample = False, temperature = 0, max\_new\_tokens = 1024 \\
\end{tabular} \\
\midrule

\begin{tabular}[c]{@{}c@{}}LLaVA-NeXT-72B\end{tabular}   & \begin{tabular}[c]{@{}l@{}} do\_sample = False, temperature = 0, max\_new\_tokens = 1024 \\
\end{tabular} \\
\midrule

\begin{tabular}[c]{@{}c@{}}LLaVA-1.6-13B\end{tabular}   & \begin{tabular}[c]{@{}l@{}} do\_sample = False, temperature = 0, max\_new\_tokens = 1024 \\
\end{tabular} \\
\midrule

\begin{tabular}[c]{@{}c@{}}LLaVA-1.6-7B\end{tabular}   & \begin{tabular}[c]{@{}l@{}} do\_sample = False, temperature = 0, max\_new\_tokens = 1024 \\
\end{tabular} \\
\midrule

\begin{tabular}[c]{@{}c@{}}DeepSeek-VL-1.3B\end{tabular}   & \begin{tabular}[c]{@{}l@{}} do\_sample = False, max\_new\_tokens = 1024
 \\
\end{tabular} \\
\midrule

\begin{tabular}[c]{@{}c@{}}DeepSeek-VL-7B\end{tabular}   & \begin{tabular}[c]{@{}l@{}} do\_sample = False, max\_new\_tokens = 1024 \\
\end{tabular} \\
\midrule

\begin{tabular}[c]{@{}c@{}}Phi3-Vision-4.2B\end{tabular}   & \begin{tabular}[c]{@{}l@{}} do\_sample = False, temperature = 0, max\_new\_tokens = 1024 \\
\end{tabular} \\
\midrule

\begin{tabular}[c]{@{}c@{}}MiniCPM-LLaMA3-V 2.5\end{tabular}   & \begin{tabular}[c]{@{}l@{}} sampling = True, temperature = 0.7 \\
\end{tabular} \\
\midrule

\begin{tabular}[c]{@{}c@{}}InternLM-XComposer2-VL-7B\end{tabular}   & \begin{tabular}[c]{@{}l@{}} do\_sample = False\\
\end{tabular} \\
\midrule

\begin{tabular}[c]{@{}c@{}}InternVL-Chat-V1.5\end{tabular}   & \begin{tabular}[c]{@{}l@{}} num\_beams = 1, do\_sample = False, max\_new\_tokens = 1024  \\
\end{tabular} \\
\midrule

\begin{tabular}[c]{@{}c@{}}GLM-4V-9B\end{tabular}   & \begin{tabular}[c]{@{}l@{}} do\_sample = True, max\_length = 1024, top\_k = 1  \\
\end{tabular} \\
\midrule

\begin{tabular}[c]{@{}c@{}}LongVA\end{tabular}   & \begin{tabular}[c]{@{}l@{}} do\_sample = False, temperature = 0, max\_new\_tokens = 1024, num\_beams = 1  \\
\end{tabular} \\
\midrule

\begin{tabular}[c]{@{}c@{}}G-LLaVA-13B\end{tabular}   & \begin{tabular}[c]{@{}l@{}}  do\_sample = True, temperature = 0.2, max\_new\_tokens = 1024 \\
\end{tabular} \\

\bottomrule
\end{tabular}
\end{table}

\begin{table}[!]
    \centering
    \footnotesize
    \caption{Model architecture of $17$ LMMs evaluated on We-Math.}
    
        \setlength\tabcolsep{0.8pt}
    \label{tab:LMMs Architecture}
    \begin{tabular}{lll}
        \toprule
        \textbf{Models} & \textbf{LLM} & \textbf{Vision Encoder}\\ 
        \midrule
        GPT-4o  & - & - \\ 
        GPT-4V  & - & - \\
        Gemini 1.5 Pro & - & - \\ 
        Qwen-VL-Max & - & - \\ 
        LLaVA-NeXT-110B & Qwen1.5-110B-Chat & CLIP-ViT-L-P14-336 \\
        LLaVA-NeXT-72B & Qwen1.5-72B-Chat & CLIP-ViT-L-P14-336 \\
        LLaVA-1.6-13B & Vicuna-13B-v1.5 & CLIP-ViT-L-P14-336 \\ 
        LLaVA-1.6-7B & Vicuna-7B-v1-5 & CLIP-ViT-L-P14-336 \\ 
        DeepSeek-VL-1.3B & DeepSeek-LLM-1.3B-base & SigLIp-L-P16-384 \\ 
        DeepSeek-VL-7B & DeepSeek-LLM-7B-base & SigLIp-L-P16-384 \& SAM-B \\ 
        Phi3-Vision-4.2B & Phi-3-mini-128K-instruct & CLIP-ViT-L-P14-336 \\
        MiniCPM-LLaMA3-V 2.5 & Llama3-8B-Instruct & SigLIp-L-P14-384\\
        InternLM-XComposer2-VL-7B & InternLM2-7B-ChatSFT &  CLIP-ViT-L-P14-336\\
        InternVL-Chat-V1.5 & InternLM2-Chat-20B & InternViT-6B-448px-V1-5 (6B)\\
        GLM-4V-9B & GLM-9B & EVA\_02\_CLIP-E-P14 (4.7B)\\
        LongVA & Qwen2-7B-Instruct & CLIP-ViT-L-P14-336\\
        G-LLaVA-13B & Vicuna-13B-v1.5 & CLIP-ViT-L-P14-336 \\
        \bottomrule
    \end{tabular}
\end{table}

\clearpage



\section{More Details on Experiment Results}
\label{subapp:results}

\subsection{Details of Model Performance}
\label{sec:Details Performance}

\textbf{The Leaderboard on \dataset.} We present the visualization results of the $\text{Score}_{\text{average}}$ under the loose and strict metric in Figure~\ref{fig:619-leaderboard}, respectively. GPT-4o shows a significant lead under both metric, and LLaVA-NeXT-110B performs the best among open-source models. Impressively, InternVL-Chat-V1.5 and GLM-4V-9B achieved excellent scores, surpassing the closed-source model Qwen-VL-Max. Additionally, some recently proposed smaller models (such as Phi-3-Vision-4.2B, InternLM-XComposer2-VL-7B, and MiniCPM-LLaMA3-V 2.5) also demonstrated outstanding performance, suggesting that optimizing training methods might partially substitute for the performance gains typically achieved by increasing the parameter count.


\textbf{Detailed Performance of Four-Dimensional Metrics.} 
Figure~\ref{fig:main_strict_4metric} and Figure~\ref{fig:main_loose_4metric} display the specific performance of LMMs under both loose and strict metric across four metrics. Focusing on the \textit{IK} metric, GPT-4o has the fewest instances under both metric, indicating that GPT-4o has the best grasp of the knowledge concepts. Furthermore, for the \textit{IG} metric, we find that GPT-4o and GPT-4V have the highest exposure compared to other models. As discussed in the previous Section~\ref{sec:Metric}, \textit{IG} issues only arise after addressing \textit{IK} issues, which further indicates that GPT-4 is progressing to the next stage. Focusing on the \textit{CM} and \textit{RM} metrics, GPT-4o and GPT-4V continue to show significant leadership. Both models excel in the \textit{CM} metric, where the number of correctly answered multi-step problems and their corresponding sub-questions is significantly higher than that of other models. Additionally, comparing GPT-4o and GPT-4V under strict metric, GPT-4o consistently outperforms GPT-4V, aligning with the $\text{Score}_{\text{average}}$ results.





\begin{figure}[ht]
    \centering
    \resizebox{\textwidth}{!}{
    \includegraphics{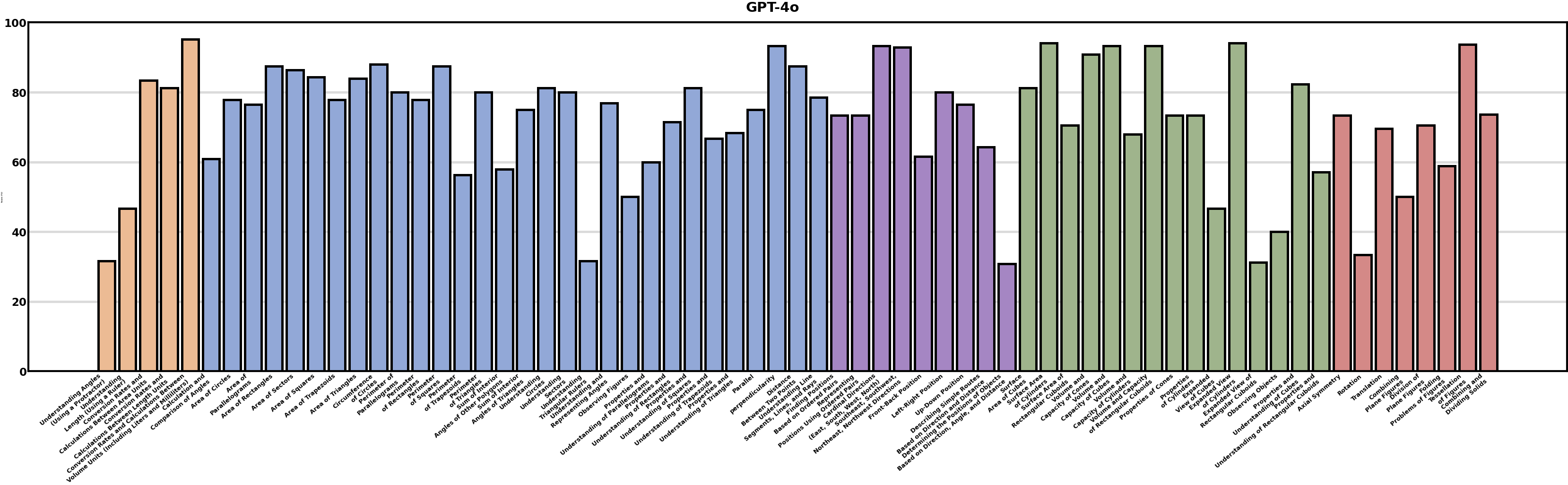}
    }
    \caption{Detailed performance of GPT-4o across 67 knowledge concepts.}
    \label{fig:GPT-4o}
\end{figure}

\begin{figure}[ht]
    \centering
    \resizebox{\textwidth}{!}{
    \includegraphics{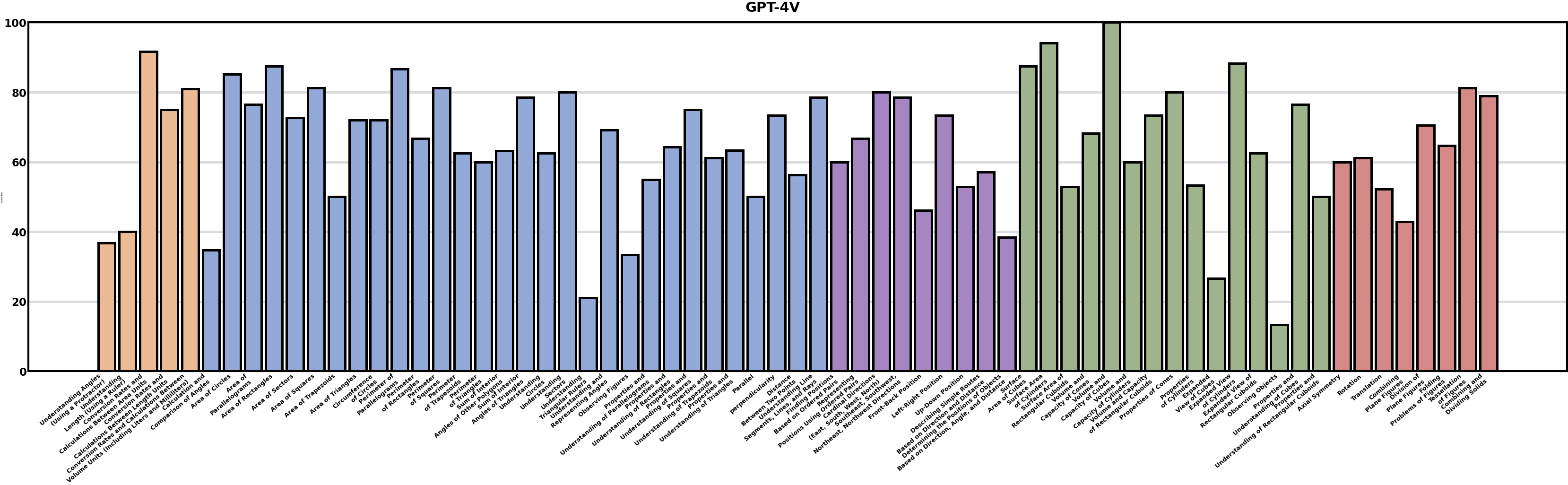}
    }
    \caption{Detailed performance of GPT-4V across 67 knowledge concepts.}
    \label{fig:GPT-4V}
\end{figure}

\begin{figure}[ht]
    \centering
    \resizebox{\textwidth}{!}{
    \includegraphics{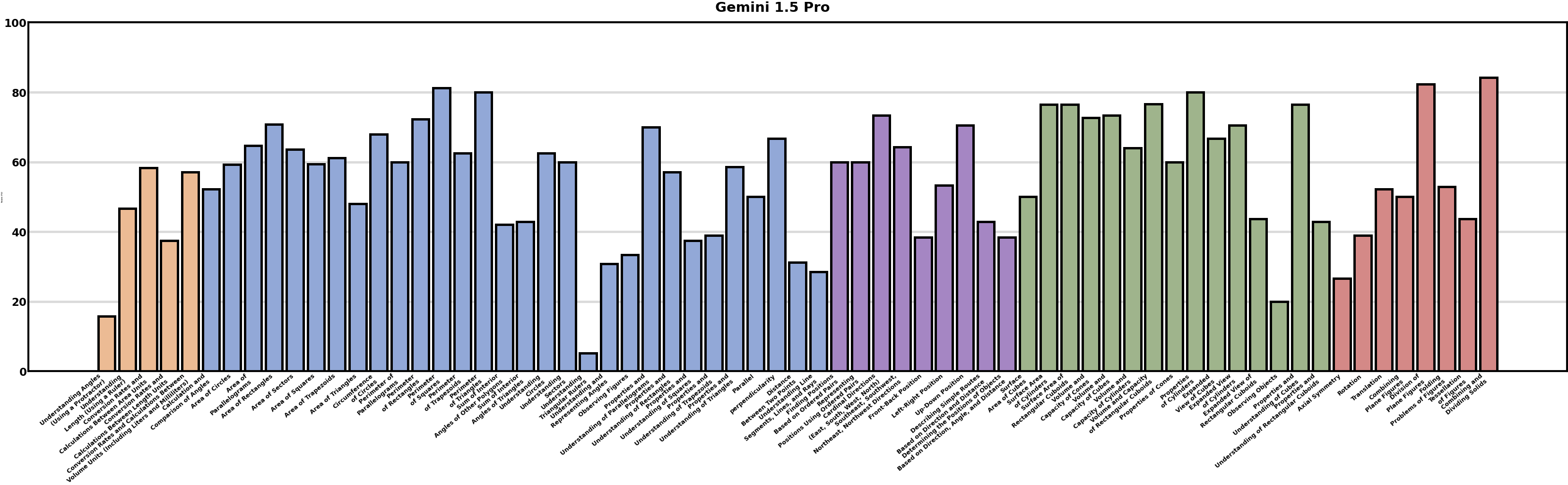}
    }
    \caption{Detailed performance of Gemini 1.5 Pro across 67 knowledge concepts.}
    \label{fig:Gemini}
\end{figure}

\begin{figure}[ht]
    \centering
    \resizebox{\textwidth}{!}{
    \includegraphics{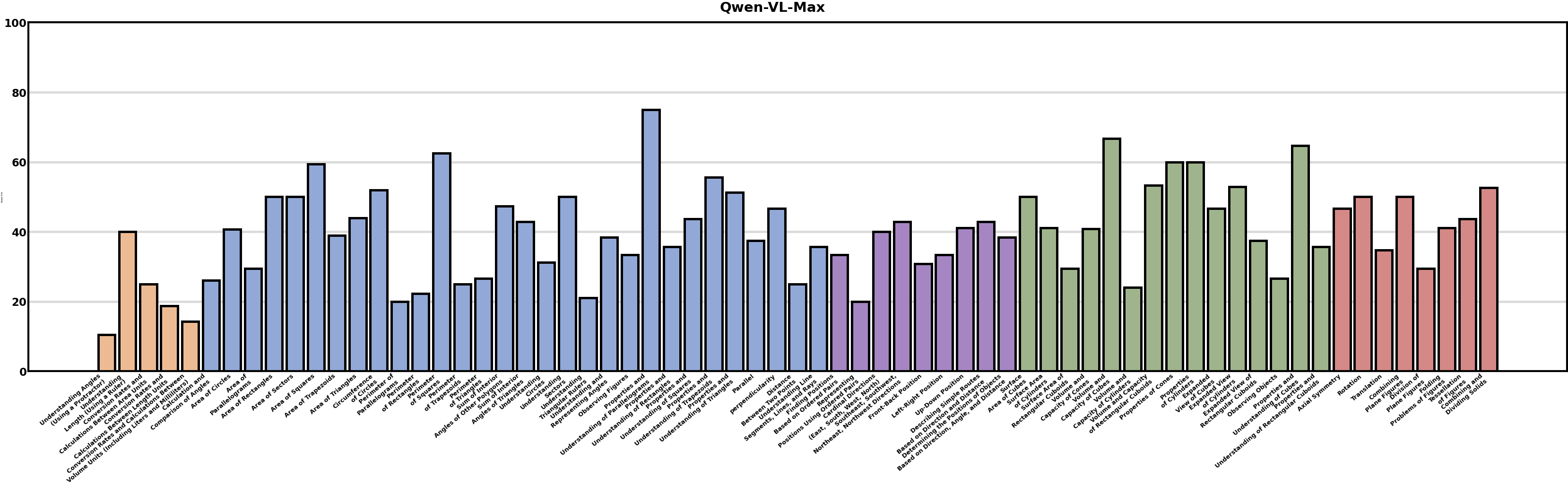}
    }
    \caption{Detailed performance of Qwen-VL-Max across 67 knowledge concepts.}
    \label{fig:Qwen-VL-Max}
\end{figure}

\begin{figure}[!t]
    \centering
    \resizebox{\textwidth}{!}{
    \includegraphics{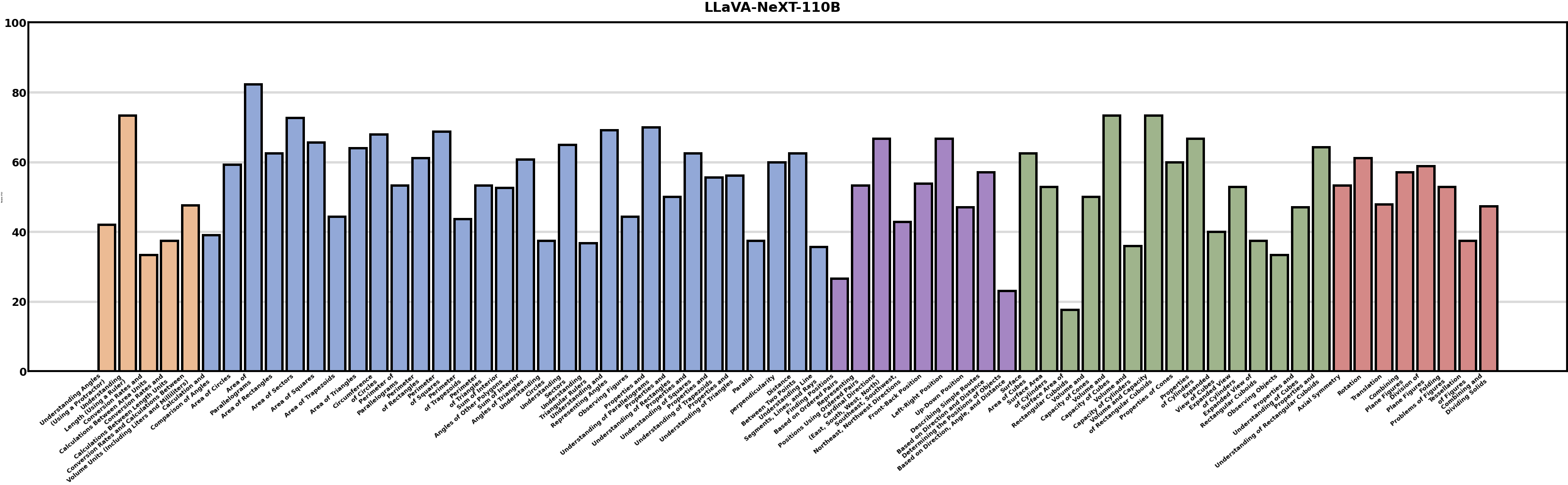}
    }
    \caption{Detailed performance of LLaVA-NeXT-110B across 67 knowledge concepts.}
    \label{fig:LLaVA-NeXT-110B}
\end{figure}

\begin{figure}[!t]
    \centering
    \resizebox{\textwidth}{!}{
    \includegraphics{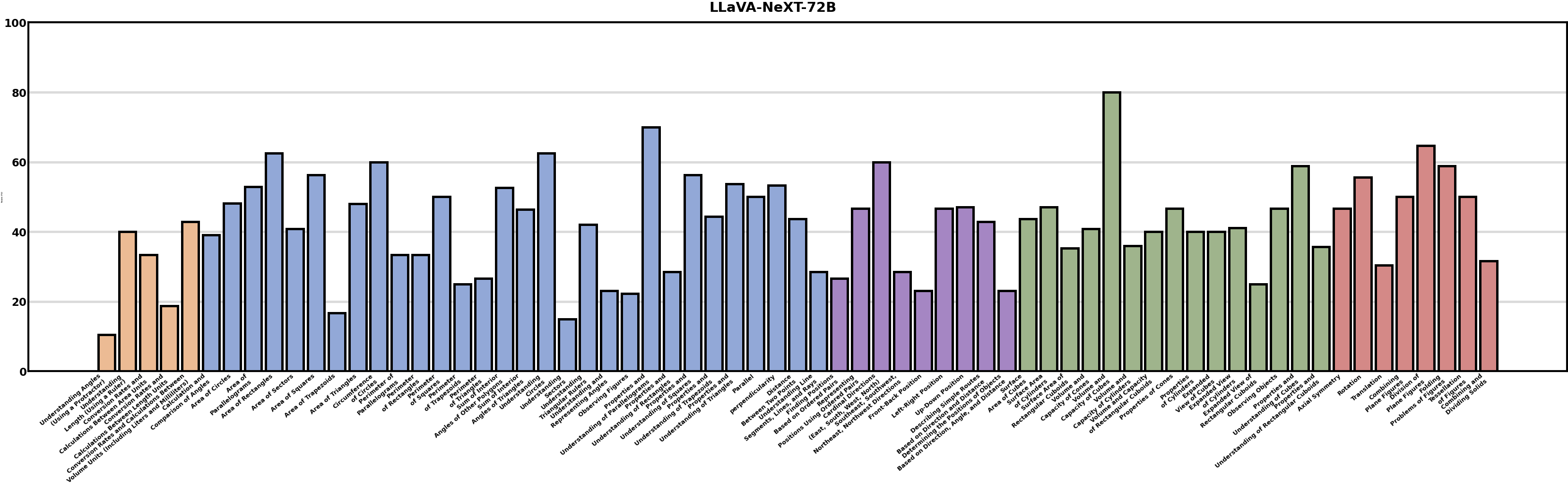}
    }
    \caption{Detailed performance of LLaVA-NeXT-72B across 67 knowledge concepts.}
    \label{fig:LLaVA-NeXT-72B}
\end{figure}

\begin{figure}[!t]
    \centering
    \resizebox{\textwidth}{!}{
    \includegraphics{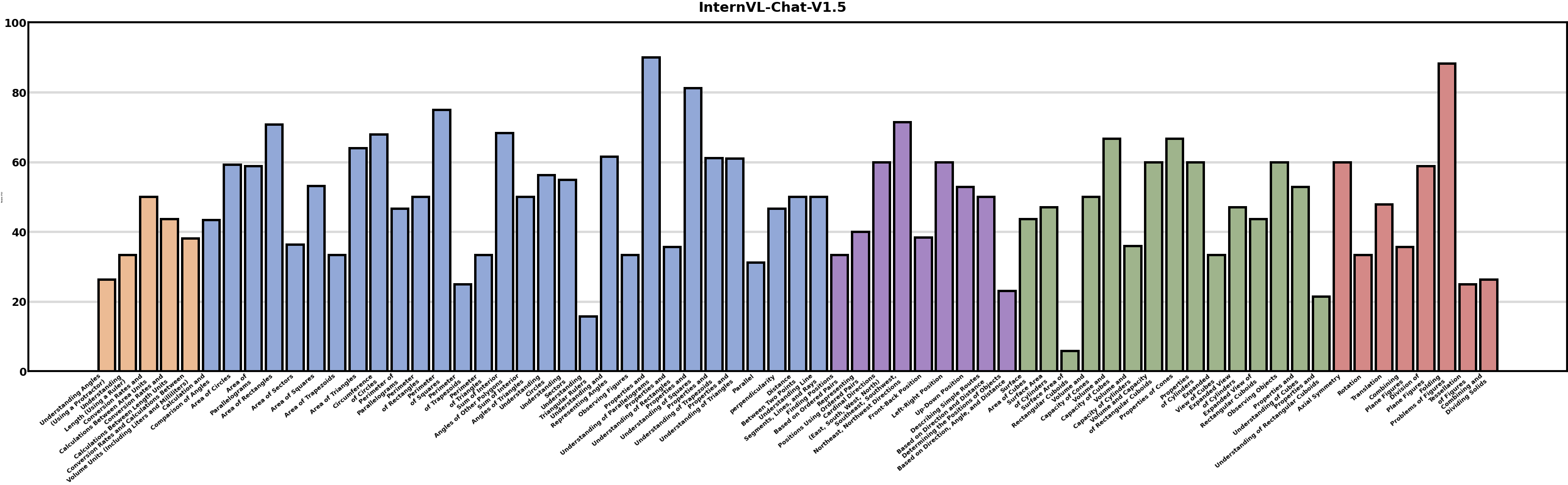}
    }
    \caption{Detailed performance of InternVL-Chat-V 1.5 across 67 knowledge concepts.}
    \label{fig:InternVL-Chat}
\end{figure}

\begin{figure}[!t]
    \centering
    \resizebox{\textwidth}{!}{
    \includegraphics{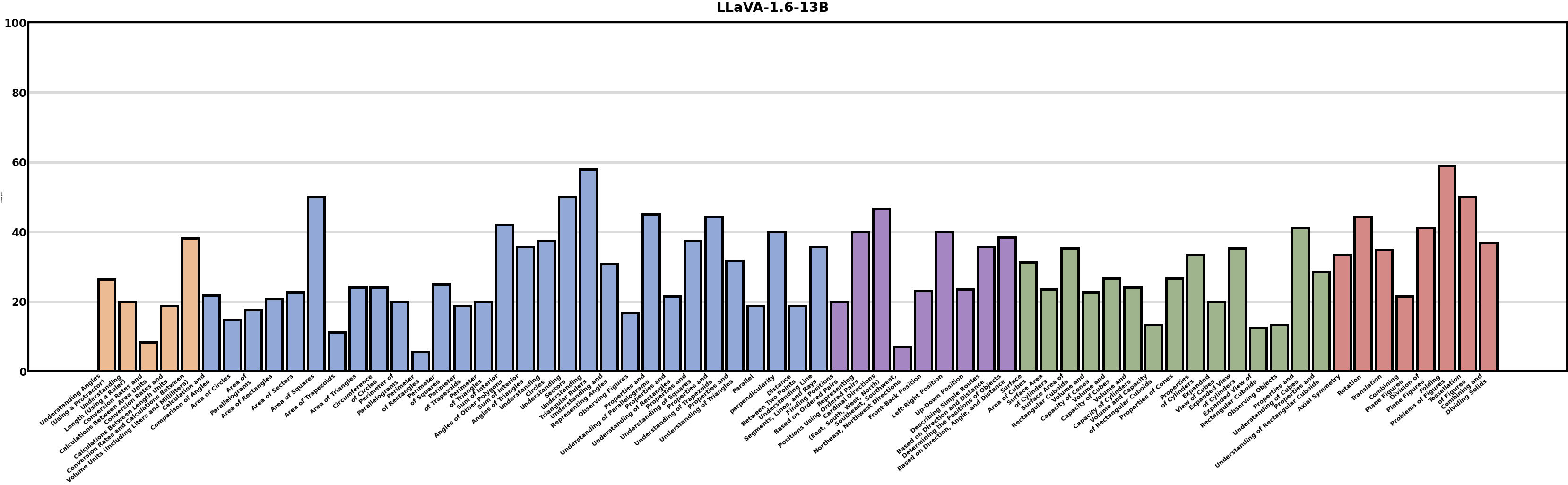}
    }
    \caption{Detailed performance of LLaVA-1.6-13B across 67 knowledge concepts.}
    \label{fig:LLaVA-1.6-13B}
\end{figure}

\begin{figure}[!t]
    \centering
    \resizebox{\textwidth}{!}{
    \includegraphics{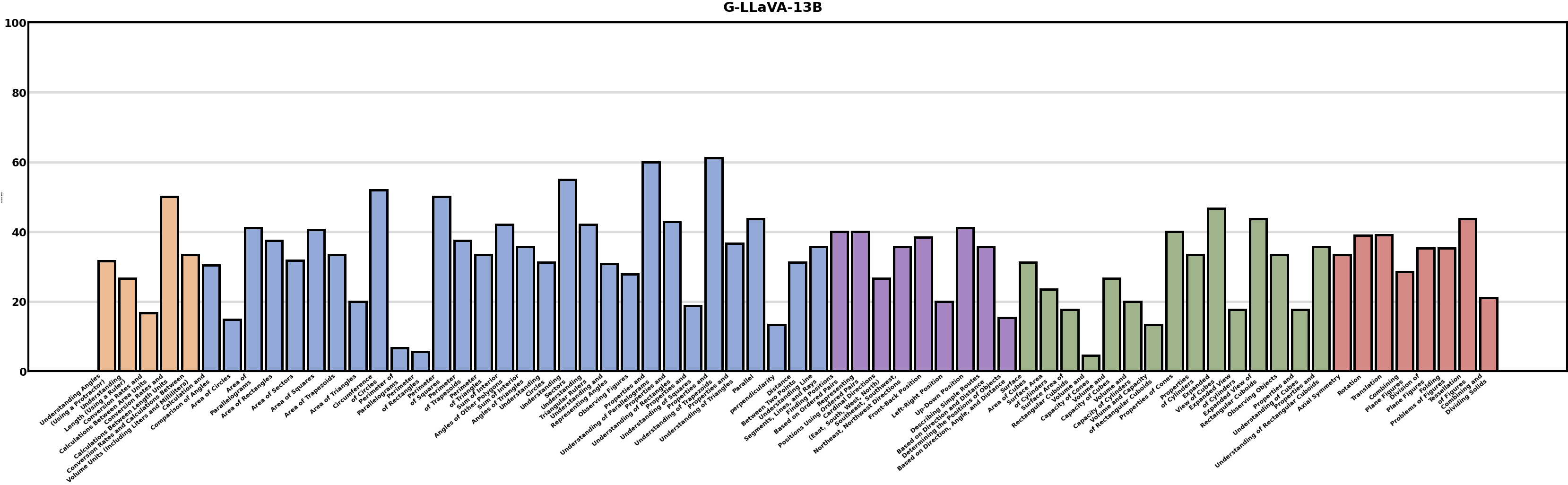}
    }
    \caption{Detailed performance of G-LLaVA-13B across 67 knowledge concepts.}
    \label{fig:G-LLaVA-13B}
\end{figure}

\begin{figure}[!t]
    \centering
    \resizebox{\textwidth}{!}{
    \includegraphics{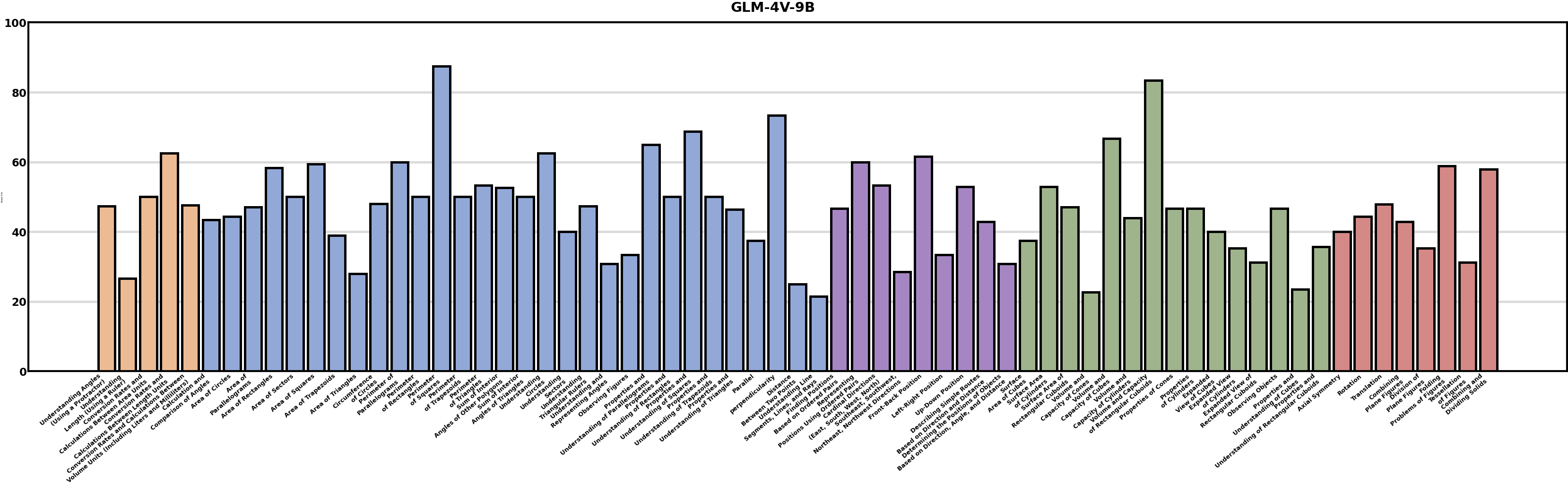}
    }
    \caption{Detailed performance of GLM-4V-9B across 67 knowledge concepts.}
    \label{fig:GLM}
\end{figure}

\begin{figure}[!t]
    \centering
    \resizebox{\textwidth}{!}{
    \includegraphics{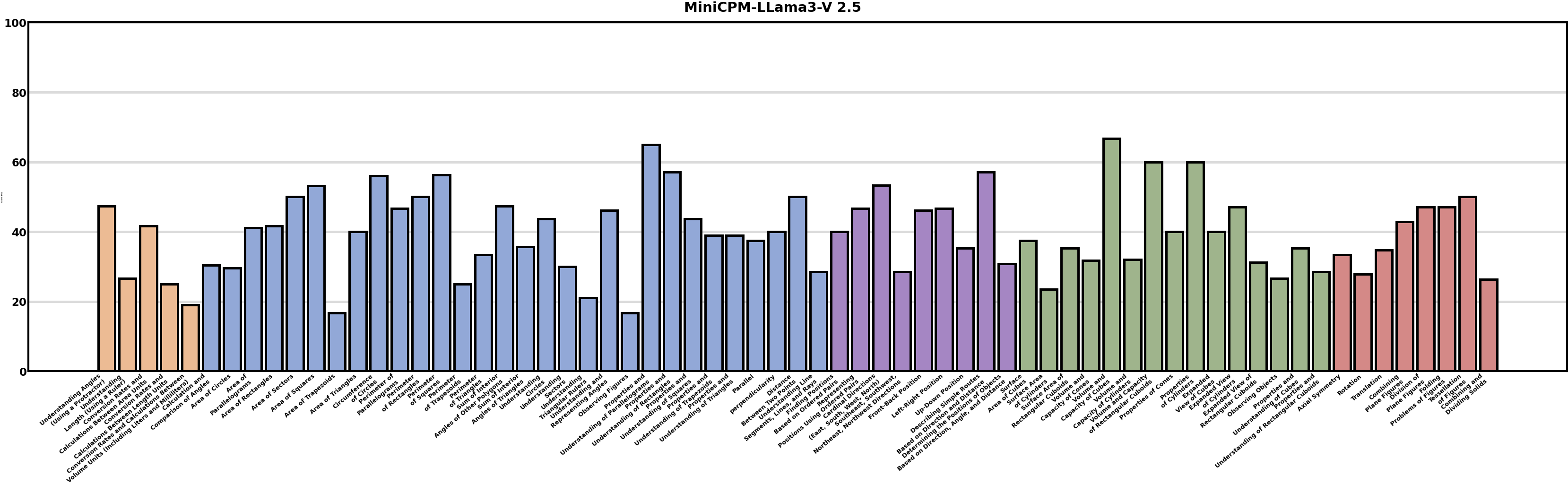}
    }
    \caption{Detailed performance of MiniCPM-LLama3-V 2.5 across 67 knowledge concepts.}
    \label{fig:MiniCPM-LLama3}
\end{figure}

\begin{figure}[!t]
    \centering
    \resizebox{\textwidth}{!}{
    \includegraphics{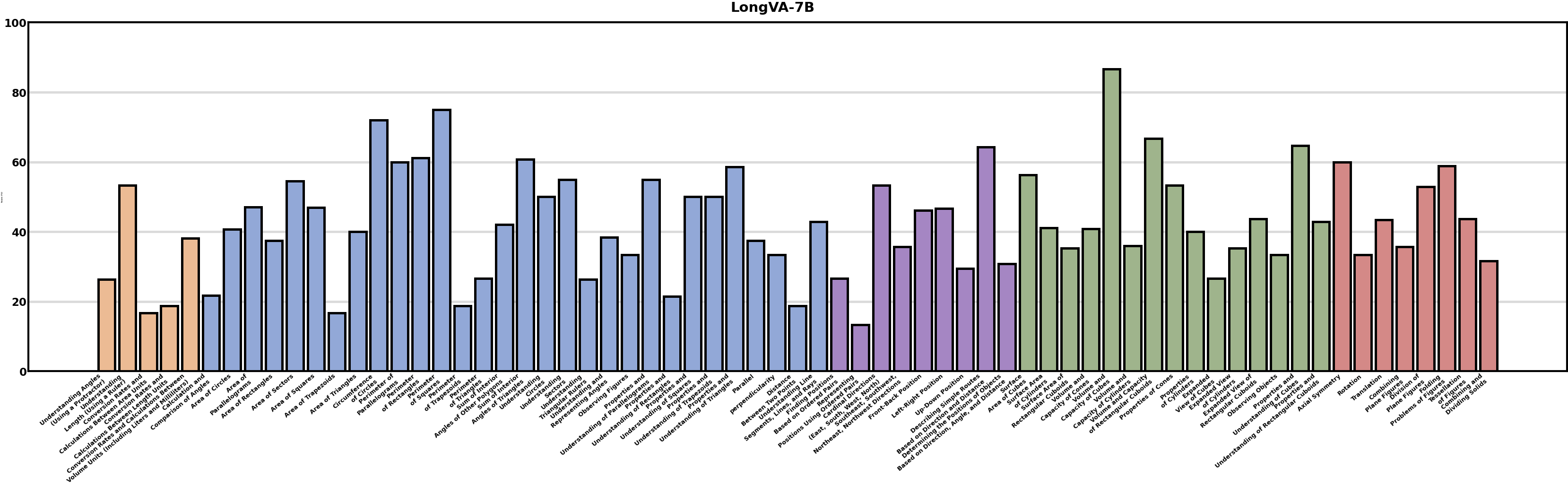}
    }
    \caption{Detailed performance of LongVA-7B across 67 knowledge concepts.}
    \label{fig:LongVA}
\end{figure}

\begin{figure}[!t]
    \centering
    \resizebox{\textwidth}{!}{
    \includegraphics{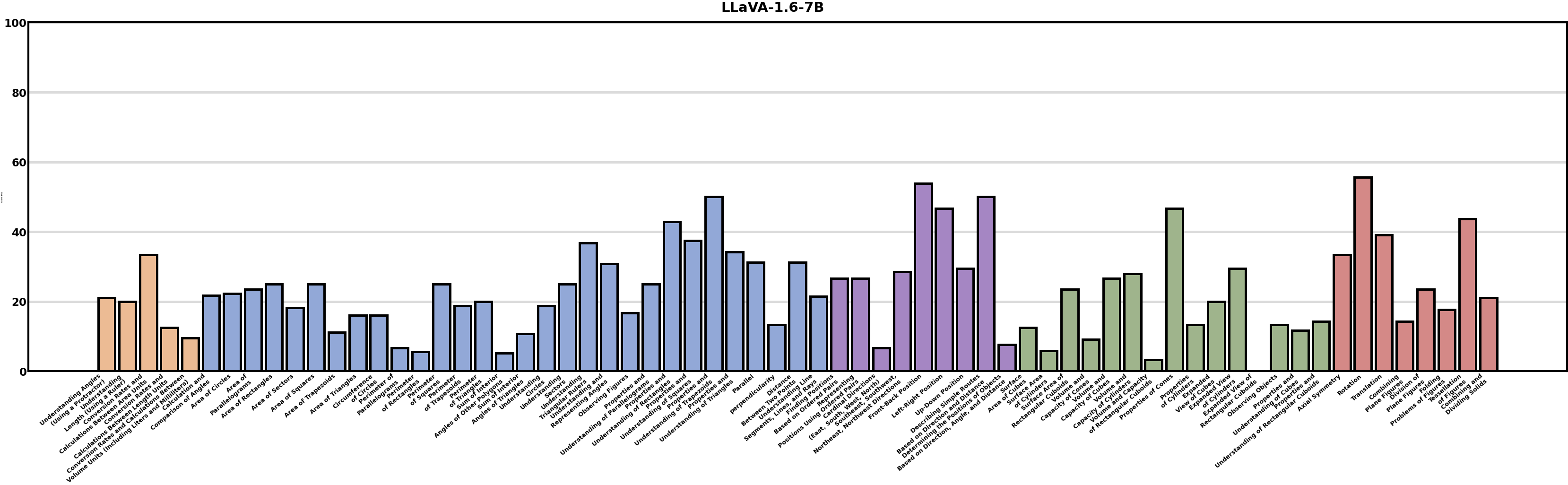}
    }
    \caption{Detailed performance of LLaVA-1.6-7B across 67 knowledge concepts.}
    \label{fig:LLaVA-1.6-7B}
\end{figure}

\begin{figure}[!t]
    \centering
    \resizebox{\textwidth}{!}{
    \includegraphics{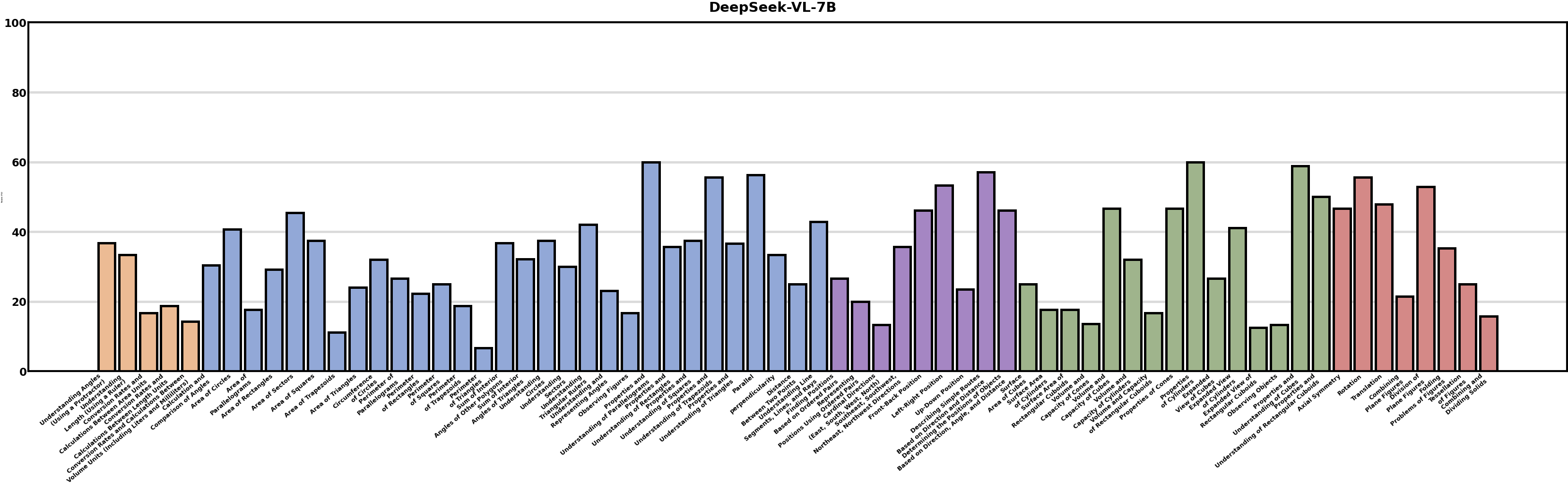}
    }
    \caption{Detailed performance of DeepSeek-VL-7B across 67 knowledge concepts.}
    \label{fig:DeepSeek-VL-7B}
\end{figure}

\begin{figure}[!t]
    \centering
    \resizebox{\textwidth}{!}{
    \includegraphics{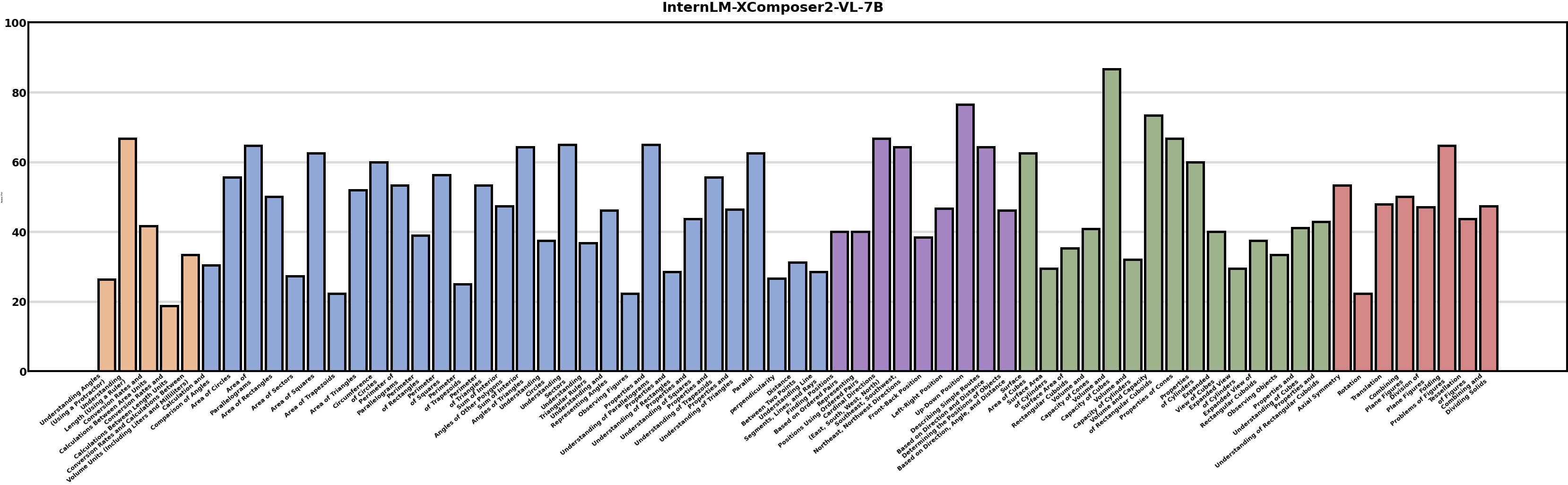}
    }
    \caption{Detailed performance of InternLM-XComposer2-VL-7B across 67 knowledge concepts.}
    \label{fig:InternLM-XComposer2-VL-7B}
\end{figure}

\begin{figure}[!t]
    \centering
    \resizebox{\textwidth}{!}{
    \includegraphics{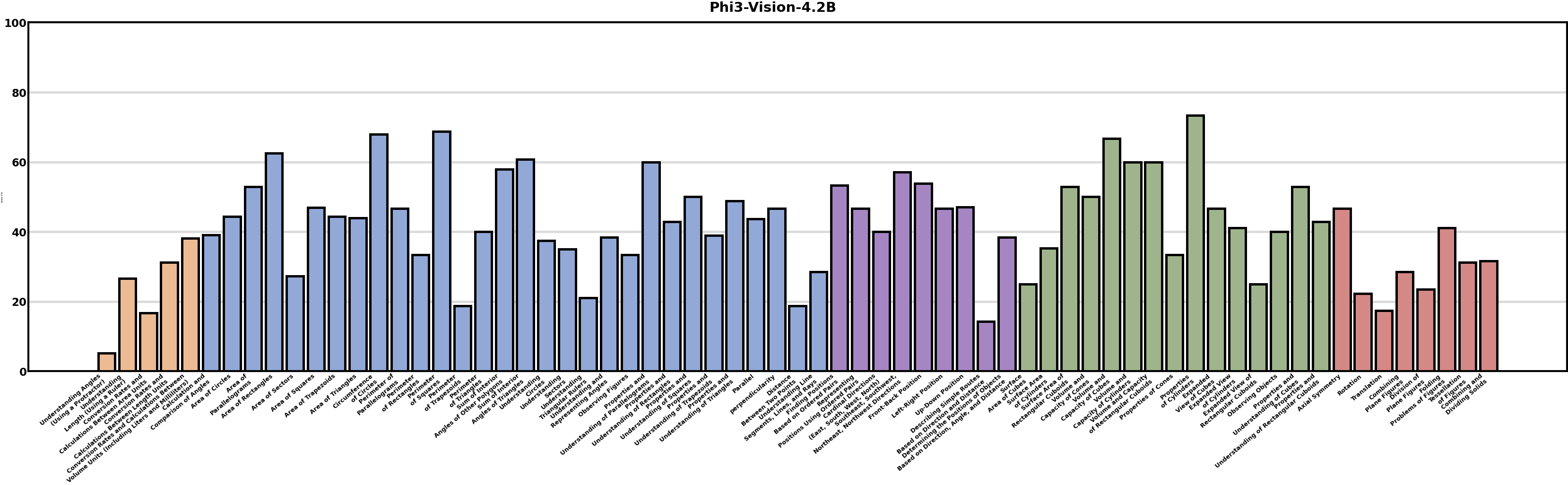}
    }
    \caption{Detailed performance of Phi3-Vision-4.2B across 67 knowledge concepts.}
    \label{fig:Phi3-Vision-4.2B}
\end{figure}

\begin{figure}[!t]
    \centering
    \resizebox{\textwidth}{!}{
    \includegraphics{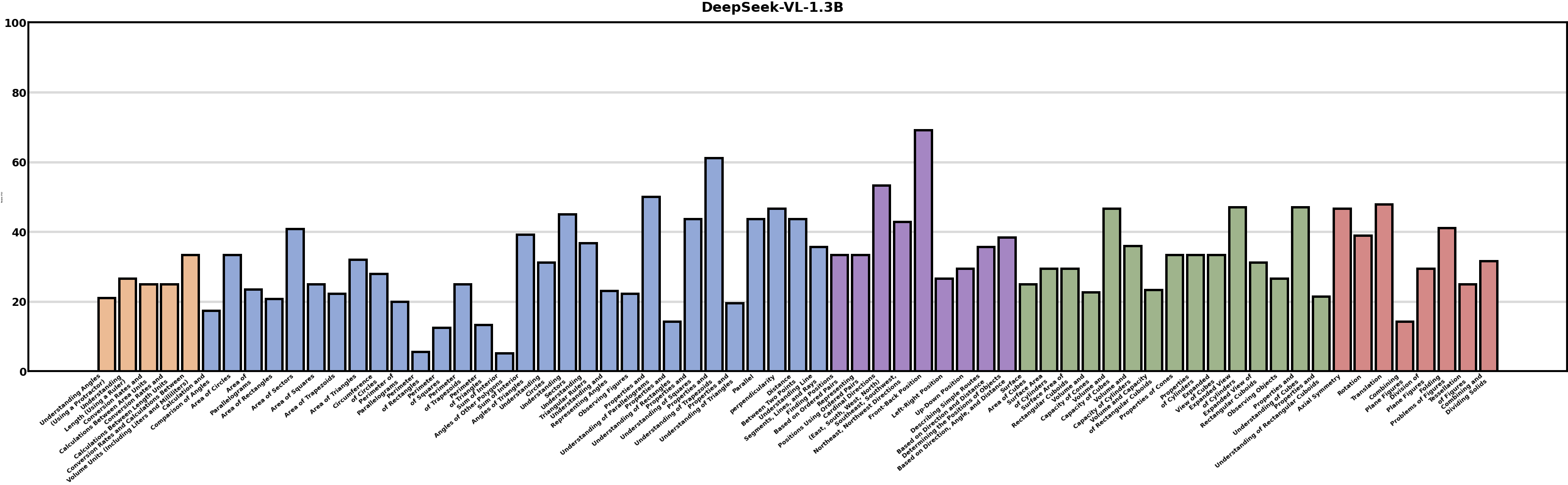}
    }
    \caption{Detailed performance of DeepSeek-VL-1.3B across 67 knowledge concepts.}
    \label{fig:DeepSeek-VL-1.3B}
\end{figure}

\textbf{Detailed Performance on Each Category.} In Figure~\ref{fig:leida-main}, we present the performance of open-source and closed-source models under the second-level nodes. In Figure~\ref{fig:GPT-4o} to Figure~\ref{fig:DeepSeek-VL-1.3B}, we detail the specific performance of 17 models across 67 knowledge concepts (based on statistics from one-step problem questions). It is evident that GPT-4o consistently leads in overall performance, but its main issue lies in measurement-related tasks. Notably, some open-source models perform worse on the simpler "Understanding and Conversion of Units" knowledge concepts compared to "Angles and Length" related concepts, while InternVL-Chat-V1.5 and MiniCPM-LLaMA3-V 2.5 exhibit more logically consistent results.



\clearpage

\subsection{Specific Error Analysis}
\label{subapp:Error Analysis}

\begin{table}[ht]
\caption{Detailed Descriptions of Error Types.}
\label{tab:Error_Types}
\begin{tabular}{c|l}
\toprule
\multicolumn{1}{c|}{\textbf{Error Type}}     & \multicolumn{1}{c}{\textbf{Explanation}}\\ 
\midrule
\begin{tabular}[c]{@{}c@{}}Knowledge Error\end{tabular}   & \begin{tabular}[c]{@{}l@{}} For a specific knowledge concept, the model is unclear or confused about \\ it, or it misuses another knowledge concept to solve the problem. \\
\end{tabular} \\
\midrule

\begin{tabular}[c]{@{}c@{}}Reason Error\end{tabular}   & \begin{tabular}[c]{@{}l@{}} Errors that occur in the logical reasoning process while using knowledge \\ concepts to solve the problem step by step.  \\
\end{tabular} \\
\midrule

\begin{tabular}[c]{@{}c@{}}Visual Error\end{tabular}   & \begin{tabular}[c]{@{}l@{}} Errors in visual perception, where the model incorrectly identifies shapes \\ or numbers in an image. \\
\end{tabular} \\
\midrule

\begin{tabular}[c]{@{}c@{}}Hallucination\end{tabular}   & \begin{tabular}[c]{@{}l@{}} The thought process introduces factors that are not consistent with the facts, \\ which are not mentioned in the context of the image or question. \\
\end{tabular} \\

\bottomrule
\end{tabular}
\end{table}

\begin{figure}[ht]
    \centering
    \resizebox{\textwidth}{!}{
    \includegraphics{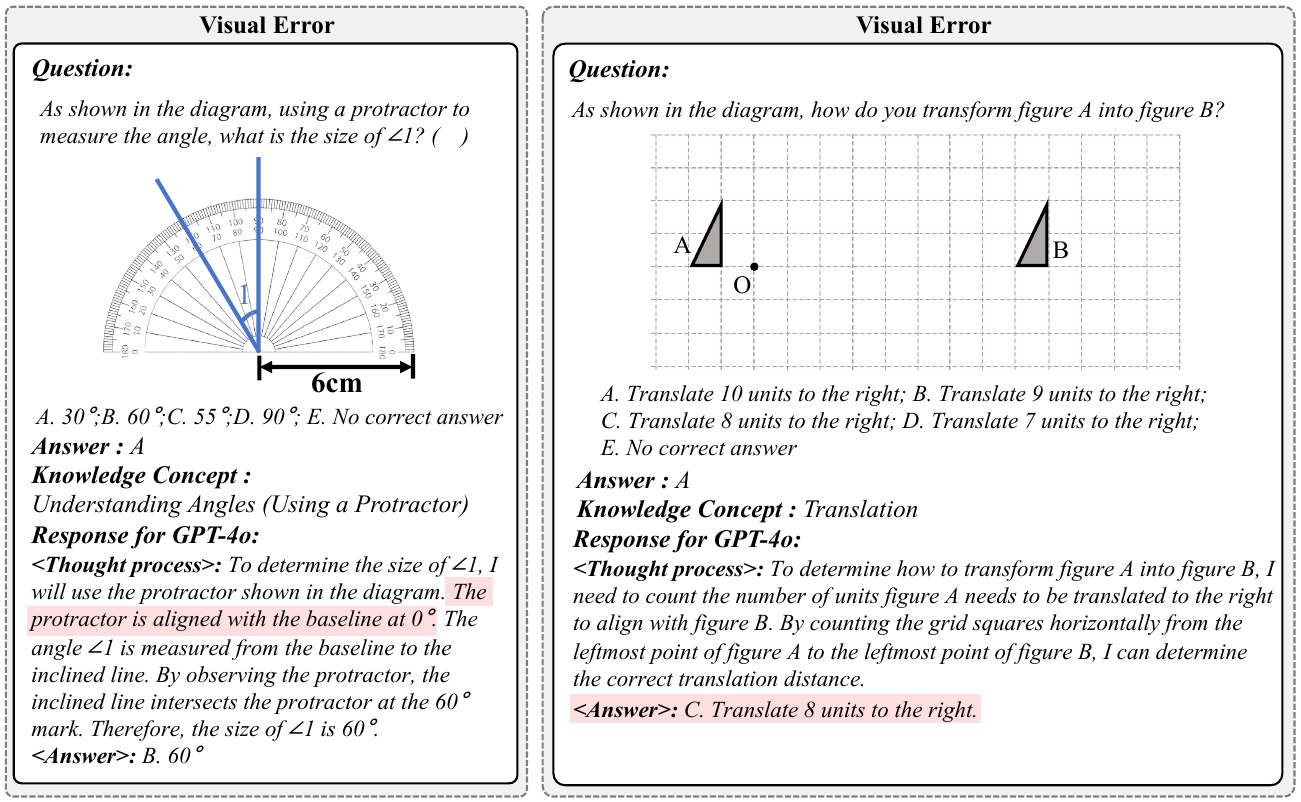}
    }
    \caption{Specific examples of Visual Error.}
    \label{fig:visual_error}
\end{figure}

\begin{figure}[!t]
    \centering
    \resizebox{\textwidth}{!}{
    \includegraphics{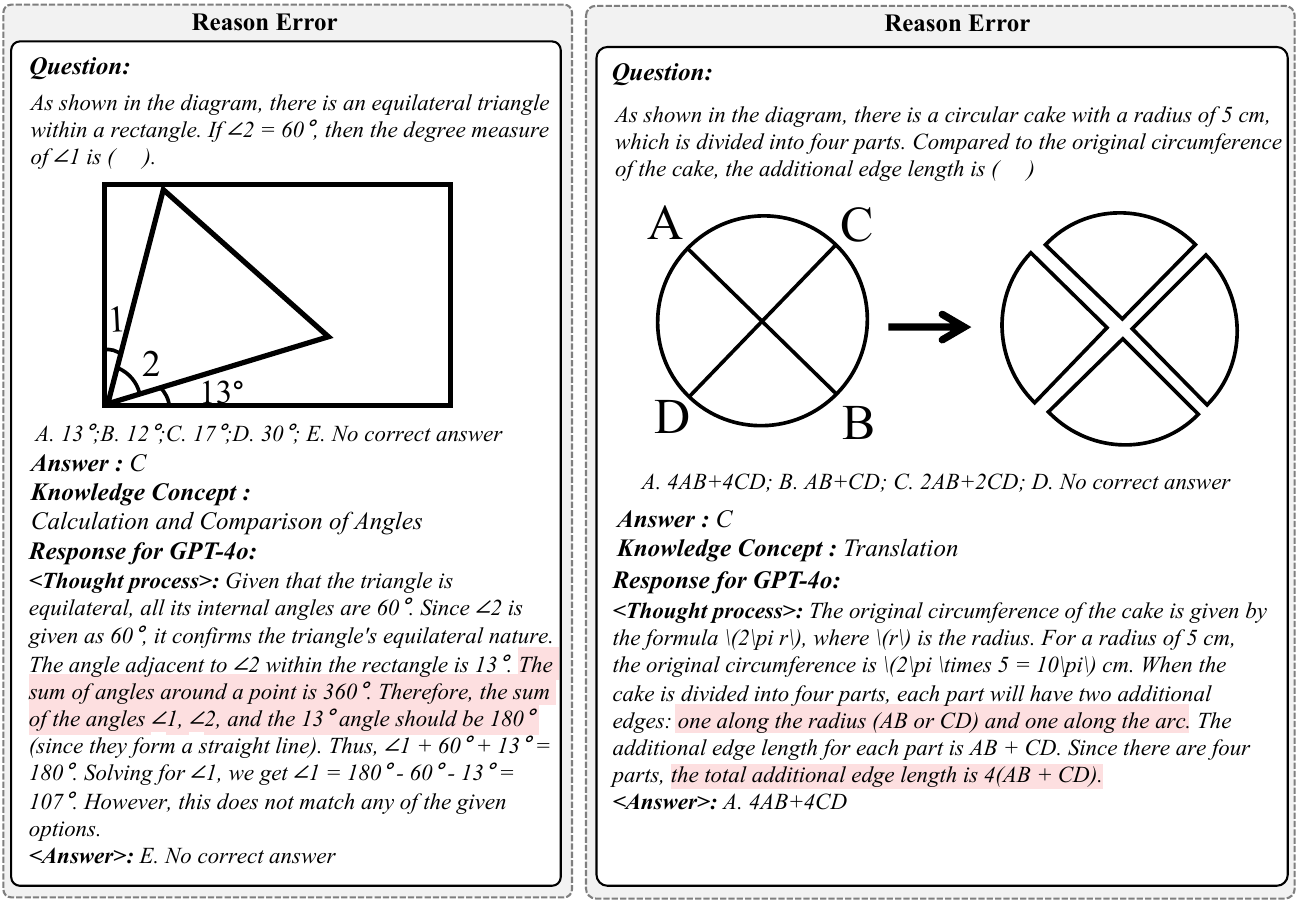}
    }
    \caption{Specific examples of Reason Error.}
    \label{fig:reason_error}
\end{figure}

\begin{figure}[!t]
    \centering
    \resizebox{\textwidth}{!}{
    \includegraphics{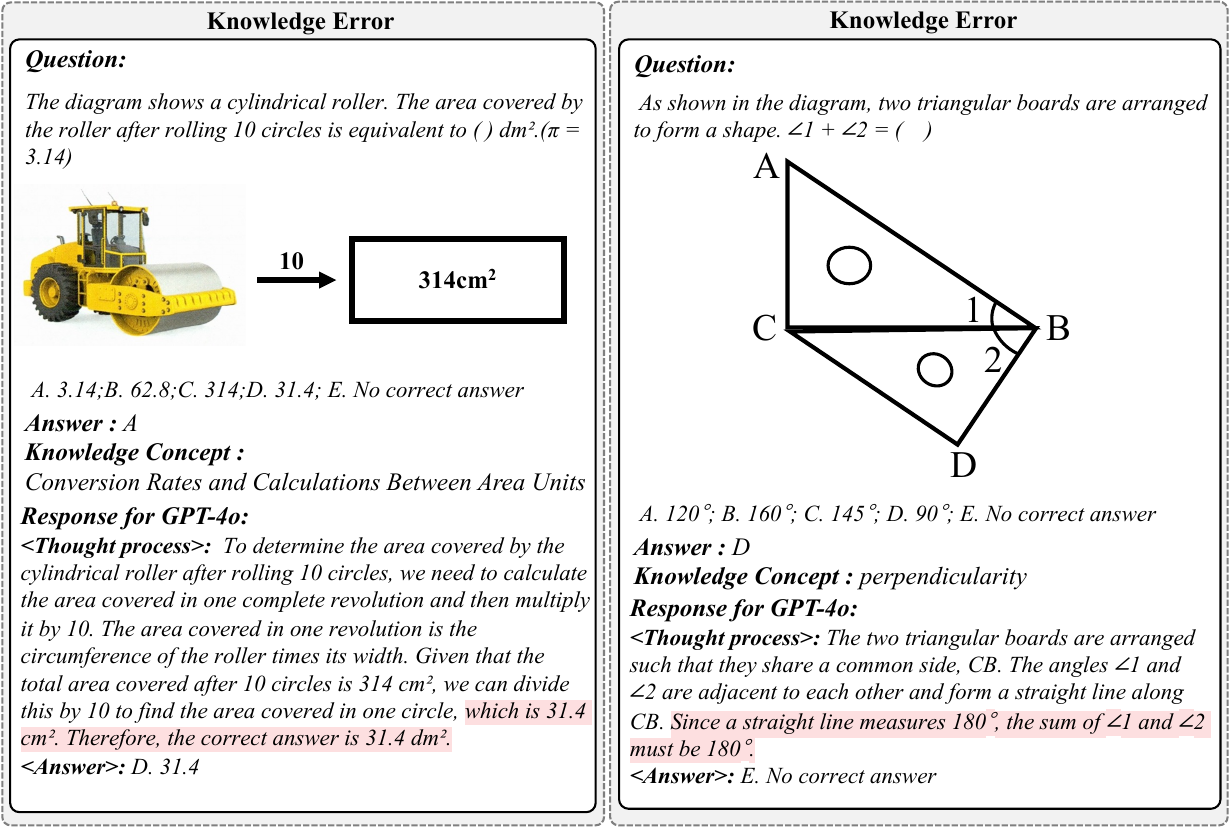}
    }
    \caption{Specific examples of Knowledge Error.}
    \label{fig:knowledge_error}
\end{figure}

\begin{figure}[!t]
    \centering
    \resizebox{\textwidth}{!}{
    \includegraphics{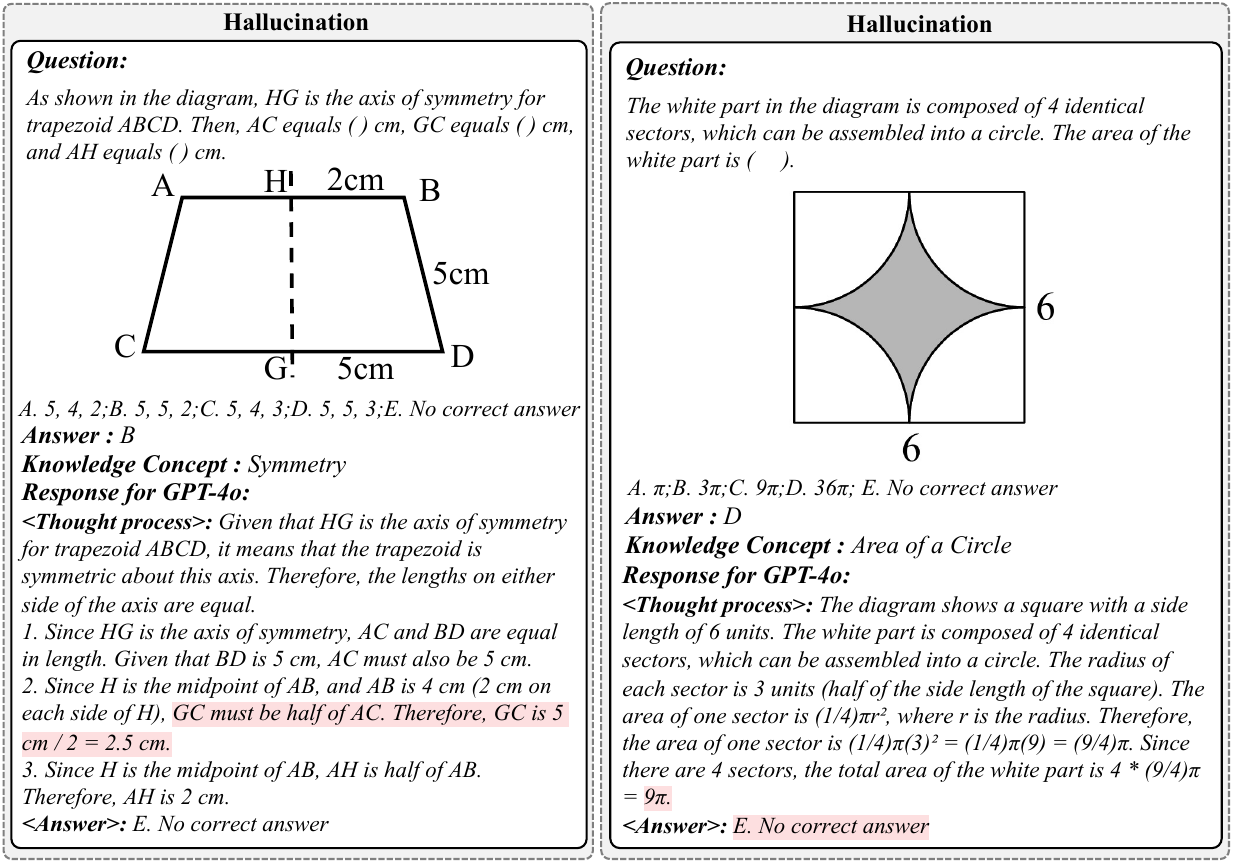}
    }
    \caption{Specific examples of Hallucination.}
    \label{fig:hallucination}
\end{figure}

\textbf{Error Types.} To delve into the failure cases of models, we detailed four typical error types in Table \ref{tab:Error_Types}. Furthermore, to facilitate a better understanding of each error type, we provide examples of each error made by GPT-4o from Figure~\ref{fig:visual_error} to Figure~\ref{fig:hallucination}. Since a single thought process in a problem can involve multiple errors and a single logical error is enough to derail a much larger solution, we consider the first error that occurs in the reasoning steps as the key error and include only this error in our statistics.

\textbf{Correspondence of Errors in Multi-Step and One-Step Problems.}
Focusing on Insufficient Knowledge, the errors in multi-step problems often correspond to those in one-step problems. This supports our approach of decomposing problems to accurately associate error types with specific knowledge concepts. Furthermore, we observe a positive correlation between the quantity of knowledge concepts and the errors in the reasoning process. As the complexity of knowledge concepts increases, the difficulty for the model to perform multi-step reasoning also increases, leading to a higher likelihood of visual recognition errors and incorrect application of knowledge concepts.

\clearpage

\section{Example Demonstration}
\label{examples}

\subsection{Description of the Knowledge Concepts}
\label{ex des}

\begin{figure}[!ht]
    \centering
    \resizebox{\textwidth}{!}{
    \includegraphics{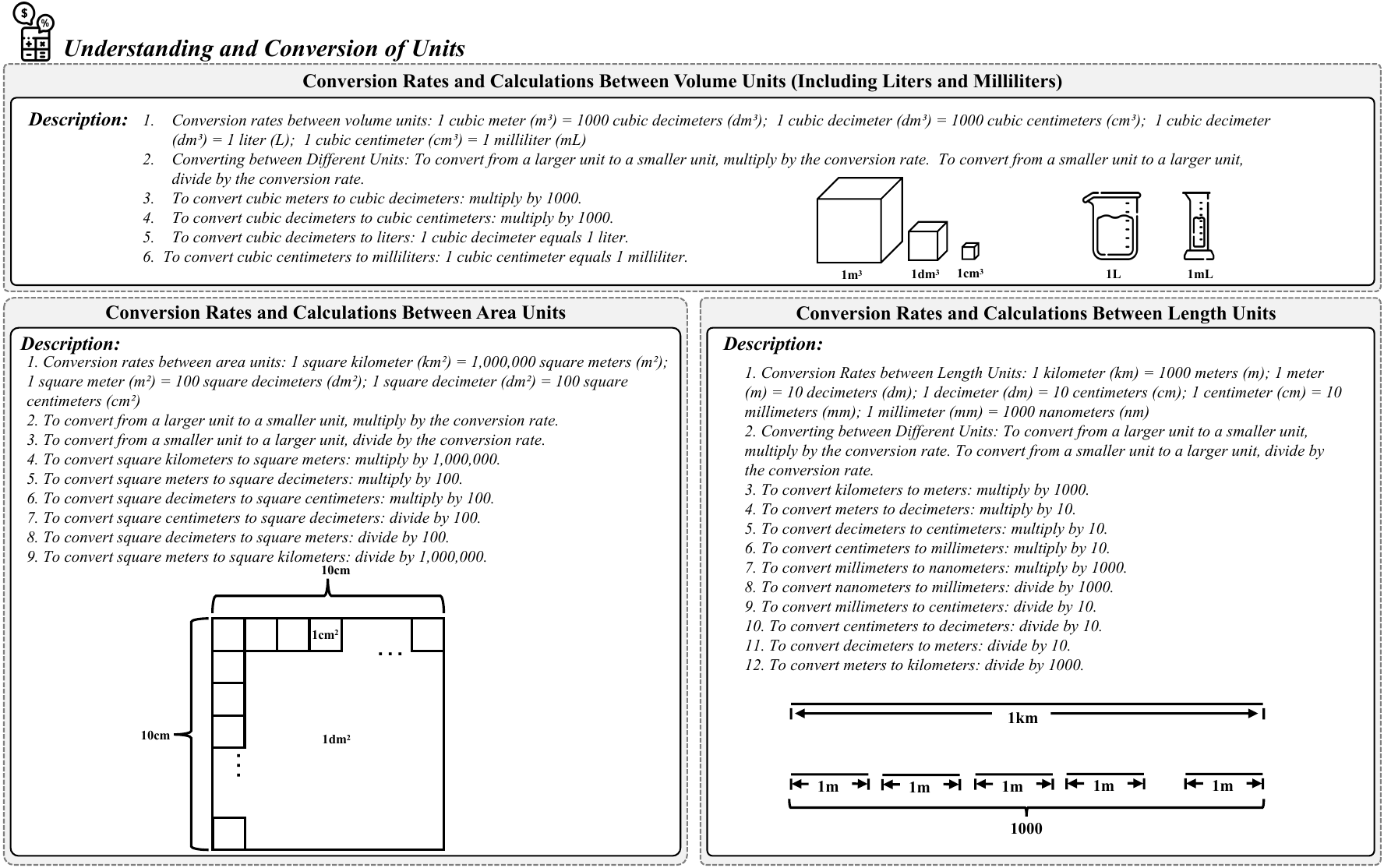}
    }
    \caption{The description of the knowledge concept "Understanding and Conversion of Units"}
    \label{fig:UCU}
\end{figure}

\begin{figure}[!ht]
    \centering
    \resizebox{\textwidth}{!}{
    \includegraphics{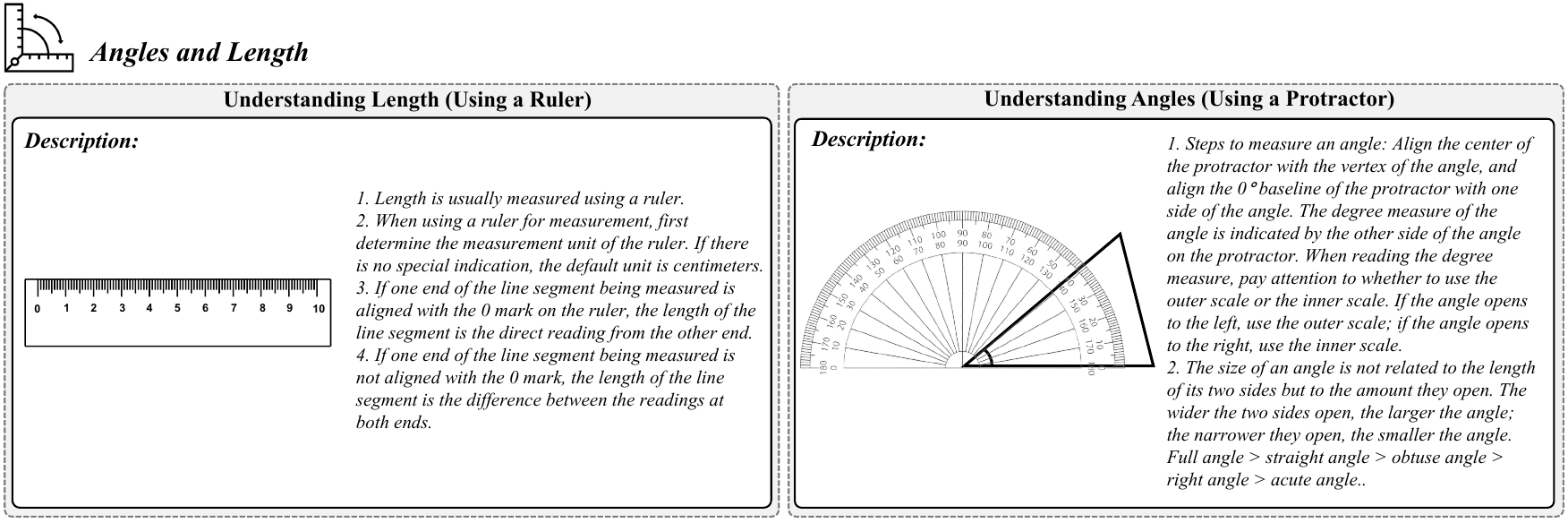}
    }
    \caption{The description of the knowledge concept "Angles and Length"}
    \label{fig:AL}
\end{figure}

\begin{figure}[!ht]
    \centering
    \resizebox{\textwidth}{!}{
    \includegraphics{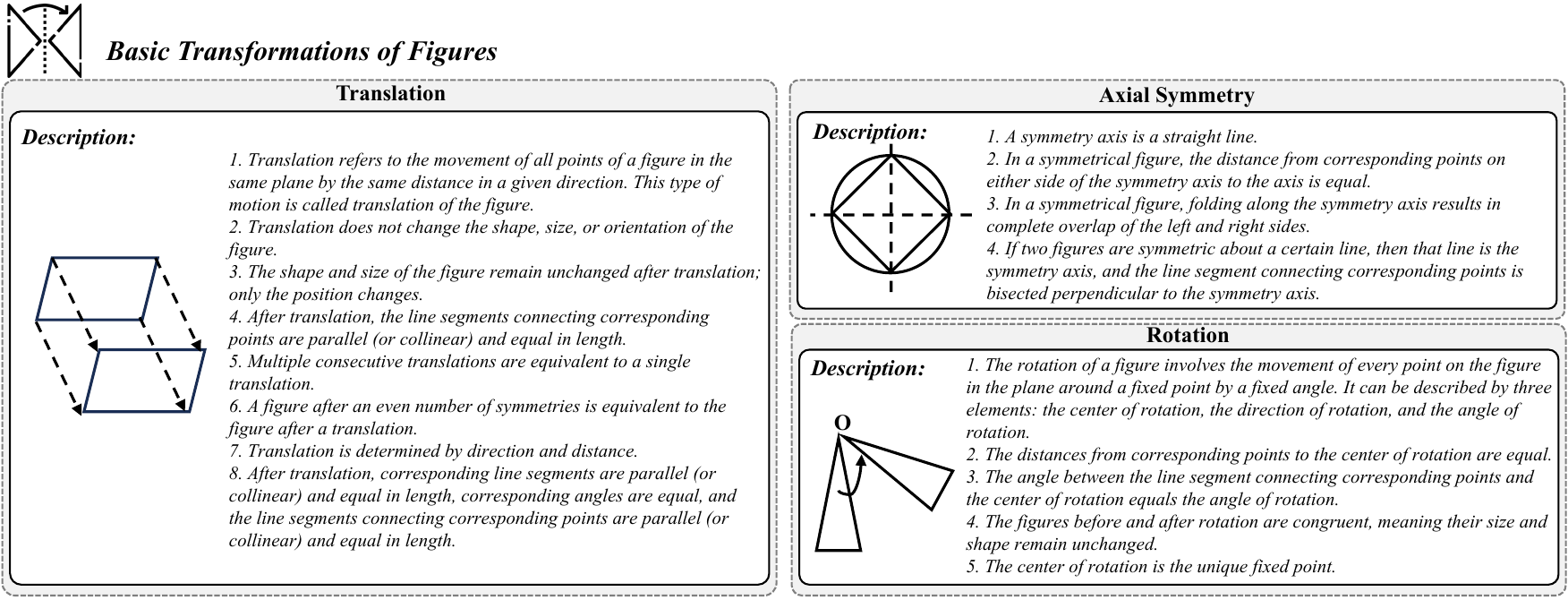}
    }
    \caption{The description of the knowledge concept "Basic Transformations of Figures"}
    \label{fig:BTF}
\end{figure}

\begin{figure}[!ht]
    \centering
    \resizebox{\textwidth}{!}{
    \includegraphics{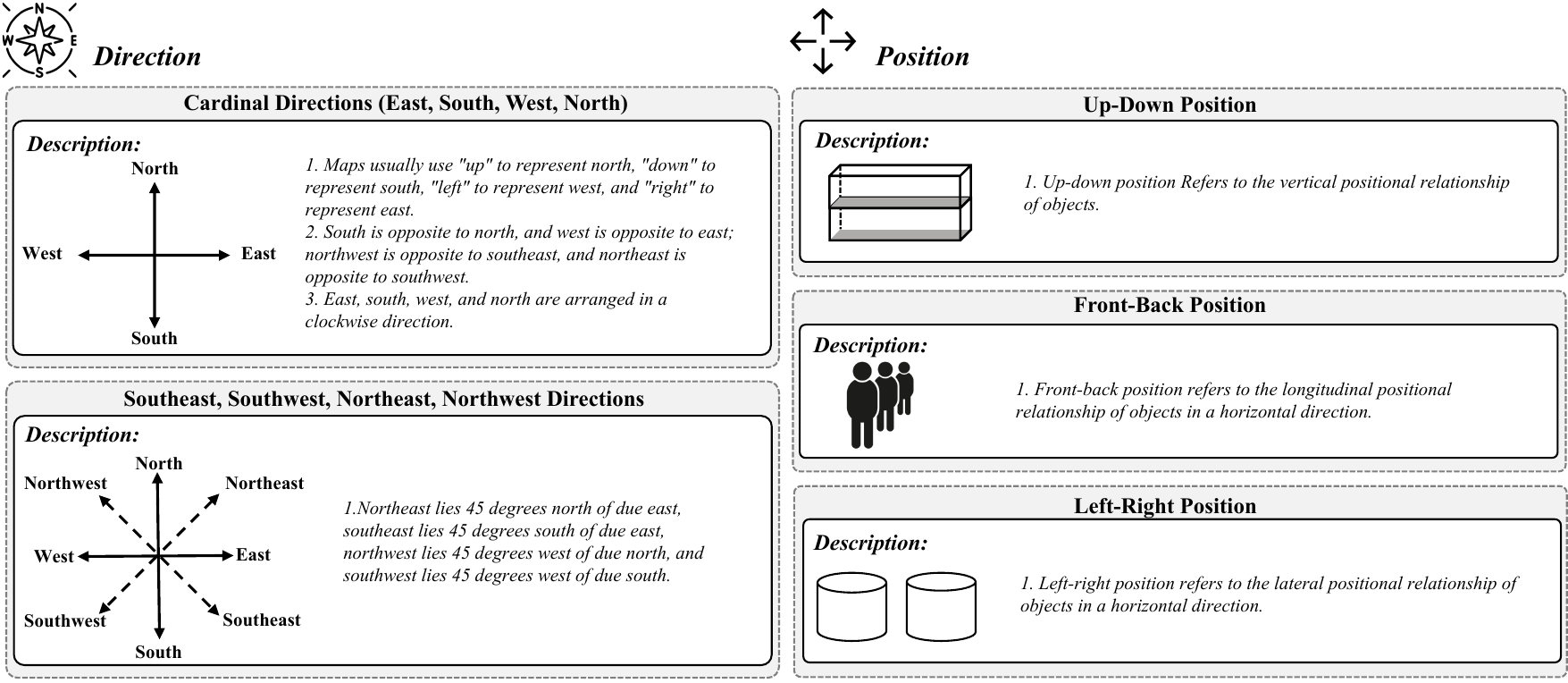}
    }
    \caption{The description of the knowledge concepts "Direction" and "Position"}
    \label{fig:Dir_Pos}
\end{figure}

\begin{figure}[!ht]
    \centering
    \resizebox{0.8\textwidth}{!}{
    \includegraphics{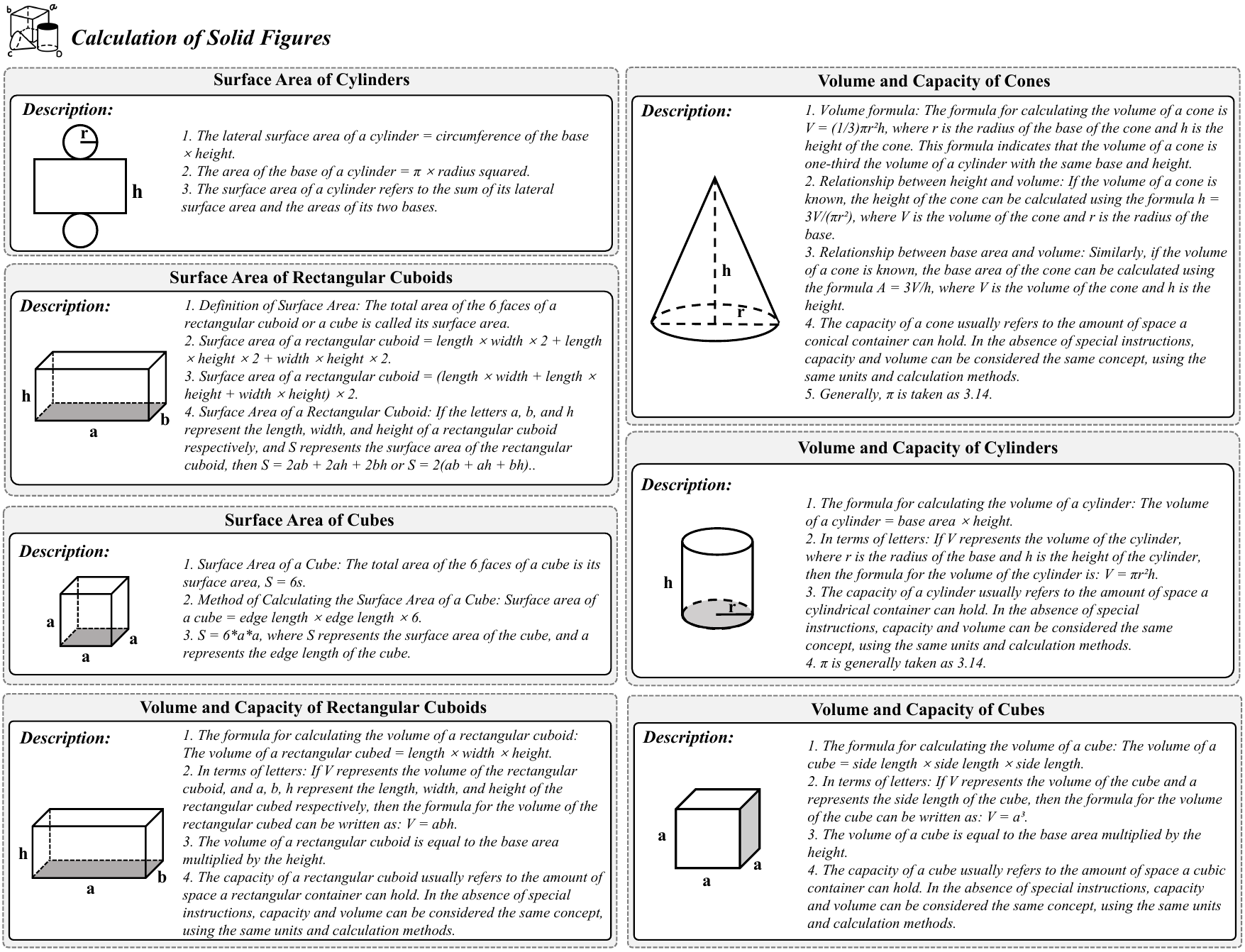}
    }
    \caption{The description of the knowledge concept "Calculation of Solid Figures"}
    \label{fig:CSF}
\end{figure}

\begin{figure}[!ht]
    \centering
    \resizebox{0.8\textwidth}{!}{
    \includegraphics{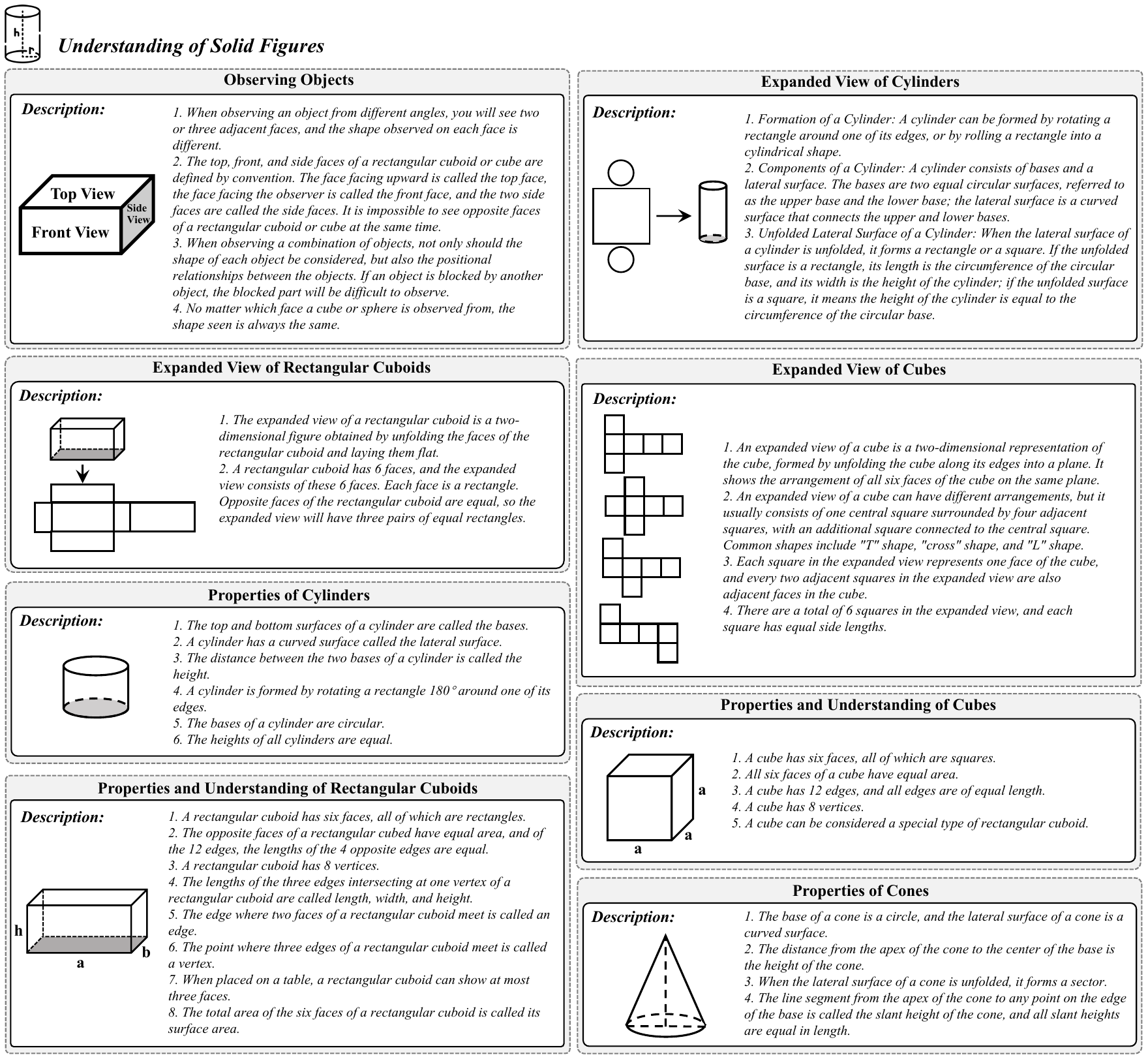}
    }
    \caption{The description of the knowledge concept "Understanding of Solid Figures"}
    \label{fig:USF}
\end{figure}

\begin{figure}[!ht]
    \centering
    \resizebox{\textwidth}{!}{
    \includegraphics{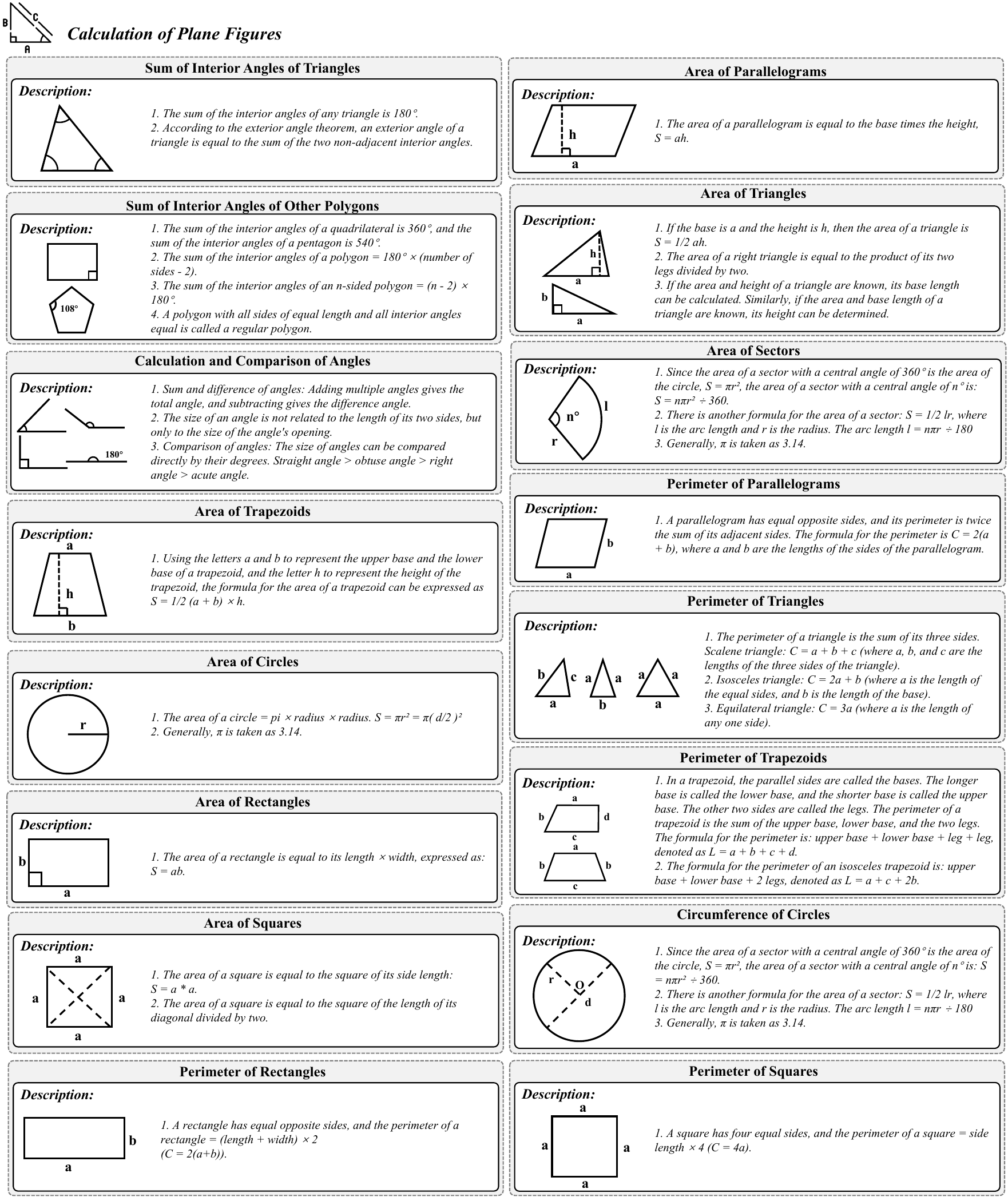}
    }
    \caption{The description of the knowledge concept "Calculation of Plane Figures"}
    \label{fig:CPF}
\end{figure}

\begin{figure}[!ht]
    \centering
    \resizebox{\textwidth}{!}{
    \includegraphics{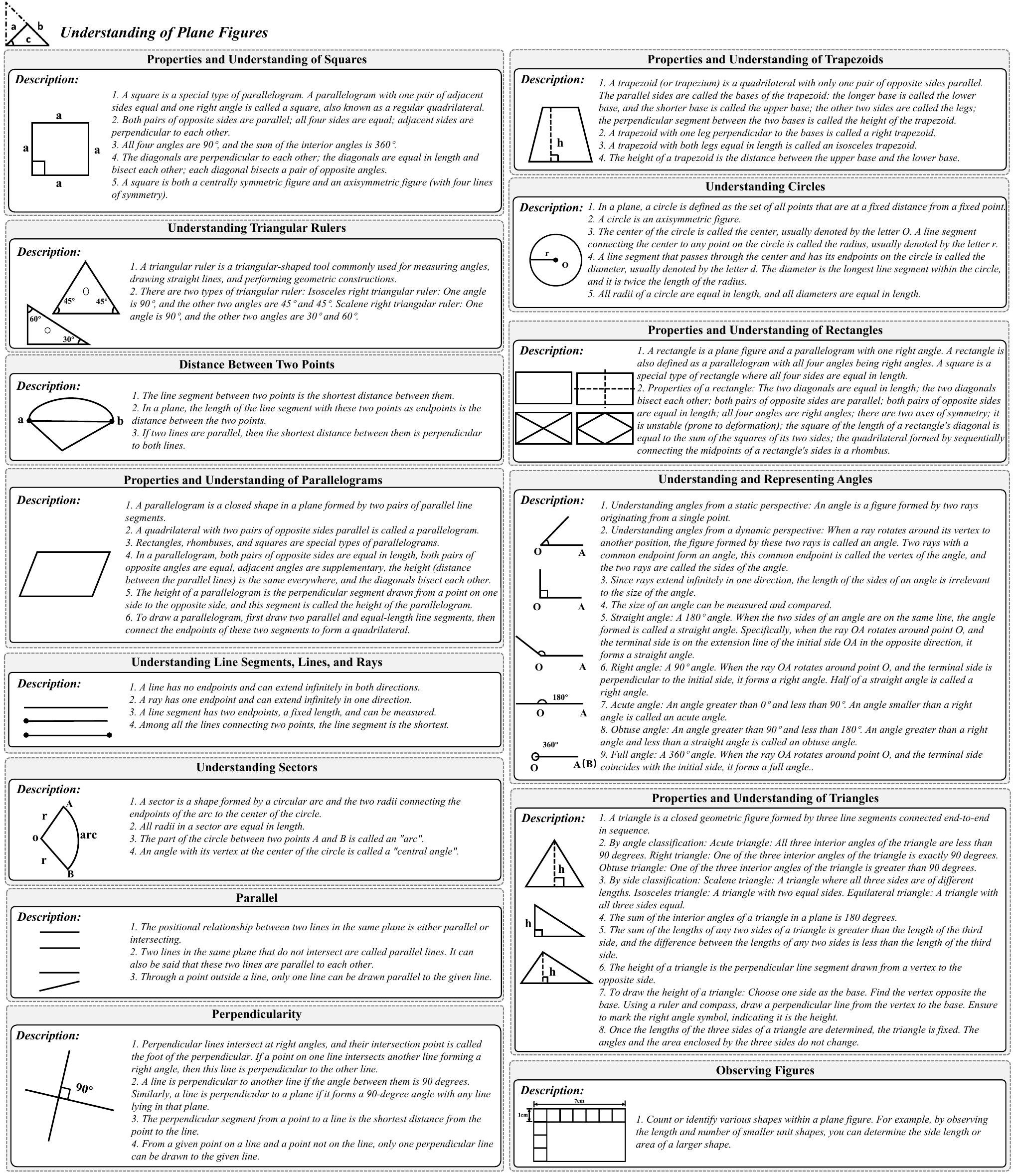}
    }
    \caption{The description of the knowledge concept "Understanding of Plane Figures"}
    \label{fig:UPF}
\end{figure}

\begin{figure}[!ht]
    \centering
    \resizebox{0.9\textwidth}{!}{
    \includegraphics{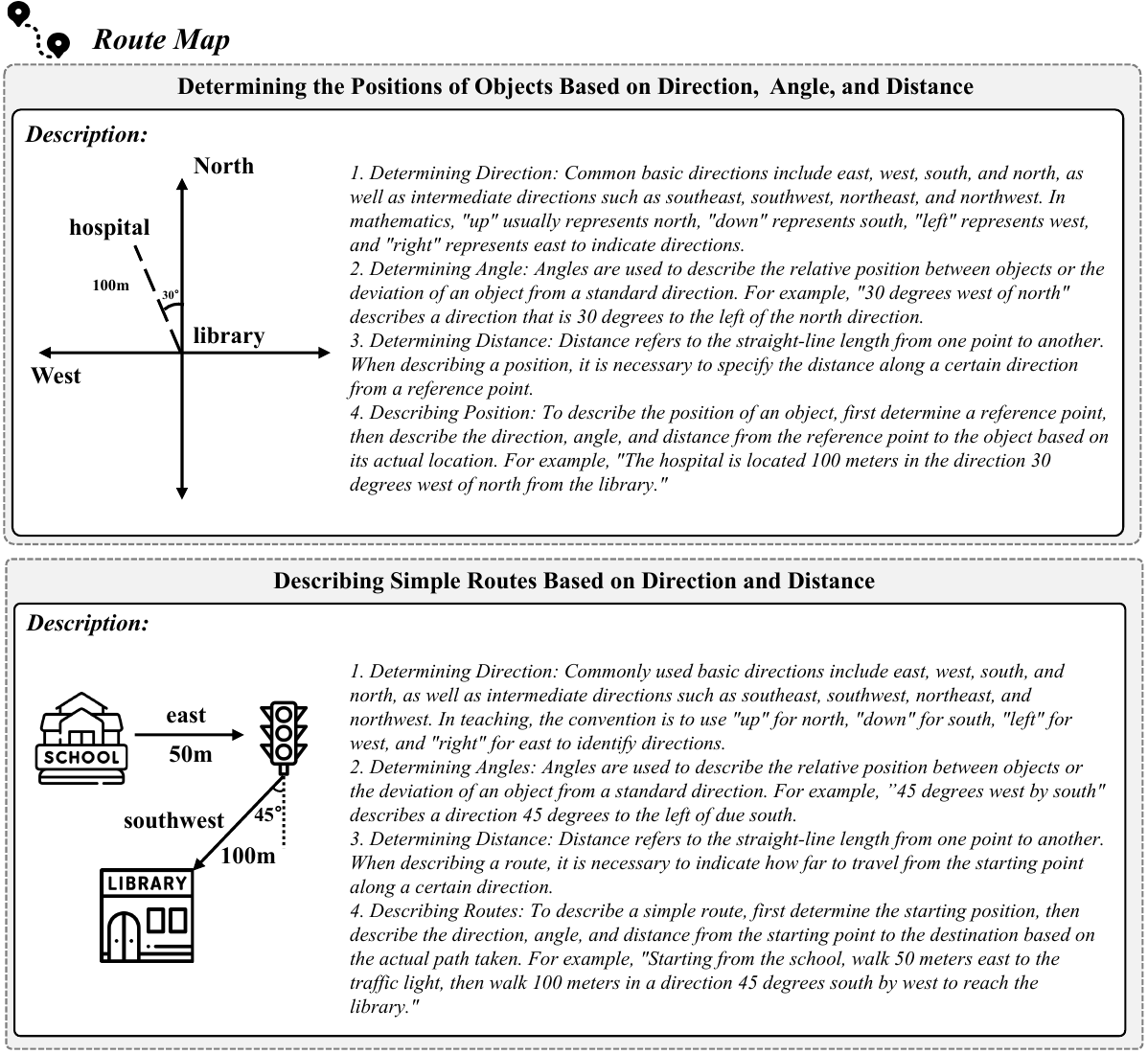}
    }
    \caption{The description of the knowledge concept "Route Map"}
    \label{fig:Rom}
\end{figure}

\begin{figure}[!ht]
    \centering
    \resizebox{0.9\textwidth}{!}{
    \includegraphics{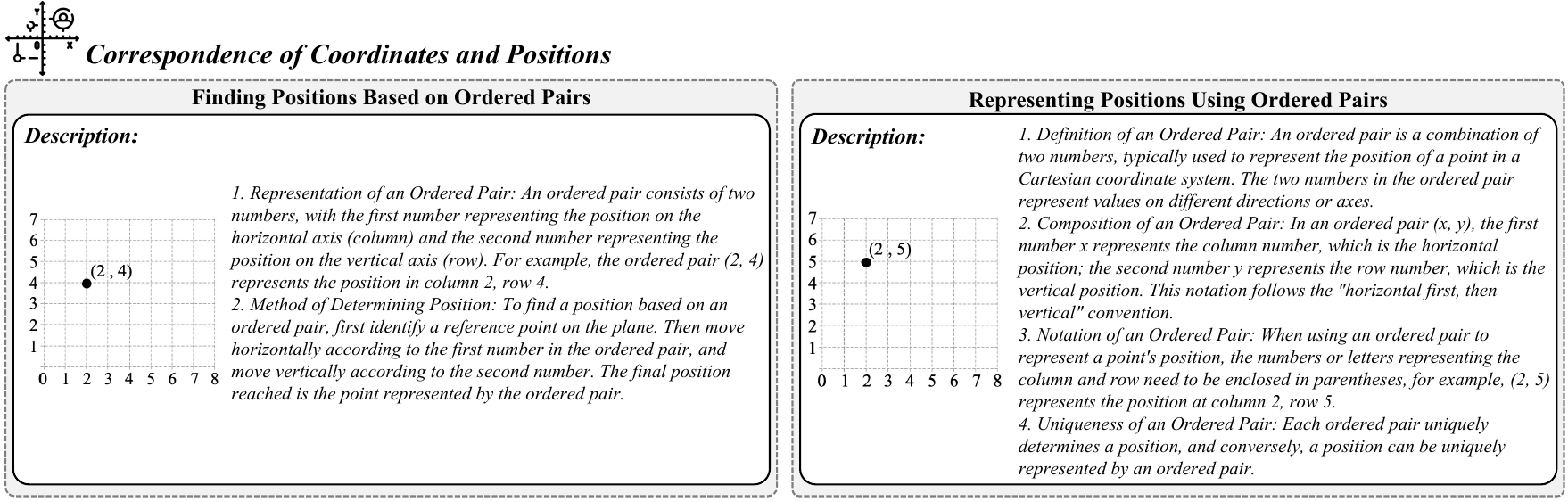}
    }
    \caption{The description of the knowledge concept "Correspondence of Coordinates and Positions"}
    \label{fig:CCP}
\end{figure}

Figure \ref{fig:kn card} and Figure~\ref{fig:UCU} to~\ref{fig:CCP} illustrate the the detailed information of knowledge concepts.

\clearpage

\subsection{Data Sources of \dataset}
\label{data sorce}
Table~\ref{tab:source1} to~\ref{tab:source5} illustrate the the detailed data source lists of \dataset.

\begin{table}[!ht]
 \caption{The data sources of \dataset (Part1, Source 1 to 50).}
 \label{tab:source1}
    \centering
    \tiny
       \begin{tabular}{|c|p{0.9\textwidth}|} 
       
    \toprule
    \textbf{Number} & \textbf{Data Source} \\ 
   \hline
        1 & [Beijing Chaoyang New Target Detection] 2022 Edition of People's Education Edition Grade 3 Volume 1: Mathematics \\ \hline
        2 & [Beijing New Target Detection] 2023 Printing People's Education Edition Grade 3 Volume 1: Mathematics \\ \hline
        3 & [Beijing Chaoyang New Target Detection] 2023 Printing People's Education Edition Grade 3 Volume 2: Mathematics \\ \hline
        4 & [Learning Objectives and Assessment] 2023-2024 Academic Year People's Education Edition Grade 3 Volume 2: Mathematics \\ \hline
        5 & [Beijing Chaoyang New Target Detection] 2020 Edition of the People's Education Edition Grade 4 Volume 1: Mathematics \\ \hline
        6 & [New Target Detection] 2022 Edition of the People's Education Press Grade 4 Volume 2: Mathematics \\ \hline
        7 & [Beijing Chaoyang New Target Detection] 2022 Edition of People's Education Edition Grade 4 Volume 1: Mathematics \\ \hline
        8 & [Beijing Chaoyang New Target Detection] 2023 Printing People's Education Edition Grade 4 Volume 2: Mathematics \\ \hline
        9 & [Beijing New Target Detection] 2023 Printing People's Education Edition Grade 4 Volume 1: Mathematics \\ \hline
        10 & [Learning Objectives and Assessment] 2023-2024 Academic Year People's Education Press Grade 4 Volume 2: Mathematics \\ \hline
        11 & [Beijing Chaoyang New Target Detection] 2021 Edition of the People's Education Edition Grade 5 Volume 1: Mathematics \\ \hline
        12 & [New Target Detection] 2022 Edition of the People's Education Edition Grade 5 Volume 2: Mathematics \\ \hline
        13 & [Beijing Chaoyang New Target Detection] 2022 Edition of the People's Education Edition Grade 5 Volume 1: Mathematics \\ \hline
        14 & [Beijing Chaoyang New Target Detection] 2023 Printing People's Education Edition Grade 5 Volume 2: Mathematics \\ \hline
        15 & [Beijing New Target Detection] 2023 Printing People's Education Edition Grade 5 Volume 1: Mathematics \\ \hline
        16 & [Learning Objectives and Tests] 2023-2024 Academic Year People's Education Edition Grade 5 Volume 2: Mathematics \\ \hline
        17 & [Beijing Chaoyang New Target Detection] 2023 Printing People's Education Edition Sixth Grade Volume 2: Mathematics \\ \hline
        18 & [New Target Detection] 2022 Edition of the People's Education Press Grade 6 Volume 2: Mathematics \\ \hline
        19 & [Beijing Chaoyang New Target Detection] 2022 Edition of People's Education Edition Grade 6 Volume 1: Mathematics \\ \hline
        20 & [Beijing Chaoyang New Target Detection] 2023 Edition of the People's Education Press Sixth Grade Volume 1: Mathematics \\ \hline
        21 & [Learning Objectives and Tests] 2023-2024 Academic Year Sixth Grade People's Education Press Volume 2: Mathematics \\ \hline
        22 & [Beijing Xicheng Learning Exploration Diagnosis] 2022 Edition of People's Education Edition Grade 3 Volume 1: Mathematics \\ \hline
        23 & [Beijing Learning Inquiry Diagnosis] 2023 Printing People's Education Press Grade 3 Volume 2: Mathematics \\ \hline
        24 & [Beijing Learning Inquiry Diagnosis] 2023 Printing People's Education Press Grade 3 Volume 1: Mathematics \\ \hline
        25 & [Learning Exploration Diagnosis] 2023-2024 Academic Year People's Education Edition Grade 3 Volume 2: Mathematics \\ \hline
        26 & [Learning Exploration Diagnosis] 2022 Edition of the People's Education Press Grade 4 Volume 2: Mathematics \\ \hline
        27 & [Beijing Xicheng Learning Exploration Diagnosis] 2022 Edition of People's Education Edition Grade 4 Volume 1: Mathematics \\ \hline
        28 & [Beijing Learning Inquiry Diagnosis] 2023 Printing People's Education Press Grade 4 Volume 2: Mathematics \\ \hline
        29 & [Beijing Learning Inquiry Diagnosis] 2023 Printing People's Education Press Grade 4 Volume 1: Mathematics \\ \hline
        30 & [Learning Exploration Diagnosis] 2023-2024 Academic Year People's Education Edition Grade 4 Volume 2: Mathematics \\ \hline
        31 & [Learning Exploration Diagnosis] 2022 Edition of the People's Education Press Grade 5 Volume 2: Mathematics \\ \hline
        32 & [Beijing Xicheng Learning Exploration Diagnosis] 2022 Edition of the People's Education Press Fifth Grade Volume 1: Mathematics \\ \hline
        33 & [Beijing Learning Inquiry Diagnosis] 2023 Printing People's Education Press Grade 5 Volume 2: Mathematics \\ \hline
        34 & [Beijing Learning Inquiry Diagnosis] 2023 Printing People's Education Press Grade 5 Volume 1: Mathematics \\ \hline
        35 & [Learning Exploration Diagnosis] 2023-2024 Academic Year People's Education Edition Grade 5 Volume 2: Mathematics \\ \hline
        36 & [Learning Exploration Diagnosis] 2022 Edition of the People's Education Press Grade 6 Volume 2: Mathematics \\ \hline
        37 & [Beijing Learning Inquiry Diagnosis] 2022 Edition of People's Education Press Grade 6 Volume 1: Mathematics \\ \hline
        38 & [Beijing Learning Exploration Diagnosis] 2023 Printing People's Education Press Grade 6 Volume 2: Mathematics \\ \hline
        39 & [Beijing Learning Inquiry Diagnosis] 2023 Printing People's Education Press Sixth Grade Volume 1: Mathematics \\ \hline
        40 & [Learning Exploration Diagnosis] 2023-2024 Academic Year People's Education Edition Sixth Grade Volume 2: Mathematics \\ \hline
        41 & [Beijing Haidian famous teachers accompany you to study and practice synchronously] 2022 Beijing Normal University Edition Grade 3 Volume 1: Mathematics \\ \hline
        42 & [Beijing Haidian famous teachers accompany you to study and practice synchronously] 2022 print Beijing Normal University edition Grade 3, Volume 2: Mathematics \\ \hline
        43 & [Beijing Companion Learning Synchronous Learning Handbook] 2023 Printing Beijing Normal University Edition Grade 3 Volume 1: Mathematics \\ \hline
        44 & [Synchronous Learning Handbook for You] 2023-2024 Academic Year Beijing Normal University Edition Grade 3 Volume 2: Mathematics \\ \hline
        45 & [Famous teachers from Haidian accompany you to study, practice and test] 2022 Beijing Normal University Grade 4 Volume 2: Mathematics \\ \hline
        46 & [Beijing Haidian famous teachers accompany you to study and practice synchronously] 2022 Beijing Normal University Edition Grade 4 Volume 1: Mathematics \\ \hline
        47 & [Beijing Haidian famous teachers accompany you to study and practice synchronously] 2022 Beijing Normal University Edition Grade 4 Volume 2: Mathematics \\ \hline
        48 & [Beijing Companion Learning Synchronous Learning Handbook] 2023 Printing Beijing Normal University Edition Grade 4 Volume 1: Mathematics \\ \hline
        49 & [Synchronous Learning Handbook for You] 2023-2024 Academic Year Beijing Normal University Edition Grade 4 Volume 2: Mathematics \\ \hline
        50 & [Haidian famous teachers accompany you to study synchronous learning and practice book] 2021 edition of fifth grade volume 1: Mathematics \\ \hline

    \end{tabular}
\end{table}

\begin{table}[!ht]
    \centering
    \tiny
    \caption{The data sources of \dataset (Part2, Source 51 to 100).}
       \begin{tabular}{|c|p{0.9\textwidth}|} 
    \toprule
    \textbf{Number} & \textbf{Data Source} \\ 
    \hline
51 & [Haidian famous teachers accompany you to study and practice synchronously] 2022 edition of Beijing Normal University Grade 5 Volume 2: Mathematics \\ \hline
        52 & [Beijing Haidian famous teachers accompany you to study synchronously] 2022 Beijing Normal University Edition Grade 5 Volume 2: Mathematics \\ \hline
        53 & [Beijing Companion Learning Synchronous Learning Handbook] 2023 Printing Beijing Normal University Edition Grade 5 Volume 1: Mathematics \\ \hline
        54 & [Synchronous Learning Handbook for You] 2023-2024 Beijing Normal University Edition Grade 5 Volume 2: Mathematics \\ \hline
        55 & [Haidian famous teachers accompany you to study synchronous learning and practice book] 2021 edition of sixth grade volume: Mathematics \\ \hline
        56 & [Haidian famous teachers accompany you to study and practice synchronously] 2022 Beijing Normal University Grade 6 Volume 2: Mathematics \\ \hline
        57 & [Beijing Haidian Famous Teacher] 2022 Beijing Normal University Edition Sixth Grade Volume 1: Mathematics \\ \hline
        58 & [Beijing Haidian famous teachers accompany you to study and practice synchronously] 2022 Beijing Normal University Edition Grade 6 Volume 2: Mathematics \\ \hline
        59 & [Beijing Companion Learning Synchronous Learning Handbook] 2023 Printing Beijing Normal University Edition Sixth Grade Volume 1: Mathematics \\ \hline
        60 & [Synchronous Learning Handbook for You] 2023-2024 Academic Year Beijing Normal University Edition Sixth Grade Volume 2: Mathematics \\ \hline
        61 & [Beijing Dongcheng Formative Independent Evaluation] 2022 Edition of People's Education Edition Grade 3 Volume 1: Mathematics \\ \hline
        62 & [Formative Self-Evaluation] 2023-2024 Academic Year Grade 3, Volume 2: Mathematics \\ \hline
        63 & [Beijing Formative Independent Assessment] 2023 Printing People's Education Press Grade 3 Volume 1: Mathematics \\ \hline
        64 & [Beijing Formative Independent Assessment] 2023 Printing People's Education Press Grade 3 Volume 1: Mathematics \\ \hline
        65 & [Beijing Dongcheng Formative Independent Evaluation] 2022 Edition of the People's Education Press Grade 4 Volume 1: Mathematics \\ \hline
        66 & [Formative Self-Evaluation] 2023-2024 Academic Year People's Education Edition Grade 4 Volume 2: Mathematics \\ \hline
        67 & [Beijing Formative Independent Assessment] 2023 Printing People's Education Press Grade 4 Volume 1: Mathematics \\ \hline
        68 & [Formative Self-Evaluation] 2022 Edition of the People's Education Press Grade 5 Volume 2: Mathematics \\ \hline
        69 & [Beijing Dongcheng Formative Independent Evaluation] 2022 Edition of the People's Education Press Fifth Grade Volume 1: Mathematics \\ \hline
        70 & [Formative Self-Evaluation] 2023-2024 Academic Year People's Education Edition Grade 5 Volume 2: Mathematics \\ \hline
        71 & [Beijing Formative Independent Assessment] 2023 Printing People's Education Edition Grade 5 Volume 1: Mathematics \\ \hline
        72 & [Formative Self-Evaluation] 2022 Edition of the People's Education Press Grade 6 Volume 2: Mathematics \\ \hline
        73 & [Beijing Dongcheng Formative Independent Evaluation] 2022 Edition of the People's Education Press Sixth Grade Volume 1: Mathematics \\ \hline
        74 & [Formative Self-Evaluation] 2023-2024 Academic Year Sixth Grade People's Education Edition Volume 2: Mathematics \\ \hline
        75 & [Beijing Formative Independent Assessment] 2023 Printing People's Education Press Sixth Grade Volume 1: Mathematics \\ \hline
        76 & [All-in-one study, practice and examination] 2023-2024 school year Beijing edition Grade 3, Volume 2: Mathematics \\ \hline
        77 & [All-in-one study, practice and examination] 2023-2024 school year People's Education Press Grade 3 Volume 2: Mathematics \\ \hline
        78 & [Beijing all-in-one study, practice and examination] 2023 Printing People's Education Press Grade 3 Volume 1: Mathematics \\ \hline
        79 & [Beijing all-in-one study, practice and examination] 2023 Beijing Edition Grade 3 Volume 2: Mathematics \\ \hline
        80 & [Beijing all-in-one study, practice and examination] 2022 Beijing Edition Grade 3 Volume 2: Mathematics \\ \hline
        81 & [Beijing all-in-one study, practice and examination] 2022 Edition of People's Education Press Grade 3 Volume 2: Mathematics \\ \hline
        82 & [Beijing all-in-one study, practice and examination] 2022 Beijing Normal University Edition Grade 3 Volume 1: Mathematics \\ \hline
        83 & [All-in-one study, practice and examination] 2023-2024 school year Beijing Edition Grade 4 Volume 2: Mathematics \\ \hline
        84 & [All-in-one study, practice and examination] 2023-2024 school year People's Education Press Grade 4 Volume 2: Mathematics \\ \hline
        85 & [Beijing all-in-one study, practice and examination] 2023 Beijing Edition Grade 4 Volume 1: Mathematics \\ \hline
        86 & [Beijing all-in-one study, practice and examination] 2023 Printing People's Education Press Grade 4 Volume 1: Mathematics \\ \hline
        87 & [Beijing all-in-one study, practice and examination] 2022 Beijing Edition Grade 4 Volume 2: Mathematics \\ \hline
        88 & [Beijing all-in-one study, practice and examination] 2022 Edition of People's Education Press Grade 4 Volume 2: Mathematics \\ \hline
        89 & [Beijing all-in-one study, practice and examination] 2022 Beijing Normal University Edition Grade 4 Volume 1: Mathematics \\ \hline
        90 & [Beijing all-in-one study, practice and examination] 2021 Edition Grade 4 Volume 1: Mathematics \\ \hline
        91 & [All-in-one study, practice and examination] 2023-2024 school year Beijing Edition Grade 5 Volume 2: Mathematics \\ \hline
        92 & [All-in-one study, practice and examination] 2023-2024 school year People's Education Press Grade 5 Volume 2: Mathematics \\ \hline
        93 & [Beijing all-in-one study, practice and examination] 2023 Beijing Edition Grade 5 Volume 1: Mathematics \\ \hline
        94 & [Beijing all-in-one study, practice and examination] 2023 Printing People's Education Edition Grade 5 Volume 1: Mathematics \\ \hline
        95 & [Beijing all-in-one study, practice and examination] 2022 Beijing Edition Grade 5, Volume 2: Mathematics \\ \hline
        96 & [Beijing all-in-one study, practice and examination] 2022 Edition of People's Education Edition Grade 5 Volume 2: Mathematics \\ \hline
        97 & [All-in-one study, practice and examination] 2022 Beijing Normal University Edition Grade 5 Volume 1: Mathematics \\ \hline
        98 & [Beijing all-in-one study, practice and examination] 2021 Edition Grade 5 Volume 1: Mathematics \\ \hline
        99 & [All-in-one study, practice and examination] 2023-2024 school year Beijing edition sixth grade volume 2: Mathematics \\ \hline
        100 & [All-in-one study, practice and examination] 2023-2024 school year People's Education Press Grade 6 Volume 2: Mathematics \\ \hline
    \end{tabular}
\end{table}

\begin{table}[!ht]
 \caption{The data sources of \dataset (Part3, Source 101 to 150).}
    \centering
    \tiny
           \begin{tabular}{|c|p{0.9\textwidth}|} 
       \toprule
    \textbf{Number} & \textbf{Data Source} \\ 
       \hline
101 & [Beijing all-in-one study, practice and examination] 2023 Beijing Edition Sixth Grade Volume 1: Mathematics \\ \hline
        102 & [Beijing all-in-one study, practice and examination] 2023 Printing People's Education Press Sixth Grade Volume 1: Mathematics \\ \hline
        103 & [Beijing all-in-one study, practice and examination] 2022 Beijing Edition Grade 6 Volume 2: Mathematics \\ \hline
        104 & [Beijing all-in-one study, practice and examination] 2022 Edition of People's Education Press Sixth Grade Volume 2: Mathematics \\ \hline
        105 & [Beijing all-in-one study, practice and examination] 2022 Printing Beijing Normal University Edition Grade 6 Volume 2: Mathematics \\ \hline
        106 & [Beijing all-in-one study, practice and examination] 2022 Beijing Normal University Edition Sixth Grade Volume 1: Mathematics \\ \hline
        107 & [Beijing all-in-one study, practice and examination] 2021 Edition Sixth Grade Volume 1: Mathematics \\ \hline
        108 & [Beijing Class Workbook] 2023 Beijing Edition Grade 4 Volume 1: Mathematics \\ \hline
        109 & [Class Workbook] 2023-2024 Academic Year Beijing Edition Grade 4 Volume 2: Mathematics \\ \hline
        110 & [Zhejiang Class Workbook] 2022 Edition People's Education Edition Grade 4 Volume 1: Mathematics \\ \hline
        111 & [Class Workbook] 2023-2024 Beijing Edition Grade 5, Volume 2: Mathematics \\ \hline
        112 & [Beijing Class Workbook] 2023 Beijing Edition Grade 5 Volume 1: Mathematics \\ \hline
        113 & [Zhejiang Class Workbook] 2022 Beijing Normal University Edition Grade 5 Volume 1: Mathematics \\ \hline
        114 & [Zhejiang Class Workbook] 2022 Edition People's Education Edition Grade 5 Volume 1: Mathematics \\ \hline
        115 & [Beijing Class Workbook] 2023 Beijing Edition Grade 6 Volume 1: Mathematics \\ \hline
        116 & [Class Workbook] 2023-2024 Academic Year Beijing Edition Grade 6 Volume 2: Mathematics \\ \hline
        117 & [Zhejiang Class Workbook] 2022 Beijing Normal University Edition Sixth Grade Volume 1: Mathematics \\ \hline
        118 & [Zhejiang Class Workbook] 2022 Edition People's Education Edition Sixth Grade Volume 1: Mathematics \\ \hline
        119 & [Mathematics Textbook] 2023-2024 Academic Year People's Education Press Grade 3 Volume 2: Mathematics \\ \hline
        120 & [Mathematics Textbook] 2022 Shanghai Education Edition Grade 3 Volume 1: Mathematics \\ \hline
        121 & [Beijing Mathematics Textbook] 2022 Beijing Edition Grade 3 Volume 1: Mathematics \\ \hline
        122 & [Beijing Mathematics Textbook] 2022 Beijing Normal University Edition Grade 3 Volume 1: Mathematics \\ \hline
        123 & [Shanghai Mathematics Textbook] 2021 Shanghai Education Edition Grade 3 Volume 2: Mathematics \\ \hline
        124 & [Beijing Mathematics Textbook] 2021 Beijing Edition Grade 3 Volume 2: Mathematics \\ \hline
        125 & [Mathematics Textbook] 2020 Beijing Normal University Edition Grade 3 Volume 2: Mathematics \\ \hline
        126 & [Beijing Mathematics Textbook] 2020 Edition People's Education Press Grade 3 Volume 1: Mathematics \\ \hline
        127 & [Mathematics Textbook] 2023-2024 Academic Year People's Education Press Grade 4 Volume 2: Mathematics \\ \hline
        128 & [Shanghai Mathematics Textbook] 2022 Shanghai Education Edition Grade 4 Volume 1: Mathematics \\ \hline
        129 & [Beijing Mathematics Textbook] 2022 Beijing Edition Grade 4 Volume 1: Mathematics \\ \hline
        130 & [Beijing Mathematics Textbook] 2022 Beijing Normal University Edition Grade 4 Volume 1: Mathematics \\ \hline
        131 & [Shanghai Mathematics Textbook] 2021 Shanghai Education Edition Grade 4 Volume 2: Mathematics \\ \hline
        132 & [Beijing Mathematics Textbook] 2021 Beijing Edition Grade 4 Volume 2: Mathematics \\ \hline
        133 & [Beijing Mathematics Textbook] 2020 Beijing Normal University Edition Grade 4 Volume 2: Mathematics \\ \hline
        134 & [Beijing Mathematics Textbook] 2019 Edition People's Education Press Grade 4 Volume 1: Mathematics \\ \hline
        135 & [Mathematics Textbook] 2023-2024 Academic Year People's Education Edition Grade 5 Volume 2: Mathematics \\ \hline
        136 & [Shanghai Mathematics Textbook] 2022 Shanghai Education Edition Grade 5 Volume 1: Mathematics \\ \hline
        137 & [Mathematics Textbook] 2022 Beijing Normal University Edition Grade 5 Volume 1: Mathematics \\ \hline
        138 & [Shanghai Mathematics Textbook] 2022 Shanghai Education Edition Grade 5 Volume 1: Mathematics \\ \hline
        139 & [Mathematics Textbook] 2022 Beijing Normal University Edition Grade 5 Volume 1: Mathematics \\ \hline
        140 & [Beijing Mathematics Textbook] 2022 Beijing Edition Grade 5 Volume 1: Mathematics \\ \hline
        141 & [Beijing Mathematics Textbook] 2021 Beijing Edition Grade 5 Volume 2: Mathematics \\ \hline
        142 & [Shanghai Mathematics Textbook] 2020 Shanghai Education Edition Grade 5 Volume 2: Mathematics \\ \hline
        143 & [Beijing Mathematics Textbook] 2020 Beijing Normal University Edition Grade 5 Volume 2: Mathematics \\ \hline
        144 & [Beijing Mathematics Textbook] 2020 Edition People's Education Edition Grade 5 Volume 1: Mathematics \\ \hline
        145 & [Beijing Mathematics Textbook] 2022 Edition People's Education Press Sixth Grade Volume 2: Mathematics \\ \hline
        146 & [Beijing Mathematics Textbook] 2022 Beijing Edition Sixth Grade Volume 1: Mathematics \\ \hline
        147 & [Mathematics Textbook] 2022 Beijing Normal University Edition Grade 6 Volume 1: Mathematics \\ \hline
        148 & [Beijing Mathematics Textbook] 2021 Beijing Normal University Edition Grade 6 Volume 2: Mathematics \\ \hline
        149 & [Beijing Mathematics Textbook] 2021 Beijing Edition Sixth Grade Volume 2: Mathematics \\ \hline
        150 & [Beijing Mathematics Textbook] 2018 Edition People's Education Press Sixth Grade Volume 1: Mathematics \\ \hline
    \end{tabular}
\end{table}

\begin{table}[!ht]
 \caption{The data sources of \dataset (Part4, Source 151 to 200).}
    \centering
    \tiny
         \begin{tabular}{|c|p{0.9\textwidth}|} 
    \toprule
    \textbf{Number} & \textbf{Data Source} \\ 
    \hline
    151 & 2022 Sichuan Liangshan Primary School to Junior High School Examination Paper (People's Education Edition): Mathematics \\ \hline
        152 & 2022 Chongqing Yubei District Primary School to Junior High School Examination Paper (People's Education Edition Examination): Mathematics \\ \hline
        153 & 2022 Guizhou Qiandongnan Primary School to Junior High School Examination Paper (People's Education Edition Examination): Mathematics \\ \hline
        154 & 2022 Anhui Fuyang Taihe County Primary School to Junior High School Examination Paper (Beijing Normal University Edition Examination): Mathematics \\ \hline
        155 & 2022 Guangdong Huizhou Huiyang District Primary School to Junior High School Examination Paper (Beijing Normal University Edition): Mathematics \\ \hline
        156 & 2022 Guangdong Shaoguan Xinfeng County Primary School to Junior High School Examination Paper (People's Education Edition): Mathematics \\ \hline
        157 & 2022 Guangdong Zhanjiang Mazhang District Primary School to Junior High School Examination Paper (Beijing Normal University Edition Examination): Mathematics \\ \hline
        158 & 2022 Gansu Dingxi Minxian Primary School to Junior High School Examination Paper (Beijing Normal University Edition): Mathematics \\ \hline
        159 & 2022 Guangdong Jieyang Jiedong District Primary School to Junior High School Examination Paper (Beijing Normal University Edition Examination): Mathematics \\ \hline
        160 & 2022 Henan Luohe Wuyang County Primary School Entrance Examination Paper (People's Education Edition): Mathematics \\ \hline
        161 & 2022 Tianjin Primary School to Junior High School Examination Paper (Primary School to Junior High School in Some Districts): Mathematics \\ \hline
        162 & 2022 Hebei Tangshan Lunan District Primary School Entrance Examination Paper (Hebei Education Edition Examination): Mathematics \\ \hline
        163 & 2022 Hebei Baoding Qingyuan District Primary School to Junior High School Examination Paper (People's Education Edition): Mathematics \\ \hline
        164 & 2022 Xinjiang Turpan Primary School to Junior High School Examination Paper (People's Education Edition): Mathematics \\ \hline
        165 & 2022 Hebei Shijiazhuang Luquan District Primary School Entrance Examination Paper (Hebei Education Edition): Mathematics \\ \hline
        166 & 2022 Hainan Ledong Li Autonomous County Primary School to Junior High School Examination Paper (People's Education Edition Examination): Mathematics \\ \hline
        167 & 2022 Henan Jiyuan Primary School to Junior High School Examination Paper (People's Education Edition): Mathematics \\ \hline
        168 & 2021 Beijing Fengtai District Primary School to Junior High School Examination Paper (People's Education Edition Examination): Mathematics \\ \hline
        169 & 2021 Yunnan Kunming Wuhua District Primary School to Junior High School Examination Paper: Mathematics \\ \hline
        170 & 2021 Yunnan Kunming Xishan District Primary School Entrance Examination Paper: Mathematics \\ \hline
        171 & 2021 Shaanxi Xi'an Beilin District Primary School to Junior High School Examination Paper (Part 2): Mathematics \\ \hline
        172 & 2021 Shaanxi Xi'an Beilin District Primary School to Junior High School Examination Paper: Mathematics \\ \hline
        173 & 2021 Zhejiang Ningbo Haishu District Primary School to Junior High School Examination Paper: Mathematics \\ \hline
        174 & 2021 Shaanxi Xi'an Weiyang District Primary School to Junior High School Examination Paper: Mathematics \\ \hline
        175 & 2021 Shaanxi Xi'an Yanta District Primary School to Junior High School Examination Paper (Part 5): Mathematics \\ \hline
        176 & 2021 Shaanxi Xi'an Beilin District Primary School to Junior High School Examination Paper: Mathematics \\ \hline
        177 & "2021 Primary to Junior High School Examination Paper in Baqiao District, Xi'an, Shaanxi: Mathematics \\ \hline
        178 & 2021 Shaanxi Xi'an Weiyang District Primary School to Junior High School Examination Paper (Part 3): Mathematics \\ \hline
        179 & "2021 Primary to Junior High School Examination Paper in Erqi District, Zhengzhou, Henan: Mathematics \\ \hline
        180 & 2021 Jiangsu Nantong Primary School to Junior High School Examination Paper (Main Urban Area): Mathematics \\ \hline
        181 & 2021 Jiangsu Suzhou Xiangcheng District Primary School to Junior High School Examination Paper 5 to 6 Direct Promotion Class: Mathematics \\ \hline
        182 & "2021 Primary to Junior High School Examination Paper for Baqiao District, Xi'an, Shaanxi (Part 5): Mathematics \\ \hline
        183 & 2021 Hunan Changsha Yuhua District Primary School to Junior High School Examination Paper: Mathematics \\ \hline
        184 & "2021 Primary to Junior High School Examination Paper of Jingxing County, Shijiazhuang, Hebei (Hebei Education Edition): Mathematics \\ \hline
        185 & 2021 Hebei Shijiazhuang Pingshan County Primary School Entrance Examination Paper (Hebei Education Edition): Mathematics \\ \hline
        186 & 2021 Hebei Shijiazhuang Lingshou County Primary School to Junior High School Examination Paper (Hebei Education Edition): Mathematics \\ \hline
        187 & "2021 Primary to Junior High School Examination Paper in Yanta District, Xi'an, Shaanxi: Mathematics \\ \hline
        188 & 2021 Shaanxi Xi'an Xincheng District Primary School to Junior High School Examination Paper (Part 3): Mathematics \\ \hline
        189 & "2021 Primary to Junior High School Examination Paper of Yuanshi County, Shijiazhuang, Hebei (Hebei Education Edition): Mathematics \\ \hline
        190 & "2021 Primary to Junior High School Examination Paper in Zhengding County, Shijiazhuang, Hebei (Hebei Education Edition): Mathematics \\ \hline
        191 & 2021 Shaanxi Xi'an Yanta District Primary School to Junior High School Examination Paper (Part 14): Mathematics \\ \hline
        192 & 2021 Chongqing Shapingba District Chongqing Nankai Middle School Primary School to Junior High School Examination Paper (Part 3): Mathematics \\ \hline
        193 & 2021 Shaanxi Xi'an Yanta District Primary School to Junior High School Examination Paper (Part 4A): Mathematics \\ \hline
        194 & "2021 Primary to Junior High School Examination Paper in Yanta District, Xi'an, Shaanxi: Mathematics \\ \hline
        195 & 2021 Shaanxi Xi'an Beilin District Primary School to Junior High School Examination Paper (II) (Hanguang Campus): Mathematics \\ \hline
        196 & 2020 Beijing Chaoyang District Primary School to Junior High School Examination Paper: Mathematics \\ \hline
        197 & 2020 Beijing Haidian District Primary School to Junior High School Examination Paper (Paper A): Mathematics \\ \hline
        198 & 2020 Beijing Haidian District Primary School to Junior High School Examination Paper (Paper B): Mathematics \\ \hline
        199 & 2020 Beijing Haidian District Primary School to Junior High School Examination Paper: Mathematics \\ \hline
        200 & 2020 Beijing Changping District Primary School to Junior High School Examination Paper: Mathematics \\ \hline

    \end{tabular}
\end{table}

\begin{table}[!ht]
 \caption{The data sources of \dataset (Part5, Source 201 to 223).}
 \label{tab:source5}
    \centering
    \tiny
           \begin{tabular}{|c|p{0.9\textwidth}|} 
    \toprule
    \textbf{Number} & \textbf{Data Source} \\ 
    \hline
201 & 2020 Hunan Changsha Yuhua District Yali Experimental Middle School Primary to Junior High School Mathematics Test Paper \\ \hline
        202 & 2020 Shenzhen Futian District Shenzhen Senior High School Primary to Junior High School Mathematics Test Paper \\ \hline
        203 & 2020 Heilongjiang Jixi Hulin City Primary School Entrance Examination Paper: Mathematics \\ \hline
        204 & 2020 Heilongjiang Qiqihar Primary School to Junior High School Examination Paper: Mathematics \\ \hline
        205 & 2020 Ningxia Wuzhong Hongsibao District Primary School Entrance Examination Paper (People's Education Edition): Mathematics \\ \hline
        206 & 2020 Liaoning Fushun Wanghua District Primary School to Junior High School Examination Paper: Mathematics \\ \hline
        207 & "2020 Primary to Junior High School Examination Paper in Xincheng District, Hohhot, Inner Mongolia: Mathematics \\ \hline
        208 & 2020 Guangdong Zhaoqing Huaiji County Primary School to Junior High School Examination Paper: Mathematics \\ \hline
        209 & 2020 Guangdong Zhaoqing Huaiji County Primary School to Junior High School Examination Paper: Mathematics \\ \hline
        210 & 2020 Fujian Quanzhou Licheng District Primary School Entrance Examination Paper: Mathematics \\ \hline
        211 & "2020 Primary to Junior High School Examination Paper in Huimin District, Hohhot, Inner Mongolia: Mathematics \\ \hline
        212 & 2020 Hebei Baoding Jingxiu District Primary School to Junior High School Examination Paper: Mathematics \\ \hline
        213 & 2020 Sichuan Chengdu Pidu District Primary School to Junior High School Examination Paper (Part 1): Mathematics \\ \hline
        214 & 2020 Sichuan Mianyang Youxian District Primary School Entrance Examination Paper (Part 6): Mathematics \\ \hline
        215 & 2020 Guangdong Shaoguan Zhenjiang District Primary School to Junior High School Examination Paper: Mathematics \\ \hline
        216 & 2020 Sichuan Mianyang Fucheng District Primary School to Junior High School Examination Paper (Part 10): Mathematics \\ \hline
        217 & 2020 Sichuan Mianyang Fucheng District Primary School to Junior High School Examination Paper (Part 5): Mathematics \\ \hline
        218 & 2020 Sichuan Chengdu Pidu District Primary School to Junior High School Examination Paper (Part 5): Mathematics \\ \hline
        219 & 2020 Hunan Changsha Yuelu District Primary School to Junior High School Examination Paper: Mathematics \\ \hline
        220 & 2020 Hunan Changsha Yuelu District Primary School to Junior High School Examination Paper: Mathematics \\ \hline
        221 & 2020 Gansu Lanzhou Chengguan District Primary School to Junior High School Examination Paper (Class Examination): Mathematics \\ \hline
        222 & 2020 Hunan Changsha Primary School to Junior High School Examination Paper: Mathematics \\ \hline
        223 & 2020 Liaoning Shenyang Primary School Primary Exam Paper: Mathematics \\ \hline
    \end{tabular}
\end{table}
\clearpage

\end{document}